\documentclass[journal]{IEEEtran}
\IEEEoverridecommandlockouts
\usepackage{amsmath,amssymb,amsfonts}
\usepackage{algorithmic}
\usepackage{algorithm}
\usepackage{array}
\usepackage{float}
\usepackage{textcomp}
\usepackage{stfloats} 
\usepackage{url} 
\usepackage{verbatim}
\usepackage{cite}
\usepackage{multirow}
\usepackage{multicol}
\usepackage{xcolor}
\usepackage{pdfpages}
\usepackage{amsmath,amssymb}
\usepackage[flushleft]{threeparttable}
\usepackage{array,booktabs,makecell}
\definecolor{colred}{rgb}{1,0,0}

\usepackage{amsmath,graphicx}
\usepackage{color}
\usepackage{epstopdf}
\usepackage{tikz}
\usepackage{balance}
\usepackage{url}
\usepackage{multirow,hhline}
\usepackage{placeins}
\usepackage{wasysym}
\usepackage{amssymb}
\usepackage[colorinlistoftodos]{todonotes}

\definecolor{colSergio}{rgb}{1., 0.8, 0.8}

\newcommand{\image}{\pgfuseimage}
\newcommand{\fpathTropi}{TropiSAR/}
\newcommand{\fpathTropisingle}{TropiSAR_singlepol/}
\newcommand{\fpathTropidual}{TropiSAR_dualpol/}
\newcommand{\fpathTropifine}{TropiSAR_finetuning/}
\newcommand{\fpathAfri}{AfriSAR/}
\newcommand{\fpathAfrifine}{AfriSAR_finetuning/}
\newcommand{\fpathAfrithreeD}{AfriSAR_3D/}
\newcommand{\fpathnewthreeDDSM}{3DDSM_withLiDARCHM/}
\newcommand{\fpathPauli}{Pauli/}

\pgfdeclareimage[width=0.45\columnwidth]{TropiSAR_Pauli}{\fpathPauli TropiSAR_Pauli.png}
\pgfdeclareimage[width=0.45\columnwidth]{TropiSAR_DTM}{\fpathPauli TropiSAR_DTM.png}
\pgfdeclareimage[width=0.45\columnwidth]{TropiSAR_CHM}{\fpathPauli TropiSAR_CHM.png}
\pgfdeclareimage[width=0.45\columnwidth]{AfriSAR_Pauli}{\fpathPauli AfriSAR_Pauli.png}
\pgfdeclareimage[width=0.45\columnwidth]{AfriSAR_DTM}{\fpathPauli AfriSAR_DTM.png}
\pgfdeclareimage[width=0.45\columnwidth]{AfriSAR_CHM}{\fpathPauli AfriSAR_CHM.png}

\pgfdeclareimage[width=0.2\textwidth]{LiDAR_CHM_2D}{\fpathTropi LiDAR_CHM_2D.png}
\pgfdeclareimage[width=0.2\textwidth]{unet_CHM_2D}{\fpathTropi unet_CHM_2D.png}
\pgfdeclareimage[width=0.2\textwidth]{FCN_CHM_2D}{\fpathTropi FCN_CHM_2D.png}
\pgfdeclareimage[width=0.2\textwidth]{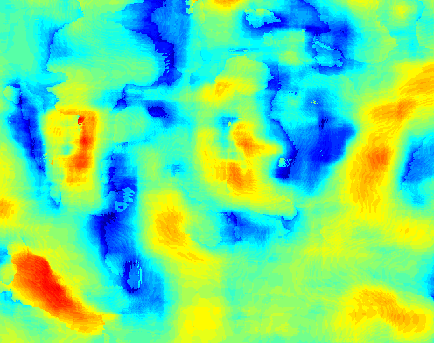}{\fpathTropi m2_CHM_2D.png}
\pgfdeclareimage[width=0.2\textwidth]{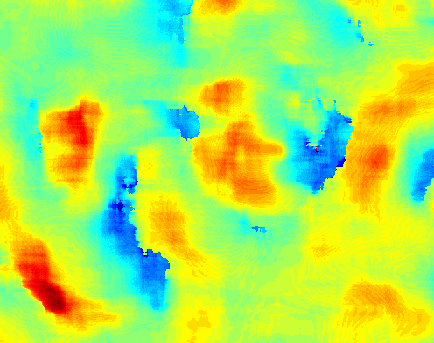}{\fpathTropi skp_CHM_2D.png}
\pgfdeclareimage[width=0.2\textwidth]{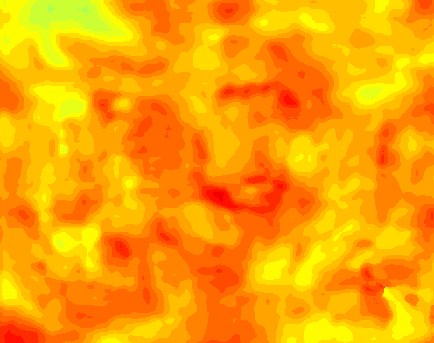}{\fpathTropi BOSS_CHM_2D.png}
\pgfdeclareimage[width=0.2\textwidth]{LiDAR_DTM_2D}{\fpathTropi LiDAR_DTM_2D.png}
\pgfdeclareimage[width=0.2\textwidth]{unet_DTM_2D}{\fpathTropi unet_DTM_2D.png}
\pgfdeclareimage[width=0.2\textwidth]{FCN_DTM_2D}{\fpathTropi FCN_DTM_2D.png}
\pgfdeclareimage[width=0.2\textwidth]{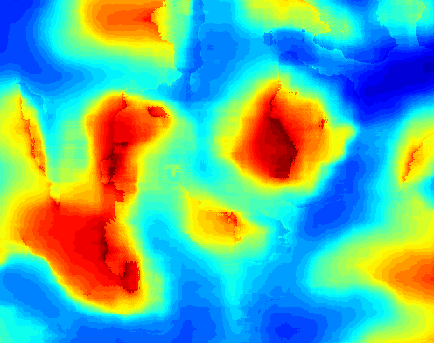}{\fpathTropi m2_DTM_2D.png}
\pgfdeclareimage[width=0.2\textwidth]{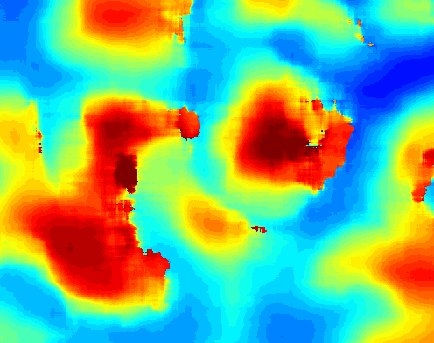}{\fpathTropi skp_DTM_2D.png}
\pgfdeclareimage[width=0.2\textwidth]{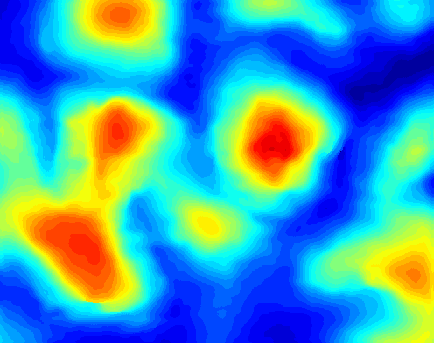}{\fpathTropi BOSS_DTM_2D.png}

\pgfdeclareimage[width=0.2\textwidth]{LiDAR_SUM_2D}{\fpathTropi LiDAR_SUM_2D.png}
\pgfdeclareimage[width=0.2\textwidth]{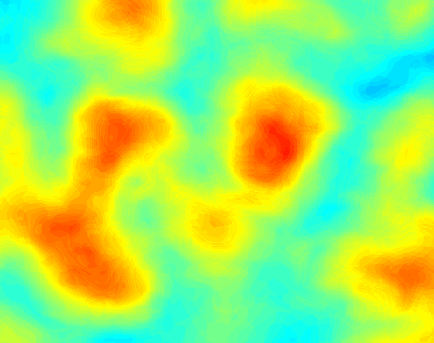}{\fpathTropi unet_SUM_2D.png}
\pgfdeclareimage[width=0.2\textwidth]{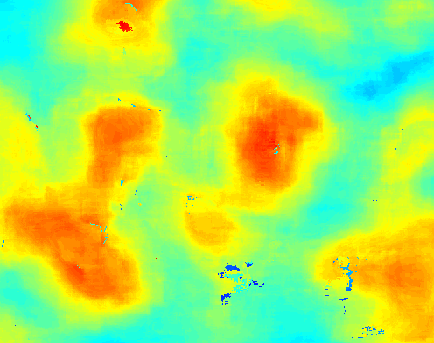}{\fpathTropi FCN_SUM_2D.png}
\pgfdeclareimage[width=0.2\textwidth]{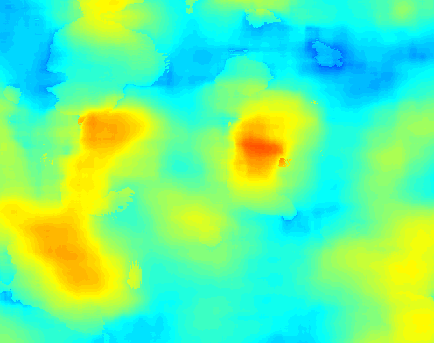}{\fpathTropi m2_SUM_2D.png}
\pgfdeclareimage[width=0.2\textwidth]{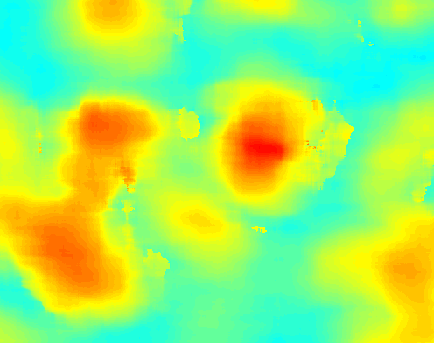}{\fpathTropi skp_SUM_2D.png}
\pgfdeclareimage[width=0.2\textwidth]{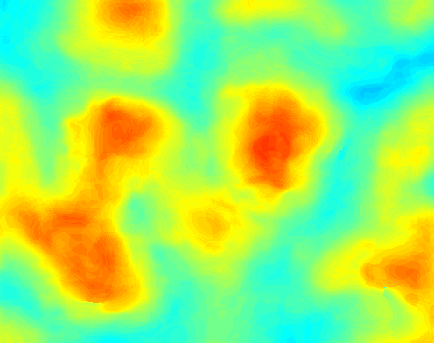}{\fpathTropi BOSS_SUM_2D.png}

\pgfdeclareimage[width=0.42\columnwidth]{LiDAR_SUM_3D}{\fpathTropi LiDAR_SUM_3D.png}
\pgfdeclareimage[width=0.42\columnwidth]{unet_SUM_3D}{\fpathTropi unet_SUM_3D.png}
\pgfdeclareimage[width=0.42\columnwidth]{FCN_SUM_3D}{\fpathTropi FCN_SUM_3D.png}
\pgfdeclareimage[width=0.42\columnwidth]{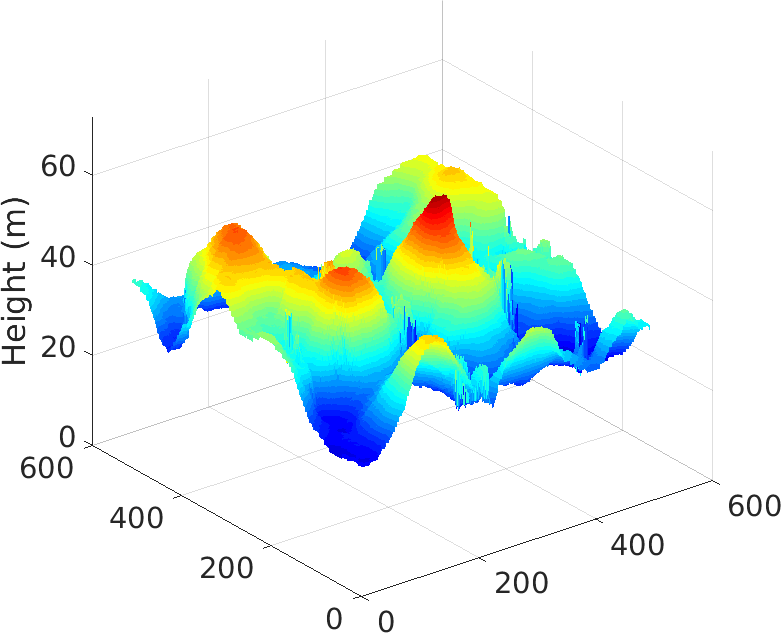}{\fpathTropi skp_SUM_3D.png}
\pgfdeclareimage[width=0.42\columnwidth]{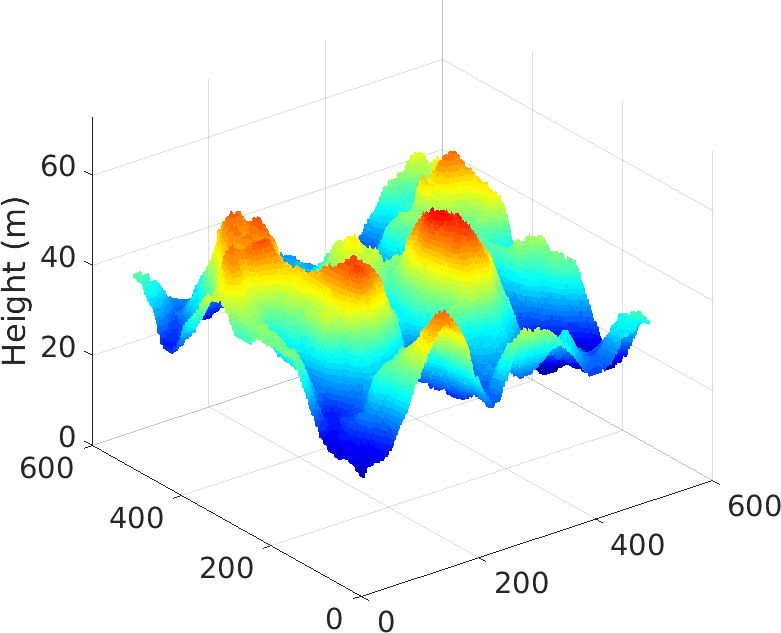}{\fpathTropi BOSS_SUM_3D.png}
\pgfdeclareimage[width=0.42\columnwidth]{LiDAR_DTM_3D}{\fpathTropi LiDAR_DTM_3D.png}
\pgfdeclareimage[width=0.42\columnwidth]{unet_DTM_3D}{\fpathTropi unet_DTM_3D.png}
\pgfdeclareimage[width=0.42\columnwidth]{FCN_DTM_3D}{\fpathTropi FCN_DTM_3D.png}
\pgfdeclareimage[width=0.42\columnwidth]{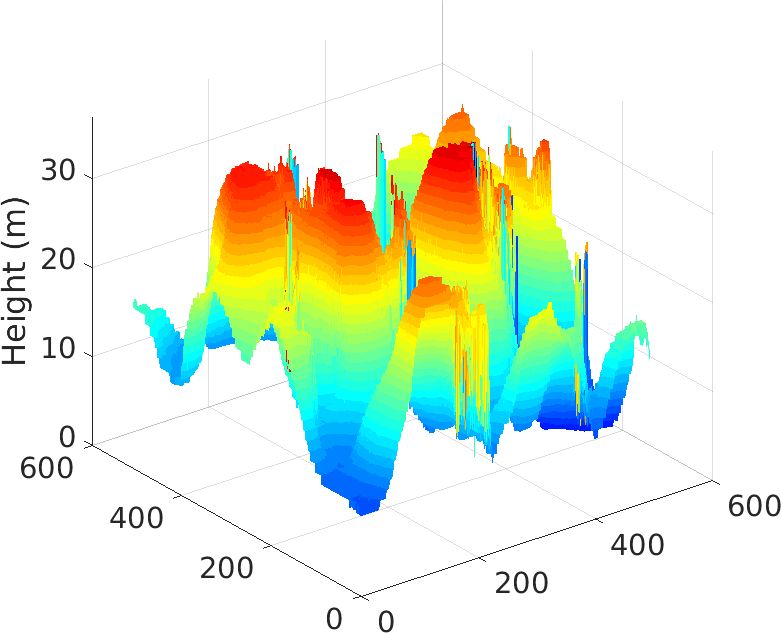}{\fpathTropi skp_DTM_3D.png}
\pgfdeclareimage[width=0.42\columnwidth]{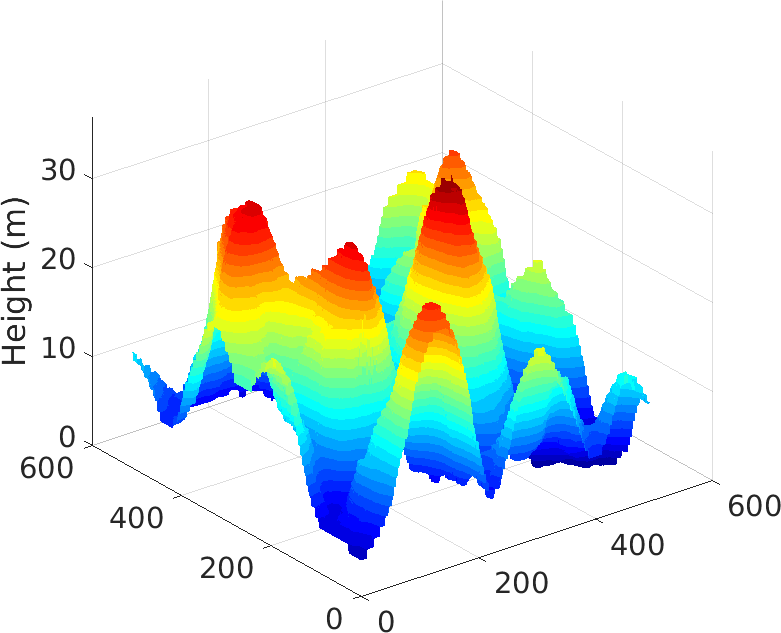}{\fpathTropi BOSS_DTM_3D.png}

\pgfdeclareimage[width=0.42\columnwidth]{LiDAR_CHM_3D}{\fpathTropi LiDAR_CHM_3D.png}
\pgfdeclareimage[width=0.42\columnwidth]{unet_CHM_3D}{\fpathTropi unet_CHM_3D.png}
\pgfdeclareimage[width=0.42\columnwidth]{fcn_CHM_3D}{\fpathTropi FCN_CHM_3D.png}
\pgfdeclareimage[width=0.42\columnwidth]{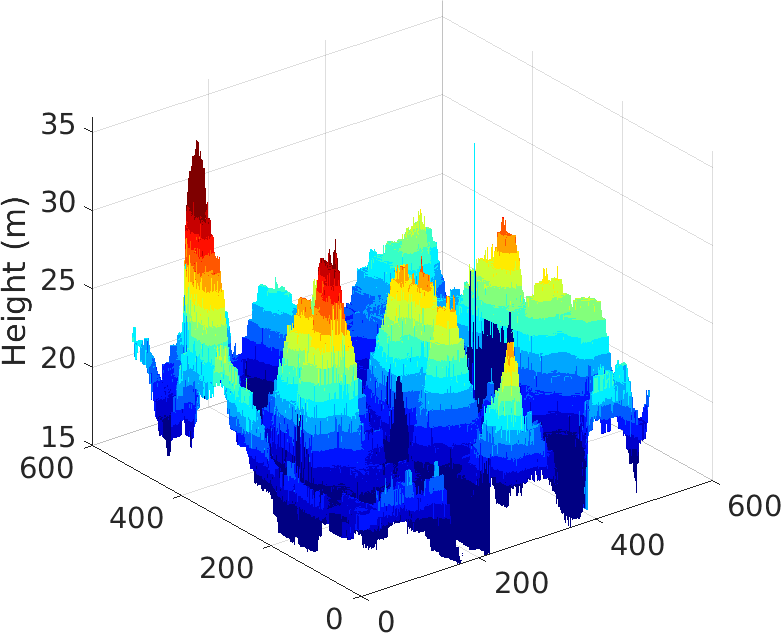}{\fpathTropi skp_CHM_3D.png}
\pgfdeclareimage[width=0.42\columnwidth]{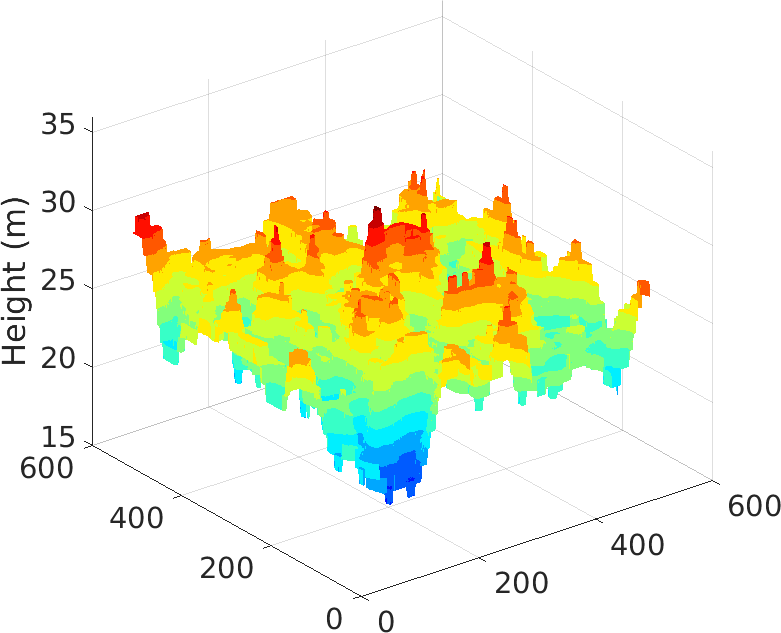}{\fpathTropi BOSS_CHM_3D.png}

\pgfdeclareimage[width=0.2100\textwidth]{CHM_profile_1}{\fpathTropi CHM_profile_line1.png}
\pgfdeclareimage[width=0.2100\textwidth]{CHM_profile_170}{\fpathTropi CHM_profile_line170.png}
\pgfdeclareimage[width=0.2100\textwidth]{CHM_profile_280}{\fpathTropi CHM_profile_line280.png}
\pgfdeclareimage[width=0.2100\textwidth]{DTM_profile_1}{\fpathTropi DTM_profile_line1.png}
\pgfdeclareimage[width=0.2100\textwidth]{DTM_profile_170}{\fpathTropi DTM_profile_line170.png}
\pgfdeclareimage[width=0.2100\textwidth]{DTM_profile_280}{\fpathTropi DTM_profile_line280.png}
\pgfdeclareimage[width=0.2100\textwidth]{SUM_profile_1}{\fpathTropi SUM_profile_line1.png}
\pgfdeclareimage[width=0.2100\textwidth]{SUM_profile_170}{\fpathTropi SUM_profile_line170.png}
\pgfdeclareimage[width=0.2100\textwidth]{SUM_profile_280}{\fpathTropi SUM_profile_line280.png}

\pgfdeclareimage[width=0.42\columnwidth]{CHM_joint_unet}{\fpathTropi CHM_jointd_unet.png}
\pgfdeclareimage[width=0.42\columnwidth]{CHM_joint_fcn}{\fpathTropi CHM_jointd_fcn.png}
\pgfdeclareimage[width=0.42\columnwidth]{CHM_joint_skp}{\fpathTropi CHM_jointd_skp.png}
\pgfdeclareimage[width=0.210\textwidth]{CHM_joint_m2}{\fpathTropi CHM_jointd_m2.png}
\pgfdeclareimage[width=0.210\textwidth]{CHM_joint_boss}{\fpathTropi CHM_jointd_boss.png}
\pgfdeclareimage[width=0.42\columnwidth]{DTM_joint_unet}{\fpathTropi DTM_jointd_unet.png}
\pgfdeclareimage[width=0.42\columnwidth]{DTM_joint_fcn}{\fpathTropi DTM_jointd_fcn.png}
\pgfdeclareimage[width=0.42\columnwidth]{DTM_joint_skp}{\fpathTropi DTM_jointd_skp.png}
\pgfdeclareimage[width=0.210\textwidth]{DTM_joint_m2}{\fpathTropi DTM_jointd_m2.png}
\pgfdeclareimage[width=0.210\textwidth]{DTM_joint_boss}{\fpathTropi DTM_jointd_boss.png}
\pgfdeclareimage[width=0.42\columnwidth]{SUM_joint_unet}{\fpathTropi SUM_jointd_unet.png}
\pgfdeclareimage[width=0.42\columnwidth]{SUM_joint_fcn}{\fpathTropi SUM_jointd_fcn.png}
\pgfdeclareimage[width=0.42\columnwidth]{SUM_joint_skp}{\fpathTropi SUM_jointd_skp.png}
\pgfdeclareimage[width=0.210\textwidth]{SUM_joint_m2}{\fpathTropi SUM_jointd_m2.png}
\pgfdeclareimage[width=0.210\textwidth]{SUM_joint_boss}{\fpathTropi SUM_jointd_boss.png}

\pgfdeclareimage[width=0.199\textwidth]{LiDAR_CHM_2D_BIG}{\fpathTropi LiDAR_CHM_2D.png}
\pgfdeclareimage[width=0.199\textwidth]{LiDAR_DTM_2D_BIG}{\fpathTropi LiDAR_DTM_2D.png}

\pgfdeclareimage[width=0.199\textwidth]{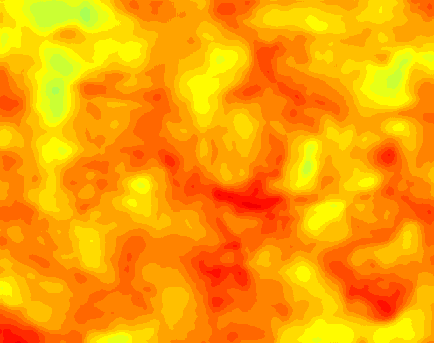}{\fpathTropisingle unet_CHM_2D_HH.png}
\pgfdeclareimage[width=0.199\textwidth]{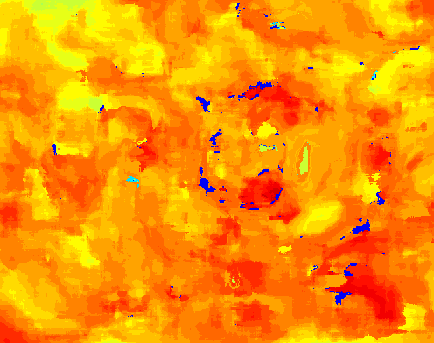}{\fpathTropisingle FCN_CHM_2D_HH.png}
\pgfdeclareimage[width=0.199\textwidth]{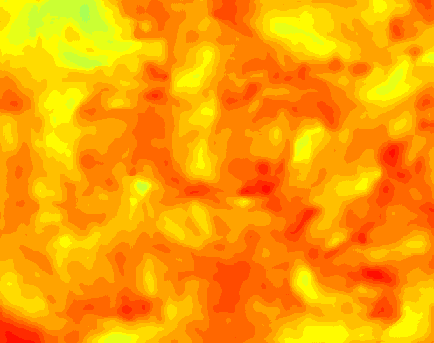}{\fpathTropisingle BOSS_CHM_2D_HH.png}
\pgfdeclareimage[width=0.199\textwidth]{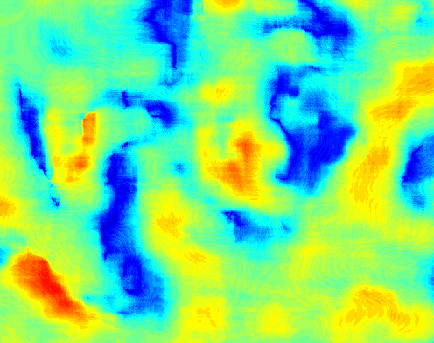}{\fpathTropisingle M2_CHM_2D_HH.png}

\pgfdeclareimage[width=0.199\textwidth]{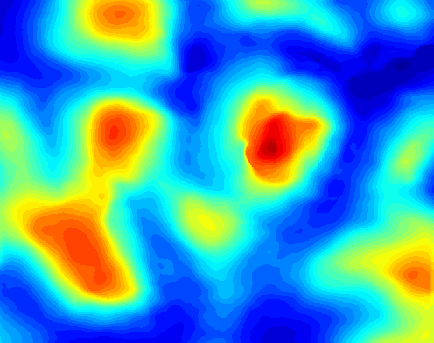}{\fpathTropisingle unet_DTM_2D_HH.png}
\pgfdeclareimage[width=0.199\textwidth]{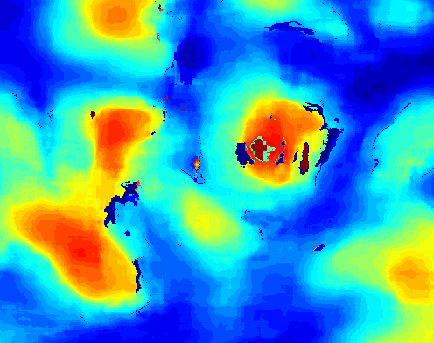}{\fpathTropisingle FCN_DTM_2D_HH.png}
\pgfdeclareimage[width=0.199\textwidth]{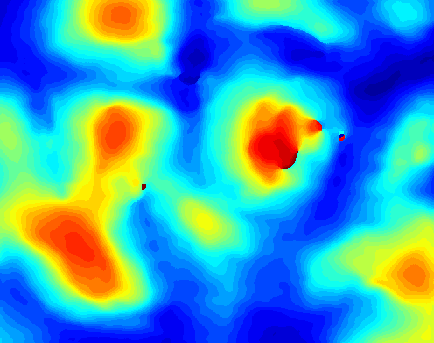}{\fpathTropisingle BOSS_DTM_2D_HH.png}
\pgfdeclareimage[width=0.199\textwidth]{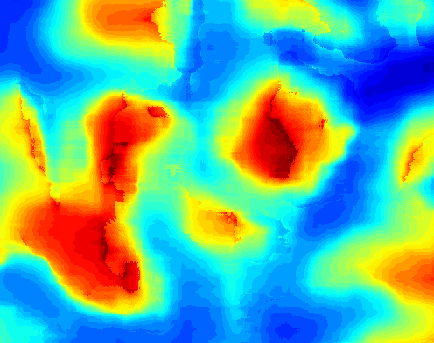}{\fpathTropisingle M2_DTM_2D_HH.png}

\pgfdeclareimage[width=0.199\textwidth]{LiDAR_SUM_2D_BIG}{\fpathTropi LiDAR_CHM_2D.png}
\pgfdeclareimage[width=0.199\textwidth]{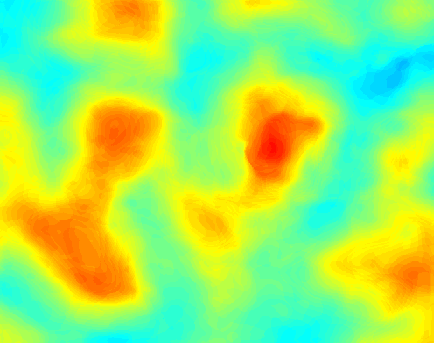}{\fpathTropisingle unet_SUM_2D_HH.png}
\pgfdeclareimage[width=0.199\textwidth]{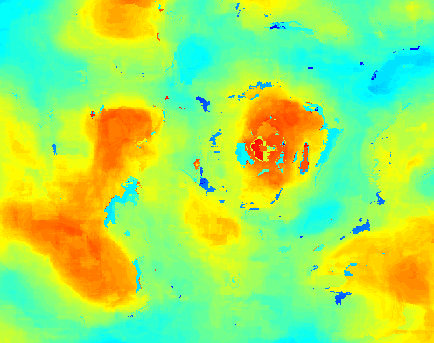}{\fpathTropisingle FCN_SUM_2D_HH.png}
\pgfdeclareimage[width=0.199\textwidth]{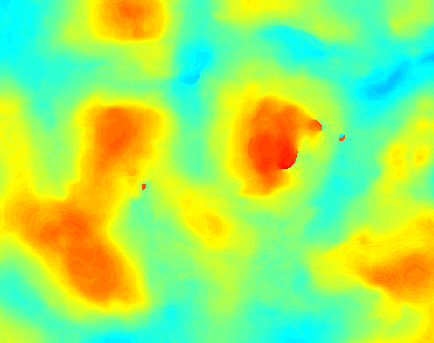}{\fpathTropisingle BOSS_SUM_2D_HH.png}
\pgfdeclareimage[width=0.199\textwidth]{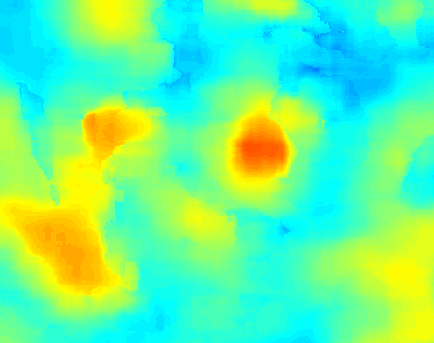}{\fpathTropisingle M2_SUM_2D_HH.png}

\pgfdeclareimage[width=0.199\textwidth]{LiDAR_SUM_3D_BIG}{\fpathTropi LiDAR_SUM_3D.png}
\pgfdeclareimage[width=0.199\textwidth]{LiDAR_DTM_3D_BIG}{\fpathTropi LiDAR_DTM_3D.png}

\pgfdeclareimage[width=0.199\textwidth]{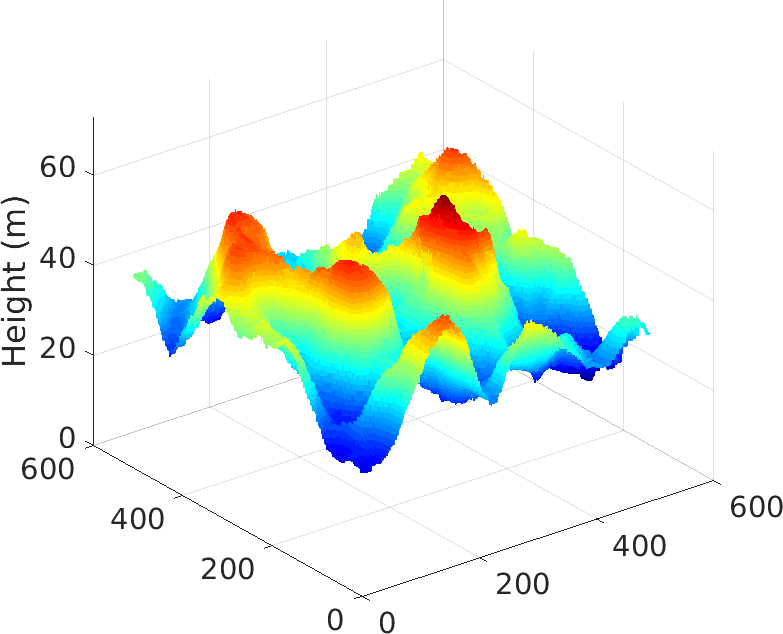}{\fpathTropisingle unet_SUM_3D_HH.png}
\pgfdeclareimage[width=0.199\textwidth]{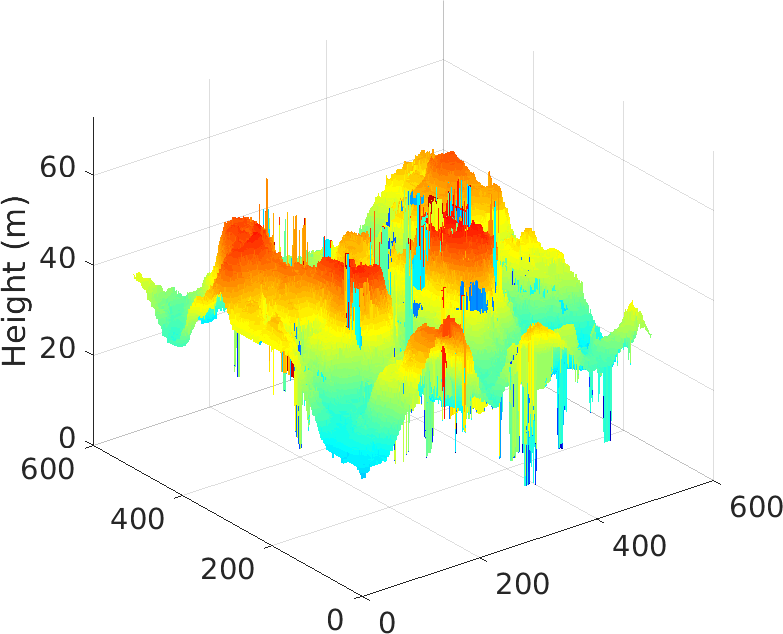}{\fpathTropisingle FCN_SUM_3D_HH.png}
\pgfdeclareimage[width=0.199\textwidth]{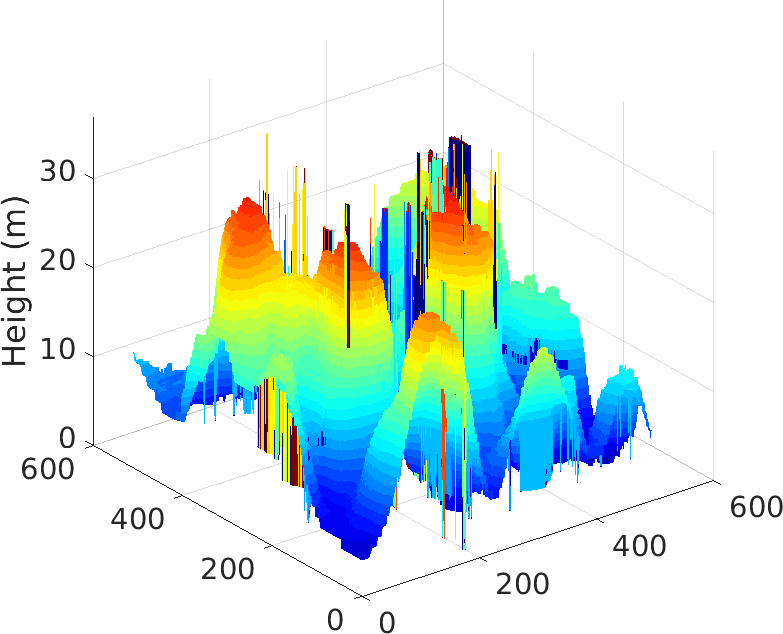}{\fpathTropisingle M2_SUM_3D_HH.png}

\pgfdeclareimage[width=0.199\textwidth]{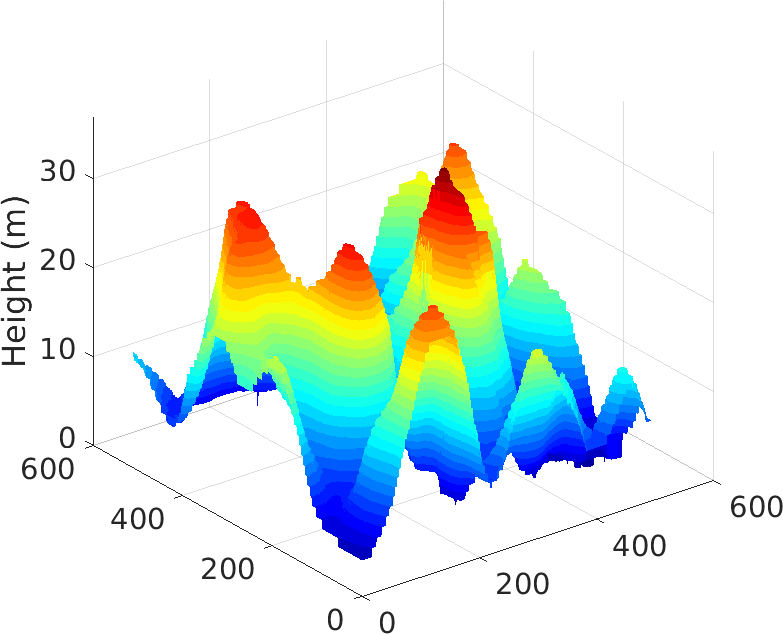}{\fpathTropisingle unet_DTM_3D_HH.png}
\pgfdeclareimage[width=0.199\textwidth]{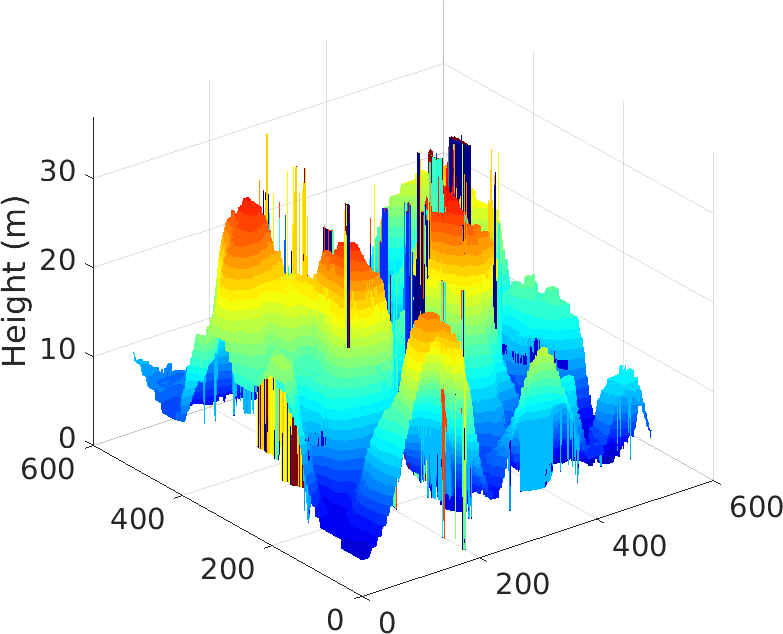}{\fpathTropisingle FCN_DTM_3D_HH.png}
\pgfdeclareimage[width=0.199\textwidth]{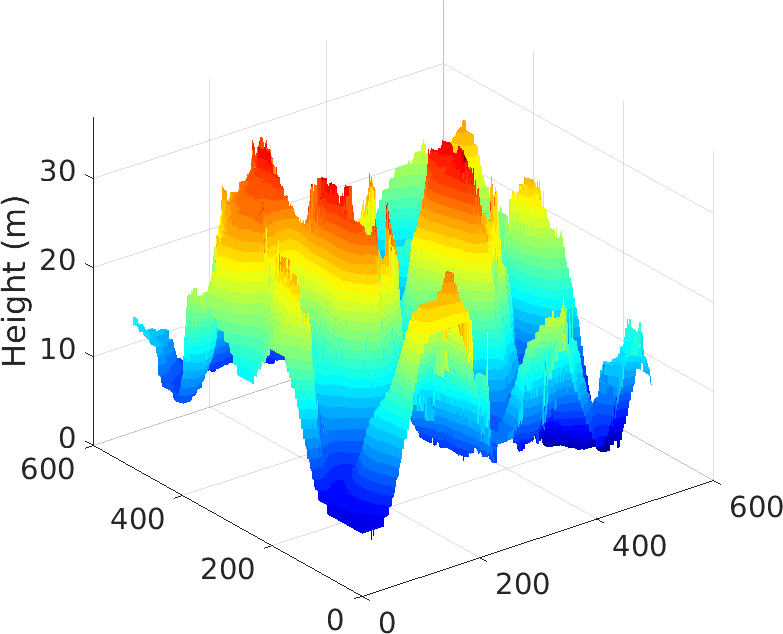}{\fpathTropisingle M2_DTM_3D_HH.png}

\pgfdeclareimage[width=0.2100\textwidth]{CHM_profile_1_HH}{\fpathTropisingle CHM_profile_line1_HH.png}
\pgfdeclareimage[width=0.2100\textwidth]{CHM_profile_170_HH}{\fpathTropisingle CHM_profile_line170_HH.png}
\pgfdeclareimage[width=0.2100\textwidth]{CHM_profile_280_HH}{\fpathTropisingle CHM_profile_line280_HH.png}
\pgfdeclareimage[width=0.2100\textwidth]{DTM_profile_1_HH}{\fpathTropisingle DTM_profile_line1_HH.png}
\pgfdeclareimage[width=0.2100\textwidth]{DTM_profile_170_HH}{\fpathTropisingle DTM_profile_line170_HH.png}
\pgfdeclareimage[width=0.2100\textwidth]{DTM_profile_280_HH}{\fpathTropisingle DTM_profile_line280_HH.png}
\pgfdeclareimage[width=0.2100\textwidth]{SUM_profile_1_HH}{\fpathTropisingle SUM_profile_line1_HH.png}
\pgfdeclareimage[width=0.2100\textwidth]{SUM_profile_170_HH}{\fpathTropisingle SUM_profile_line170_HH.png}
\pgfdeclareimage[width=0.2100\textwidth]{SUM_profile_280_HH}{\fpathTropisingle SUM_profile_line280_HH.png}

\pgfdeclareimage[width=0.26\columnwidth]{CHM_joint_unet_HH}{\fpathTropisingle CHM_jointd_unet_HH.png}
\pgfdeclareimage[width=0.26\columnwidth]{CHM_joint_fcn_HH}{\fpathTropisingle CHM_jointd_fcn_HH.png}
\pgfdeclareimage[width=0.26\columnwidth]{CHM_joint_boss_HH}{\fpathTropisingle CHM_jointd_boss_HH.png}
\pgfdeclareimage[width=0.26\columnwidth]{CHM_joint_m2_HH}{\fpathTropisingle CHM_jointd_m2_HH.png}
\pgfdeclareimage[width=0.26\columnwidth]{DTM_joint_unet_HH}{\fpathTropisingle DTM_jointd_unet_HH.png}
\pgfdeclareimage[width=0.26\columnwidth]{DTM_joint_fcn_HH}{\fpathTropisingle DTM_jointd_fcn_HH.png}
\pgfdeclareimage[width=0.26\columnwidth]{DTM_joint_boss_HH}{\fpathTropisingle DTM_jointd_boss_HH.png}
\pgfdeclareimage[width=0.26\columnwidth]{DTM_joint_m2_HH}{\fpathTropisingle DTM_jointd_m2_HH.png}
\pgfdeclareimage[width=0.26\columnwidth]{SUM_joint_unet_HH}{\fpathTropisingle SUM_jointd_unet_HH.png}
\pgfdeclareimage[width=0.26\columnwidth]{SUM_joint_fcn_HH}{\fpathTropisingle SUM_jointd_fcn_HH.png}
\pgfdeclareimage[width=0.26\columnwidth]{SUM_joint_boss_HH}{\fpathTropisingle SUM_jointd_boss_HH.png}
\pgfdeclareimage[width=0.26\columnwidth]{SUM_joint_m2_HH}{\fpathTropisingle SUM_jointd_m2_HH.png}

\pgfdeclareimage[width=0.199\textwidth]{LiDAR_CHM_2D_BIG}{\fpathTropi LiDAR_CHM_2D.png}
\pgfdeclareimage[width=0.199\textwidth]{LiDAR_DTM_2D_BIG}{\fpathTropi LiDAR_DTM_2D.png}

\pgfdeclareimage[width=0.199\textwidth]{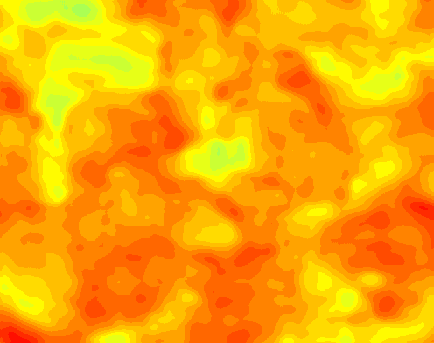}{\fpathTropisingle unet_CHM_2D_HV.png}
\pgfdeclareimage[width=0.199\textwidth]{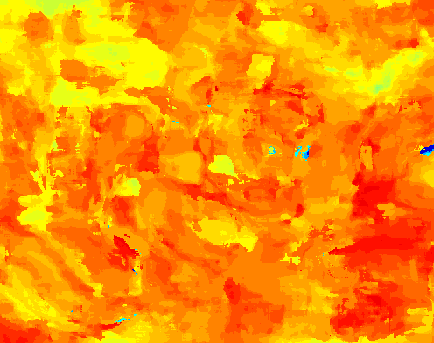}{\fpathTropisingle FCN_CHM_2D_HV.png}
\pgfdeclareimage[width=0.199\textwidth]{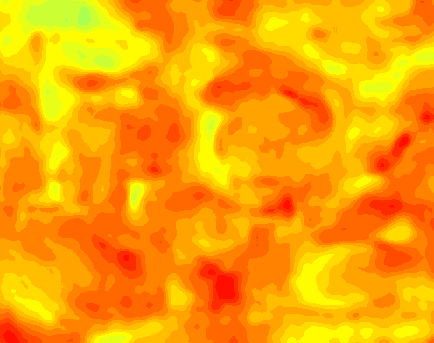}{\fpathTropisingle BOSS_CHM_2D_HV.png}
\pgfdeclareimage[width=0.199\textwidth]{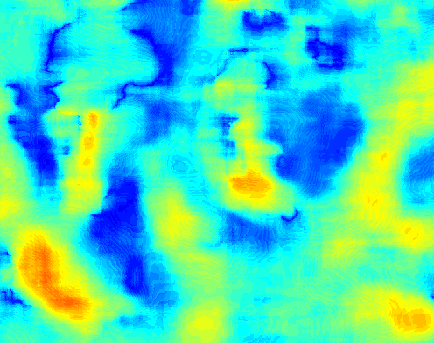}{\fpathTropisingle M2_CHM_2D_HV.png}
\pgfdeclareimage[width=0.199\textwidth]{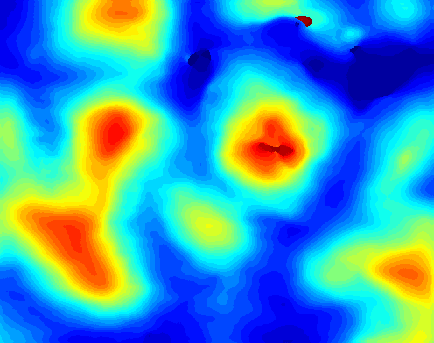}{\fpathTropisingle unet_DTM_2D_HV.png}
\pgfdeclareimage[width=0.199\textwidth]{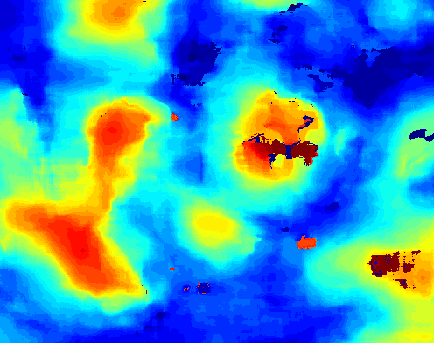}{\fpathTropisingle FCN_DTM_2D_HV.png}
\pgfdeclareimage[width=0.199\textwidth]{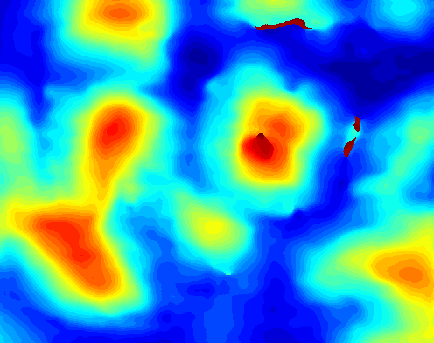}{\fpathTropisingle BOSS_DTM_2D_HV.png}
\pgfdeclareimage[width=0.199\textwidth]{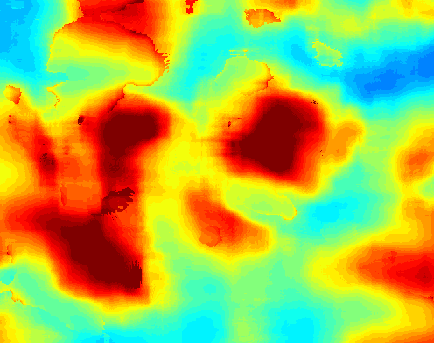}{\fpathTropisingle M2_DTM_2D_HV.png}

\pgfdeclareimage[width=0.199\textwidth]{LiDAR_SUM_2D_BIG}{\fpathTropi LiDAR_SUM_2D.png}
\pgfdeclareimage[width=0.199\textwidth]{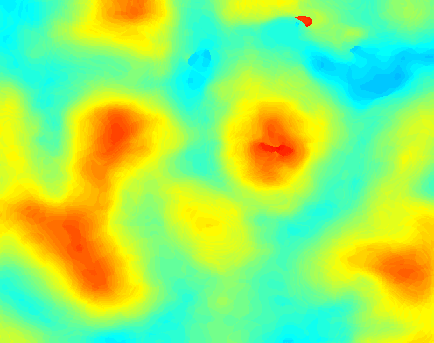}{\fpathTropisingle unet_SUM_2D_HV.png}
\pgfdeclareimage[width=0.199\textwidth]{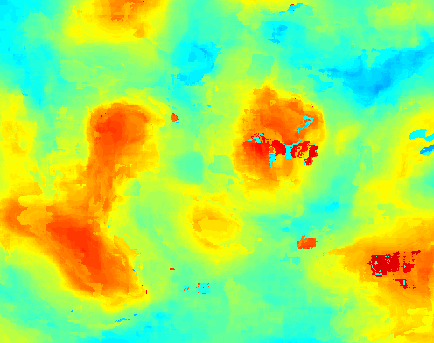}{\fpathTropisingle FCN_SUM_2D_HV.png}
\pgfdeclareimage[width=0.199\textwidth]{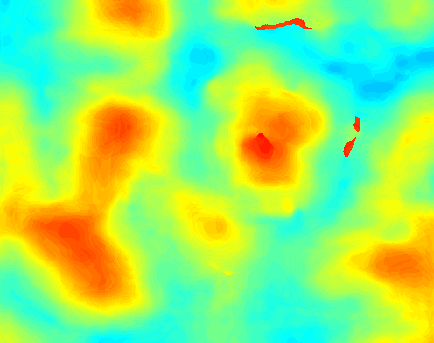}{\fpathTropisingle BOSS_SUM_2D_HV.png}
\pgfdeclareimage[width=0.199\textwidth]{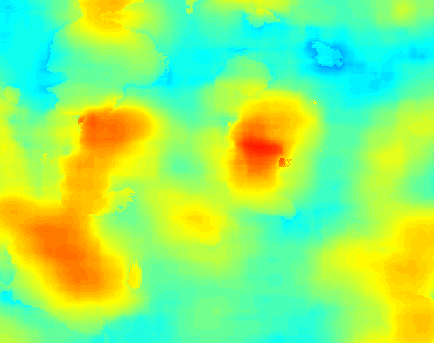}{\fpathTropisingle M2_SUM_2D_HV.png}

\pgfdeclareimage[width=0.199\textwidth]{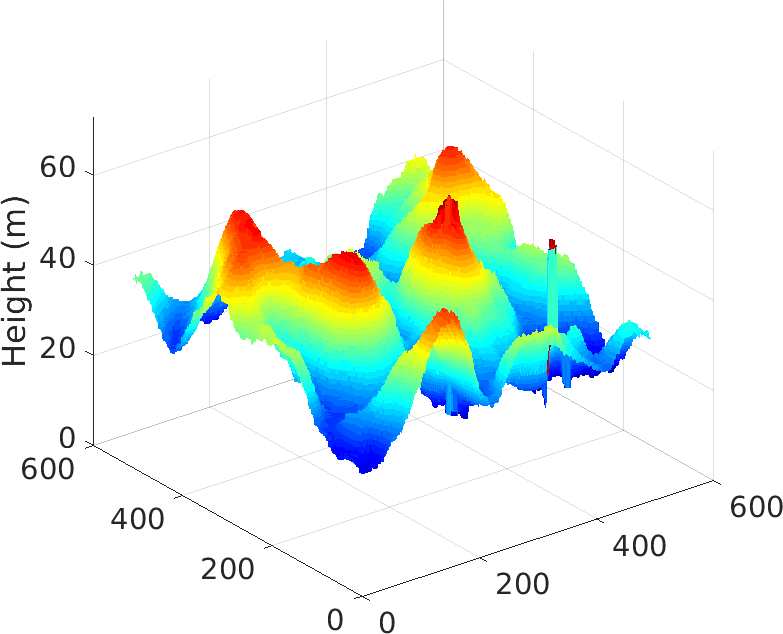}{\fpathTropisingle unet_SUM_3D_HV.png}
\pgfdeclareimage[width=0.199\textwidth]{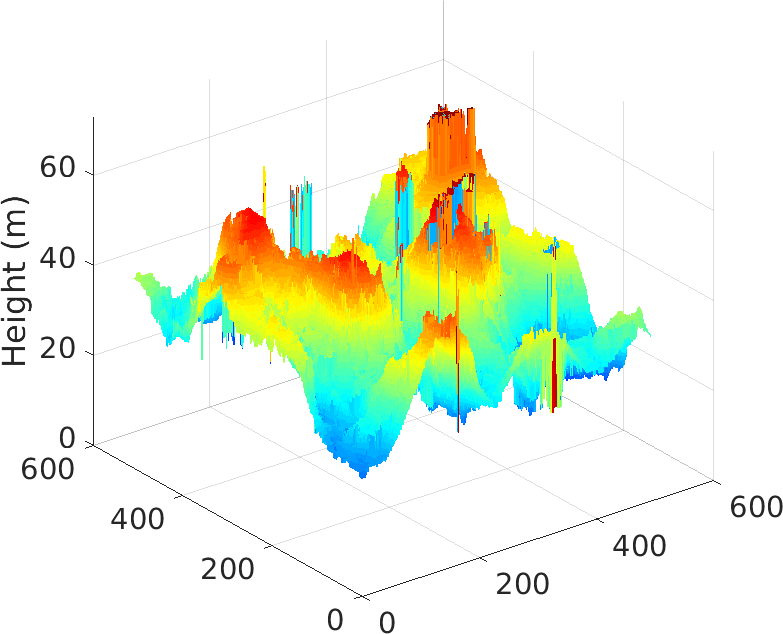}{\fpathTropisingle FCN_SUM_3D_HV.png}
\pgfdeclareimage[width=0.199\textwidth]{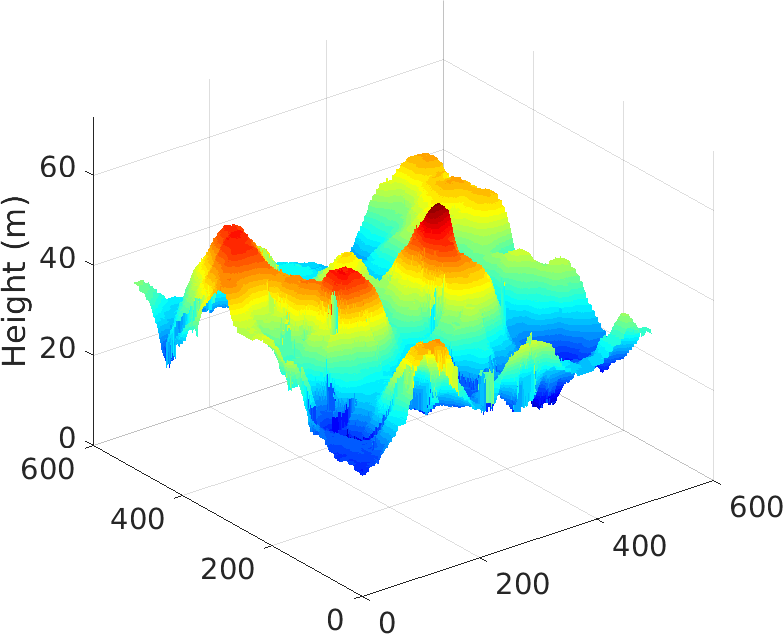}{\fpathTropisingle M2_SUM_3D_HV.png}
\pgfdeclareimage[width=0.199\textwidth]{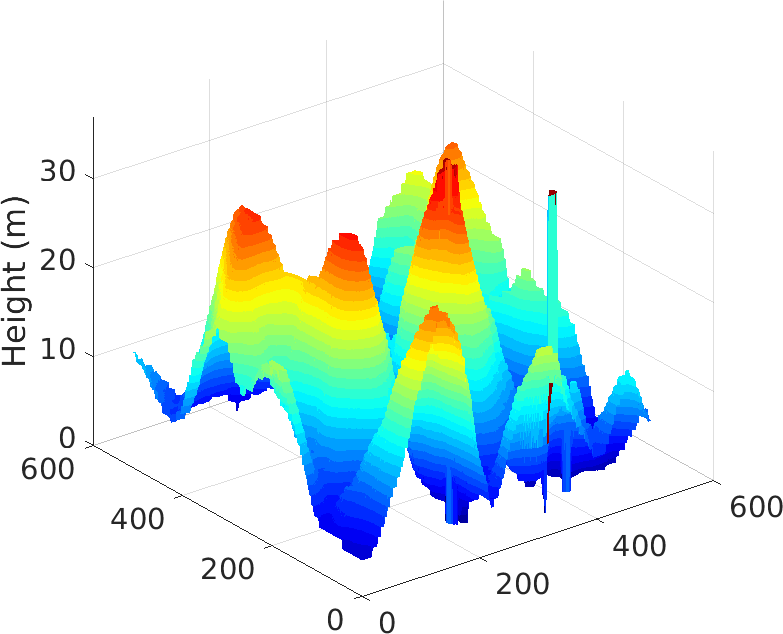}{\fpathTropisingle unet_DTM_3D_HV.png}
\pgfdeclareimage[width=0.199\textwidth]{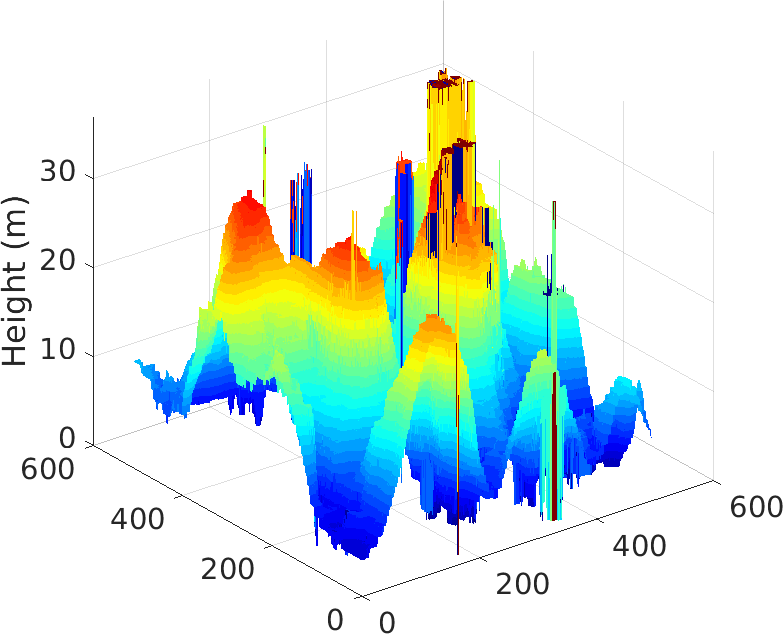}{\fpathTropisingle FCN_DTM_3D_HV.png}
\pgfdeclareimage[width=0.199\textwidth]{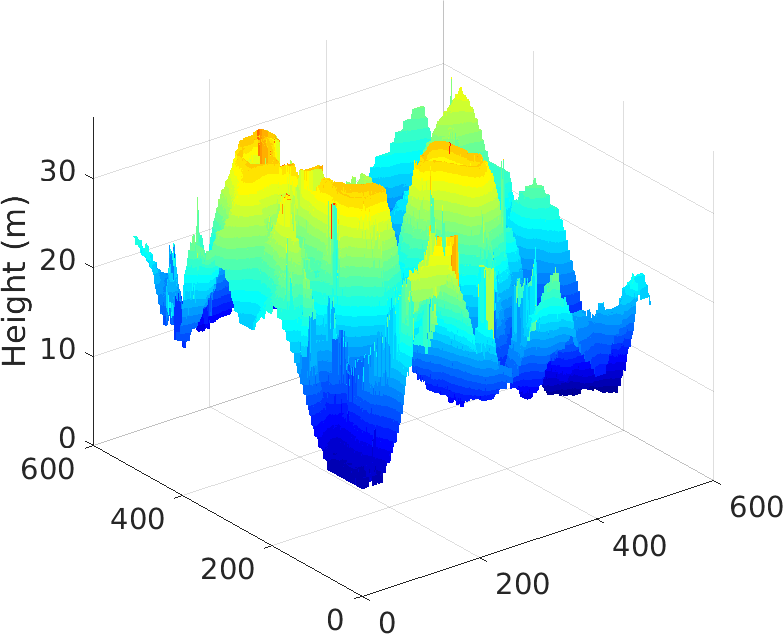}{\fpathTropisingle M2_DTM_3D_HV.png}

\pgfdeclareimage[width=0.2100\textwidth]{CHM_profile_1_HV}{\fpathTropisingle CHM_profile_line1_HV.png}
\pgfdeclareimage[width=0.2100\textwidth]{CHM_profile_170_HV}{\fpathTropisingle CHM_profile_line170_HV.png}
\pgfdeclareimage[width=0.2100\textwidth]{CHM_profile_280_HV}{\fpathTropisingle CHM_profile_line280_HV.png}
\pgfdeclareimage[width=0.2100\textwidth]{DTM_profile_1_HV}{\fpathTropisingle DTM_profile_line1_HV.png}
\pgfdeclareimage[width=0.2100\textwidth]{DTM_profile_170_HV}{\fpathTropisingle DTM_profile_line170_HV.png}
\pgfdeclareimage[width=0.2100\textwidth]{DTM_profile_280_HV}{\fpathTropisingle DTM_profile_line280_HV.png}
\pgfdeclareimage[width=0.2100\textwidth]{SUM_profile_1_HV}{\fpathTropisingle SUM_profile_line1_HV.png}
\pgfdeclareimage[width=0.2100\textwidth]{SUM_profile_170_HV}{\fpathTropisingle SUM_profile_line170_HV.png}
\pgfdeclareimage[width=0.2100\textwidth]{SUM_profile_280_HV}{\fpathTropisingle SUM_profile_line280_HV.png}

\pgfdeclareimage[width=0.26\columnwidth]{CHM_joint_unet_HV}{\fpathTropisingle CHM_jointd_unet_HV.png}
\pgfdeclareimage[width=0.26\columnwidth]{CHM_joint_fcn_HV}{\fpathTropisingle CHM_jointd_fcn_HV.png}
\pgfdeclareimage[width=0.26\columnwidth]{CHM_joint_boss_HV}{\fpathTropisingle CHM_jointd_boss_HV.png}
\pgfdeclareimage[width=0.26\columnwidth]{CHM_joint_m2_HV}{\fpathTropisingle CHM_jointd_m2_HV.png}

\pgfdeclareimage[width=0.26\columnwidth]{DTM_joint_unet_HV}{\fpathTropisingle DTM_jointd_unet_HV.png}
\pgfdeclareimage[width=0.26\columnwidth]{DTM_joint_fcn_HV}{\fpathTropisingle DTM_jointd_fcn_HV.png}
\pgfdeclareimage[width=0.26\columnwidth]{DTM_joint_boss_HV}{\fpathTropisingle DTM_jointd_boss_HV.png}
\pgfdeclareimage[width=0.26\columnwidth]{DTM_joint_m2_HV}{\fpathTropisingle DTM_jointd_m2_HV.png}

\pgfdeclareimage[width=0.26\columnwidth]{SUM_joint_unet_HV}{\fpathTropisingle SUM_jointd_unet_HV.png}
\pgfdeclareimage[width=0.26\columnwidth]{SUM_joint_fcn_HV}{\fpathTropisingle SUM_jointd_fcn_HV.png}
\pgfdeclareimage[width=0.26\columnwidth]{SUM_joint_boss_HV}{\fpathTropisingle SUM_jointd_boss_HV.png}
\pgfdeclareimage[width=0.26\columnwidth]{SUM_joint_m2_HV}{\fpathTropisingle SUM_jointd_m2_HV.png}

\pgfdeclareimage[width=0.199\textwidth]{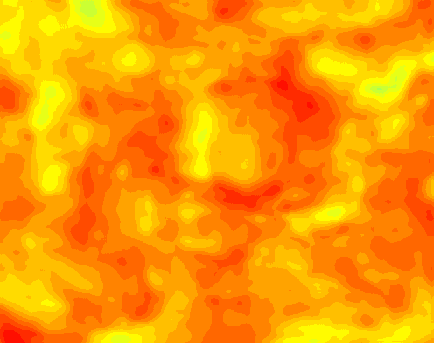}{\fpathTropisingle unet_CHM_2D_VV.png}
\pgfdeclareimage[width=0.199\textwidth]{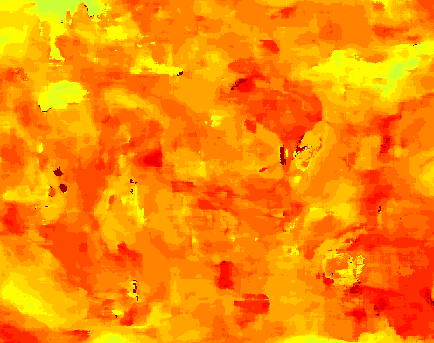}{\fpathTropisingle FCN_CHM_2D_VV.png}
\pgfdeclareimage[width=0.199\textwidth]{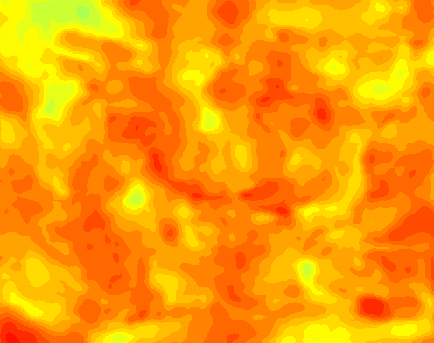}{\fpathTropisingle BOSS_CHM_2D_VV.png}
\pgfdeclareimage[width=0.199\textwidth]{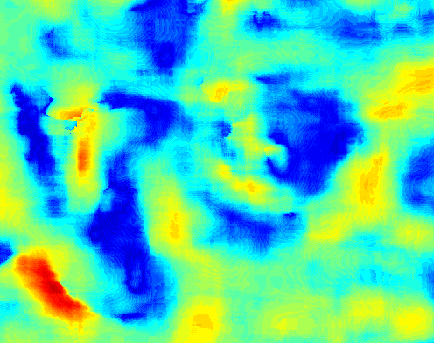}{\fpathTropisingle M2_CHM_2D_VV.png}

\pgfdeclareimage[width=0.199\textwidth]{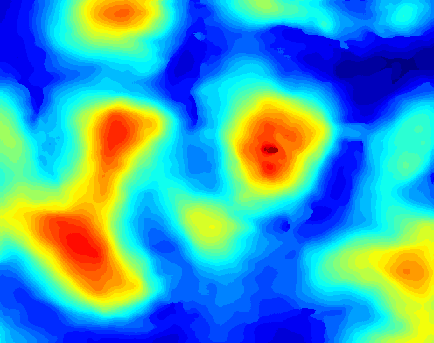}{\fpathTropisingle unet_DTM_2D_VV.png}
\pgfdeclareimage[width=0.199\textwidth]{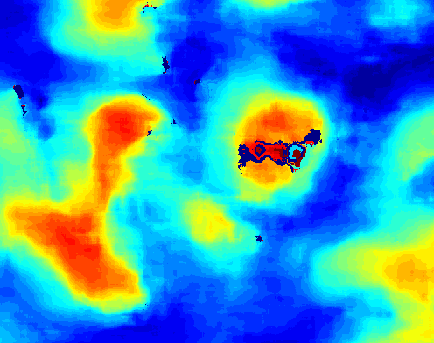}{\fpathTropisingle FCN_DTM_2D_VV.png}
\pgfdeclareimage[width=0.199\textwidth]{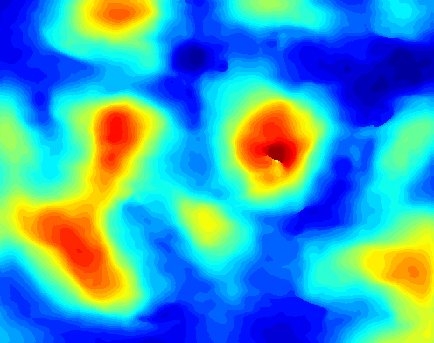}{\fpathTropisingle BOSS_DTM_2D_VV.png}
\pgfdeclareimage[width=0.199\textwidth]{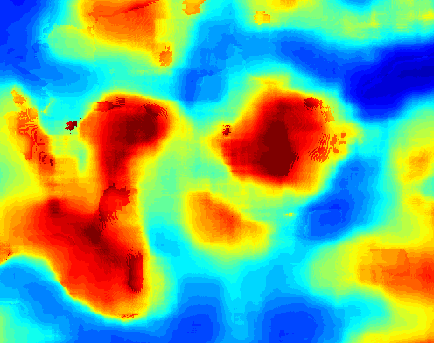}{\fpathTropisingle BOSS_DTM_2D_VV.png}

\pgfdeclareimage[width=0.199\textwidth]{LiDAR_SUM_2D_BIG}{\fpathTropi LiDAR_SUM_2D.png}
\pgfdeclareimage[width=0.199\textwidth]{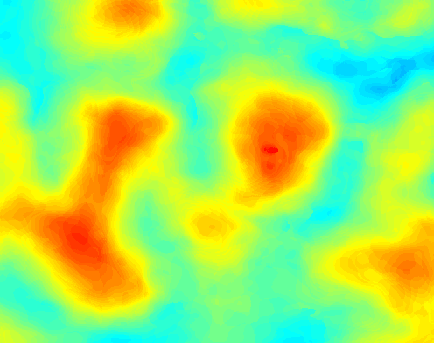}{\fpathTropisingle unet_SUM_2D_VV.png}
\pgfdeclareimage[width=0.199\textwidth]{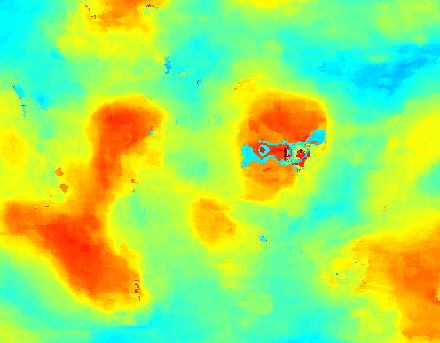}{\fpathTropisingle FCN_SUM_2D_VV.png}
\pgfdeclareimage[width=0.199\textwidth]{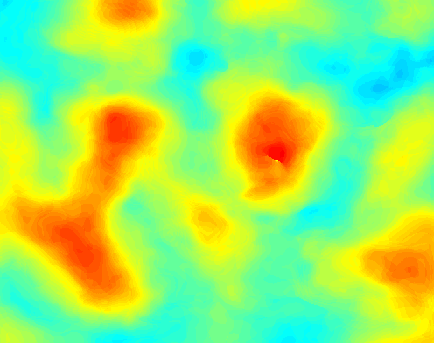}{\fpathTropisingle BOSS_SUM_2D_VV.png}
\pgfdeclareimage[width=0.199\textwidth]{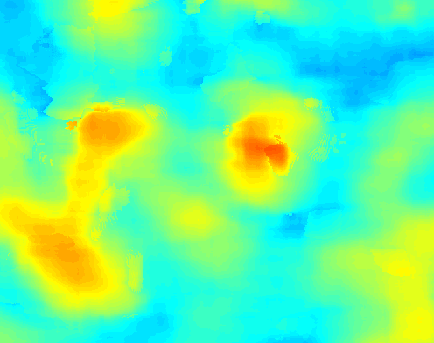}{\fpathTropisingle M2_SUM_2D_VV.png}

\pgfdeclareimage[width=0.199\textwidth]{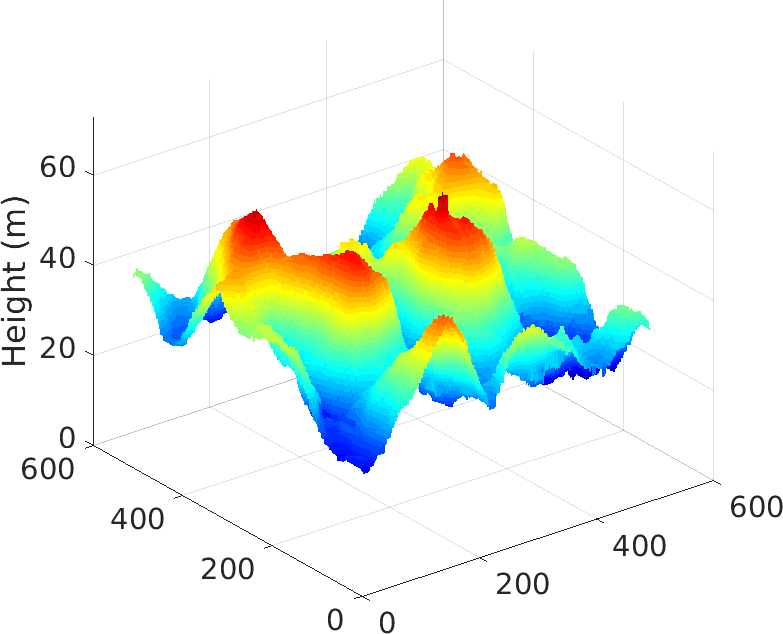}{\fpathTropisingle unet_SUM_3D_VV.png}
\pgfdeclareimage[width=0.199\textwidth]{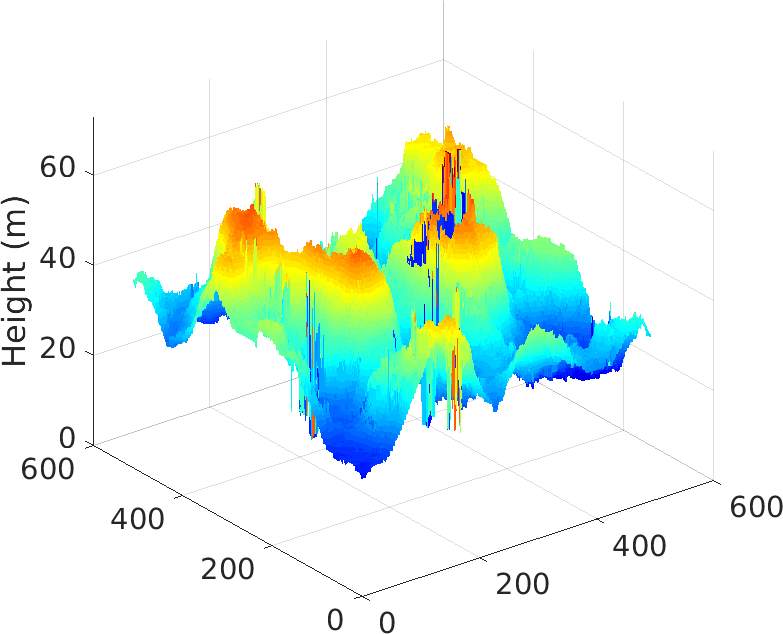}{\fpathTropisingle FCN_SUM_3D_VV.png}
\pgfdeclareimage[width=0.199\textwidth]{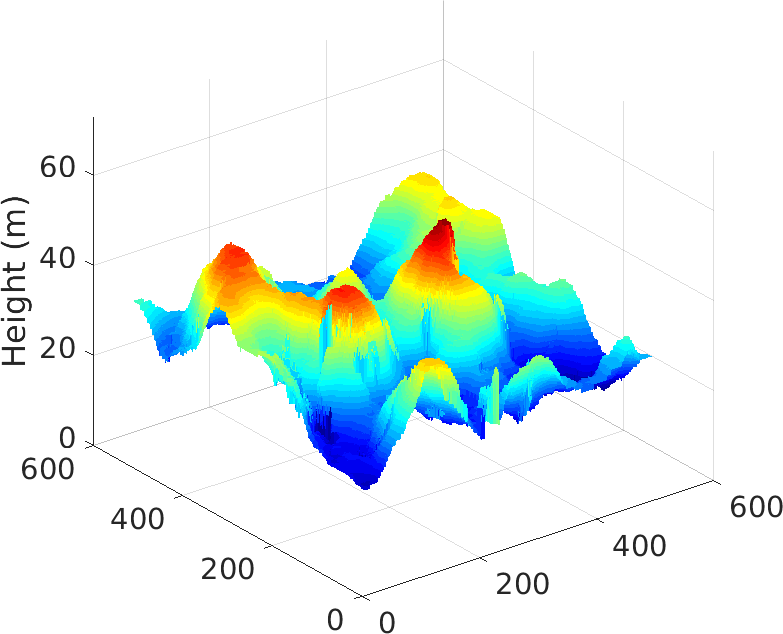}{\fpathTropisingle M2_SUM_3D_VV.png}

\pgfdeclareimage[width=0.199\textwidth]{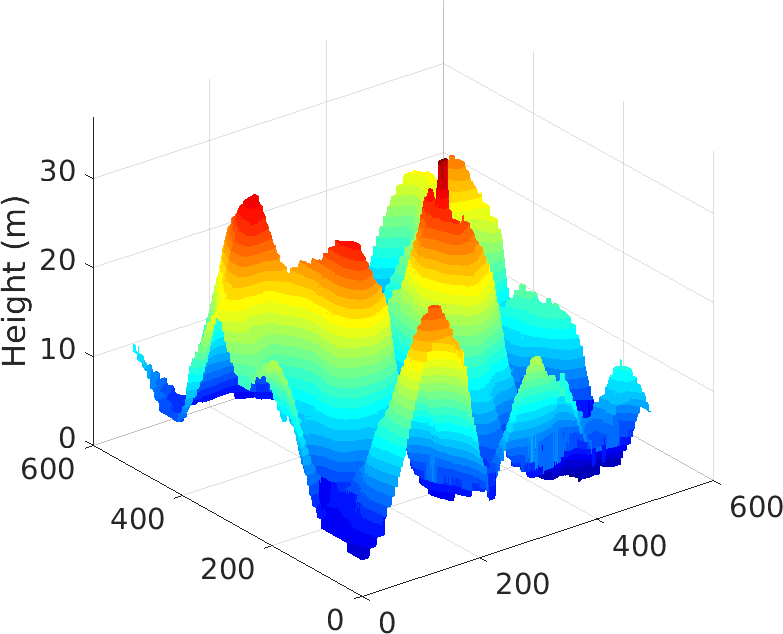}{\fpathTropisingle unet_DTM_3D_VV.png}
\pgfdeclareimage[width=0.199\textwidth]{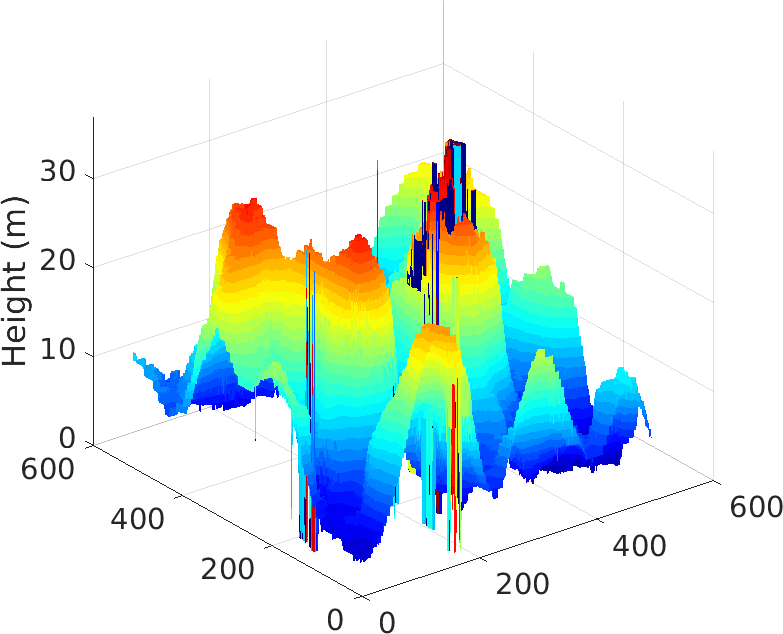}{\fpathTropisingle FCN_DTM_3D_VV.png}
\pgfdeclareimage[width=0.199\textwidth]{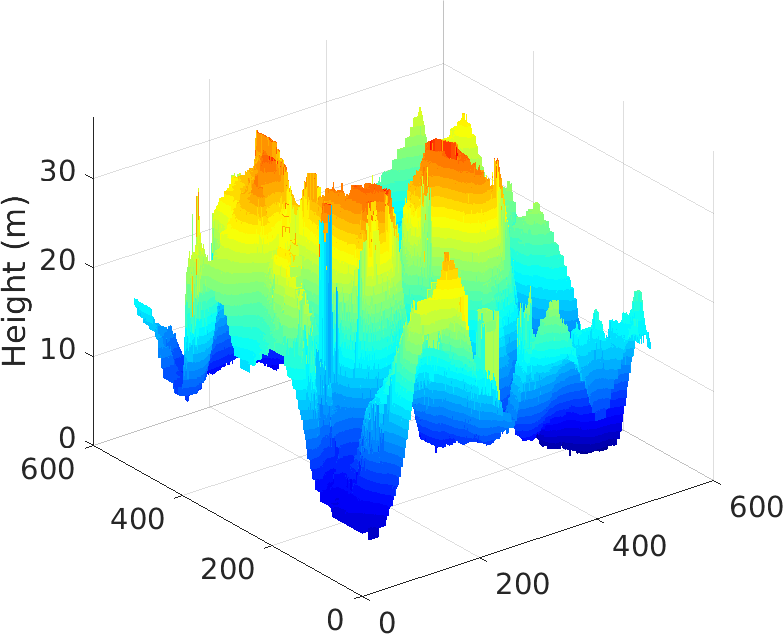}{\fpathTropisingle M2_DTM_3D_VV.png}

\pgfdeclareimage[width=0.2100\textwidth]{CHM_profile_1_VV}{\fpathTropisingle CHM_profile_line1_VV.png}
\pgfdeclareimage[width=0.2100\textwidth]{CHM_profile_170_VV}{\fpathTropisingle CHM_profile_line170_VV.png}
\pgfdeclareimage[width=0.2100\textwidth]{CHM_profile_280_VV}{\fpathTropisingle CHM_profile_line280_VV.png}
\pgfdeclareimage[width=0.2100\textwidth]{DTM_profile_1_VV}{\fpathTropisingle DTM_profile_line1_VV.png}
\pgfdeclareimage[width=0.2100\textwidth]{DTM_profile_170_VV}{\fpathTropisingle DTM_profile_line170_VV.png}
\pgfdeclareimage[width=0.2100\textwidth]{DTM_profile_280_VV}{\fpathTropisingle DTM_profile_line280_VV.png}
\pgfdeclareimage[width=0.2100\textwidth]{SUM_profile_1_VV}{\fpathTropisingle SUM_profile_line1_VV.png}
\pgfdeclareimage[width=0.2100\textwidth]{SUM_profile_170_VV}{\fpathTropisingle SUM_profile_line170_VV.png}
\pgfdeclareimage[width=0.2100\textwidth]{SUM_profile_280_VV}{\fpathTropisingle SUM_profile_line280_VV.png}

\pgfdeclareimage[width=0.26\columnwidth]{CHM_joint_unet_VV}{\fpathTropisingle CHM_jointd_unet_VV.png}
\pgfdeclareimage[width=0.26\columnwidth]{CHM_joint_fcn_VV}{\fpathTropisingle CHM_jointd_fcn_VV.png}
\pgfdeclareimage[width=0.26\columnwidth]{CHM_joint_boss_VV}{\fpathTropisingle CHM_jointd_boss_VV.png}
\pgfdeclareimage[width=0.26\columnwidth]{CHM_joint_m2_VV}{\fpathTropisingle CHM_jointd_m2_VV.png}
\pgfdeclareimage[width=0.26\columnwidth]{DTM_joint_unet_VV}{\fpathTropisingle DTM_jointd_unet_VV.png}
\pgfdeclareimage[width=0.26\columnwidth]{DTM_joint_fcn_VV}{\fpathTropisingle DTM_jointd_fcn_VV.png}
\pgfdeclareimage[width=0.26\columnwidth]{DTM_joint_boss_VV}{\fpathTropisingle DTM_jointd_boss_VV.png}
\pgfdeclareimage[width=0.26\columnwidth]{DTM_joint_m2_VV}{\fpathTropisingle DTM_jointd_m2_VV.png}
\pgfdeclareimage[width=0.26\columnwidth]{SUM_joint_unet_VV}{\fpathTropisingle SUM_jointd_unet_VV.png}
\pgfdeclareimage[width=0.26\columnwidth]{SUM_joint_fcn_VV}{\fpathTropisingle SUM_jointd_fcn_VV.png}
\pgfdeclareimage[width=0.26\columnwidth]{SUM_joint_boss_VV}{\fpathTropisingle SUM_jointd_boss_VV.png}
\pgfdeclareimage[width=0.26\columnwidth]{SUM_joint_m2_VV}{\fpathTropisingle SUM_jointd_m2_VV.png}

\pgfdeclareimage[width=0.199\textwidth]{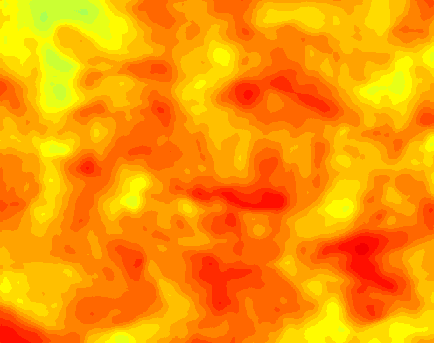}{\fpathTropidual unet_CHM_2D_HHHV.png}
\pgfdeclareimage[width=0.199\textwidth]{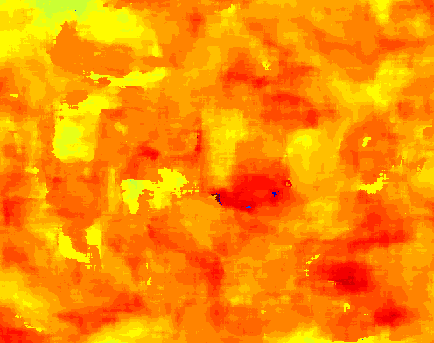}{\fpathTropidual FCN_CHM_2D_HHHV.png}
\pgfdeclareimage[width=0.199\textwidth]{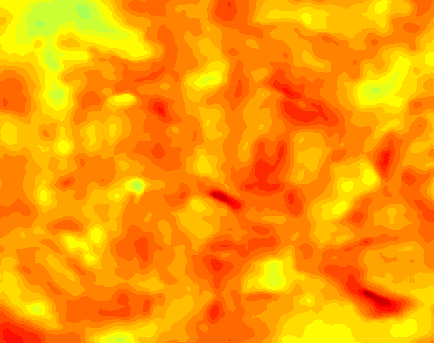}{\fpathTropidual BOSS_CHM_2D_HHHV.png}
\pgfdeclareimage[width=0.199\textwidth]{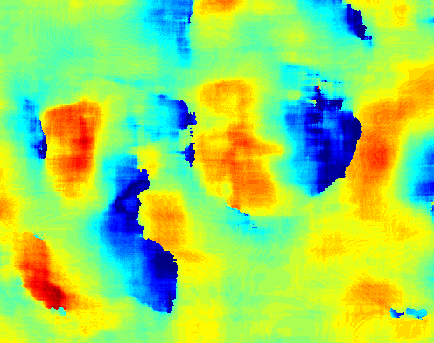}{\fpathTropidual SKP_CHM_2D_HHHV.png}

\pgfdeclareimage[width=0.199\textwidth]{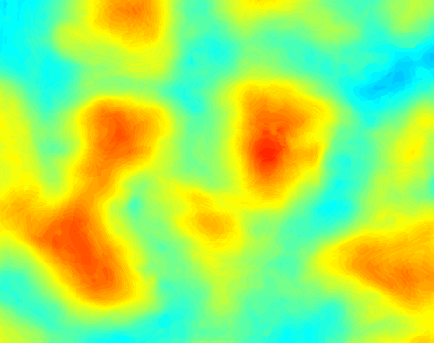}{\fpathTropidual unet_SUM_2D_HHHV.png}
\pgfdeclareimage[width=0.199\textwidth]{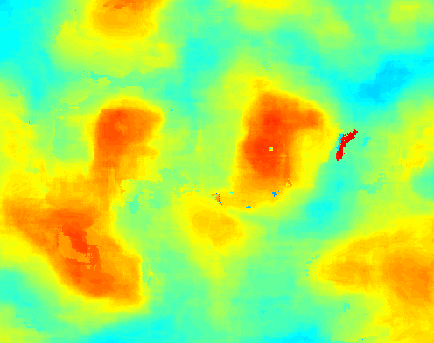}{\fpathTropidual FCN_SUM_2D_HHHV.png}
\pgfdeclareimage[width=0.199\textwidth]{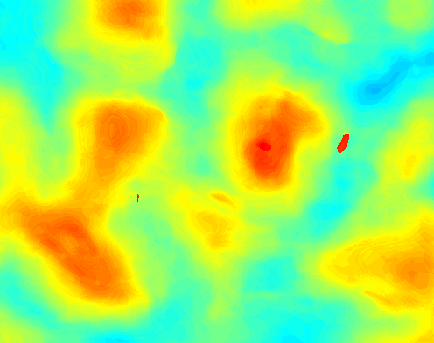}{\fpathTropidual BOSS_SUM_2D_HHHV.png}
\pgfdeclareimage[width=0.199\textwidth]{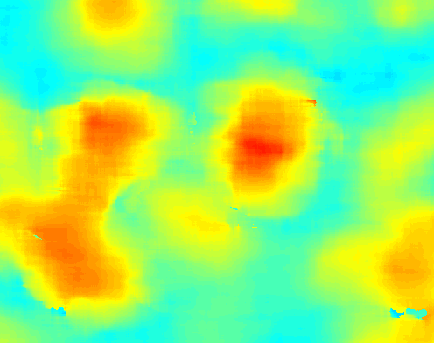}{\fpathTropidual SKP_SUM_2D_HHHV.png}

\pgfdeclareimage[width=0.199\textwidth]{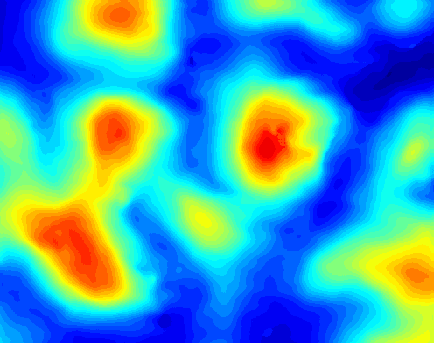}{\fpathTropidual unet_DTM_2D_HHHV.png}
\pgfdeclareimage[width=0.199\textwidth]{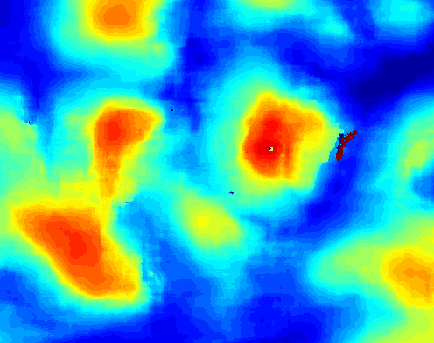}{\fpathTropidual FCN_DTM_2D_HHHV.png}
\pgfdeclareimage[width=0.199\textwidth]{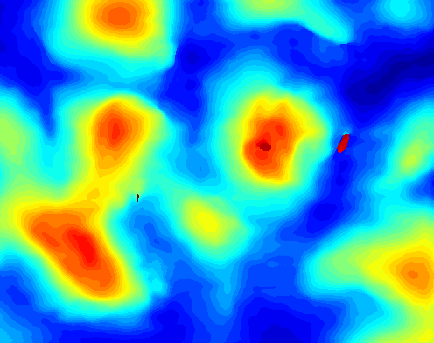}{\fpathTropidual BOSS_DTM_2D_HHHV.png}
\pgfdeclareimage[width=0.199\textwidth]{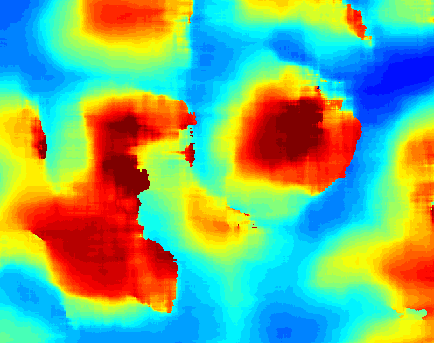}{\fpathTropidual SKP_DTM_2D_HHHV.png}

\pgfdeclareimage[width=0.199\textwidth]{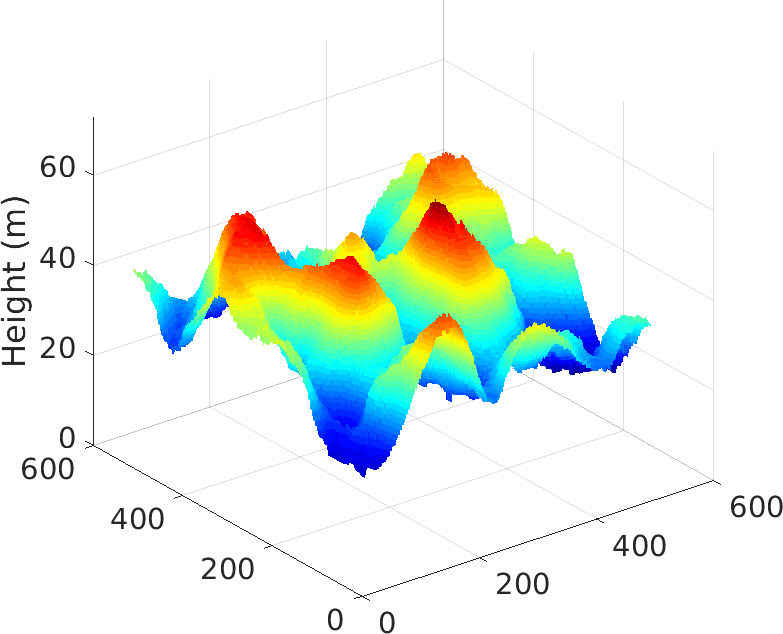}{\fpathTropidual unet_SUM_3D_HHHV.png}
\pgfdeclareimage[width=0.199\textwidth]{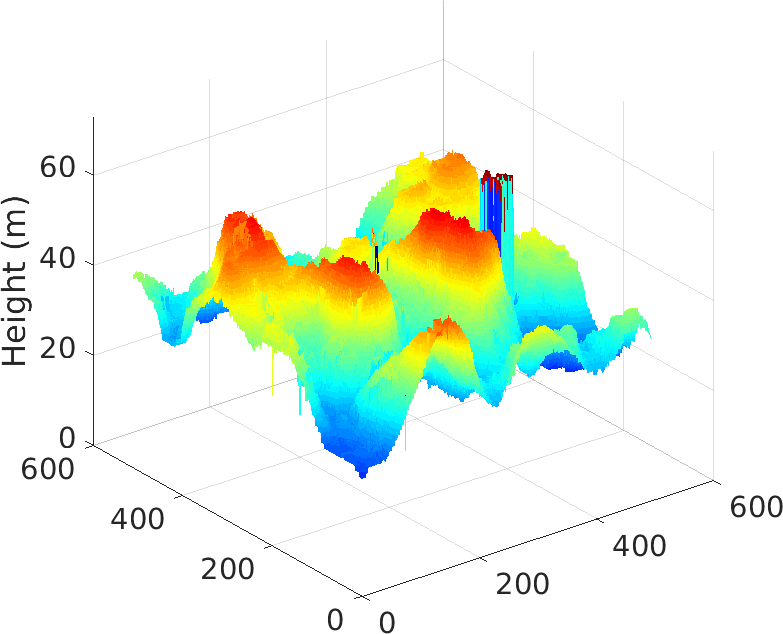}{\fpathTropidual FCN_SUM_3D_HHHV.png}
\pgfdeclareimage[width=0.199\textwidth]{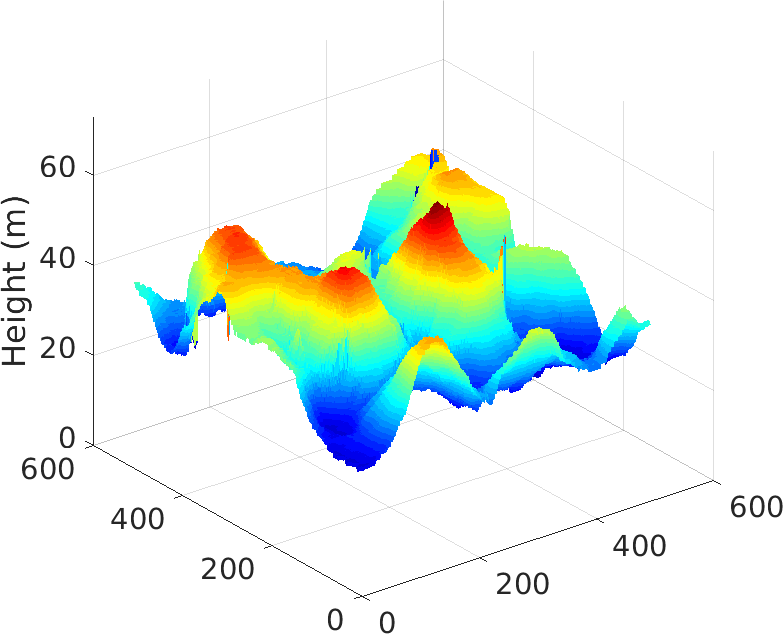}{\fpathTropidual SKP_SUM_3D_HHHV.png}

\pgfdeclareimage[width=0.199\textwidth]{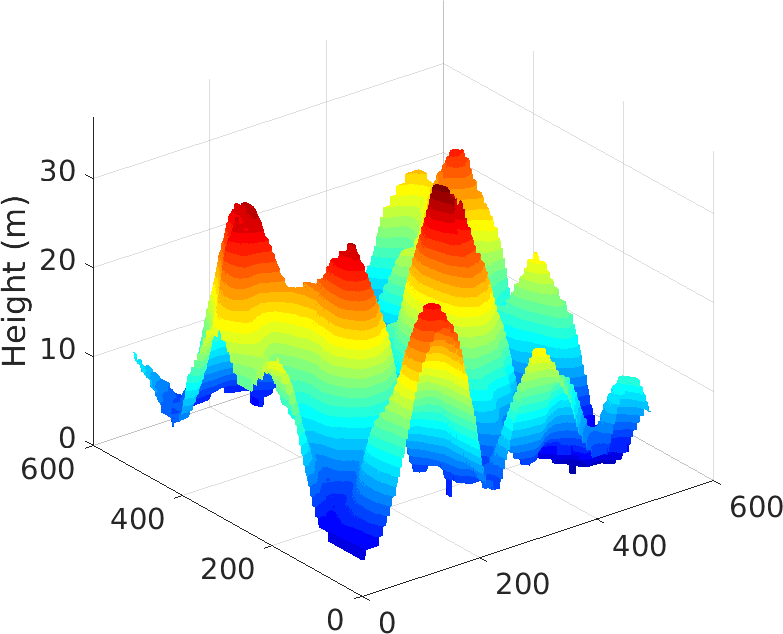}{\fpathTropidual unet_DTM_3D_HHHV.png}
\pgfdeclareimage[width=0.199\textwidth]{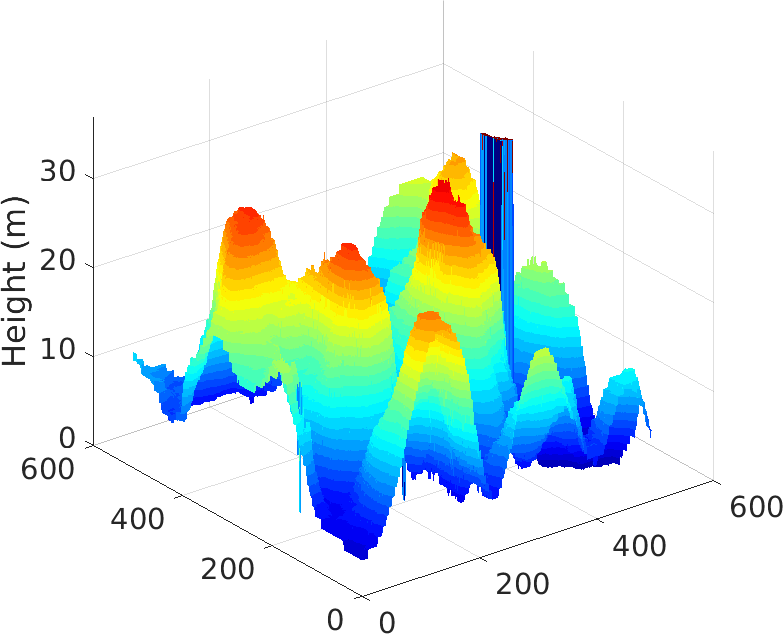}{\fpathTropidual FCN_DTM_3D_HHHV.png}
\pgfdeclareimage[width=0.199\textwidth]{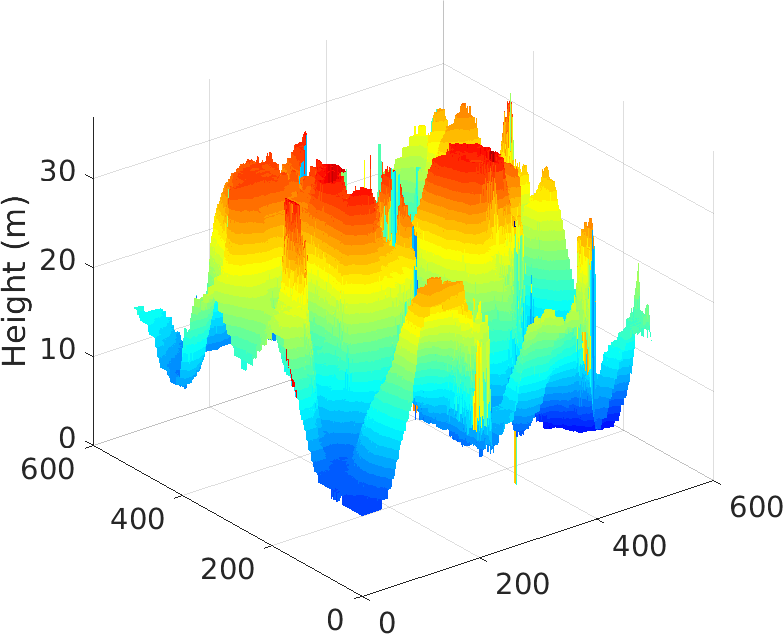}{\fpathTropidual SKP_DTM_3D_HHHV.png}

\pgfdeclareimage[width=0.2100\textwidth]{CHM_profile_1_HHHV}{\fpathTropidual CHM_profile_line1_HHHV.png}
\pgfdeclareimage[width=0.2100\textwidth]{CHM_profile_170_HHHV}{\fpathTropidual CHM_profile_line170_HHHV.png}
\pgfdeclareimage[width=0.2100\textwidth]{CHM_profile_280_HHHV}{\fpathTropidual CHM_profile_line280_HHHV.png}
\pgfdeclareimage[width=0.2100\textwidth]{DTM_profile_1_HHHV}{\fpathTropidual DTM_profile_line1_HHHV.png}
\pgfdeclareimage[width=0.2100\textwidth]{DTM_profile_170_HHHV}{\fpathTropidual DTM_profile_line170_HHHV.png}
\pgfdeclareimage[width=0.2100\textwidth]{DTM_profile_280_HHHV}{\fpathTropidual DTM_profile_line280_HHHV.png}
\pgfdeclareimage[width=0.2100\textwidth]{SUM_profile_1_HHHV}{\fpathTropidual SUM_profile_line1_HHHV.png}
\pgfdeclareimage[width=0.2100\textwidth]{SUM_profile_170_HHHV}{\fpathTropidual SUM_profile_line170_HHHV.png}
\pgfdeclareimage[width=0.2100\textwidth]{SUM_profile_280_HHHV}{\fpathTropidual SUM_profile_line280_HHHV.png}

\pgfdeclareimage[width=0.26\columnwidth]{CHM_joint_unet_HHHV}{\fpathTropidual CHM_jointd_unet_HHHV.png}
\pgfdeclareimage[width=0.26\columnwidth]{CHM_joint_fcn_HHHV}{\fpathTropidual CHM_jointd_fcn_HHHV.png}
\pgfdeclareimage[width=0.26\columnwidth]{CHM_joint_boss_HHHV}{\fpathTropidual CHM_jointd_boss_HHHV.png}
\pgfdeclareimage[width=0.26\columnwidth]{CHM_joint_skp_HHHV}{\fpathTropidual CHM_jointd_skp_HHHV.png}
\pgfdeclareimage[width=0.26\columnwidth]{DTM_joint_unet_HHHV}{\fpathTropidual DTM_jointd_unet_HHHV.png}
\pgfdeclareimage[width=0.26\columnwidth]{DTM_joint_fcn_HHHV}{\fpathTropidual DTM_jointd_fcn_HHHV.png}
\pgfdeclareimage[width=0.26\columnwidth]{DTM_joint_boss_HHHV}{\fpathTropidual DTM_jointd_boss_HHHV.png}
\pgfdeclareimage[width=0.26\columnwidth]{DTM_joint_skp_HHHV}{\fpathTropidual DTM_jointd_skp_HHHV.png}
\pgfdeclareimage[width=0.26\columnwidth]{SUM_joint_unet_HHHV}{\fpathTropidual SUM_jointd_unet_HHHV.png}
\pgfdeclareimage[width=0.26\columnwidth]{SUM_joint_fcn_HHHV}{\fpathTropidual SUM_jointd_fcn_HHHV.png}
\pgfdeclareimage[width=0.26\columnwidth]{SUM_joint_boss_HHHV}{\fpathTropidual SUM_jointd_boss_HHHV.png}
\pgfdeclareimage[width=0.26\columnwidth]{SUM_joint_skp_HHHV}{\fpathTropidual SUM_jointd_skp_HHHV.png}

\pgfdeclareimage[width=0.199\textwidth]{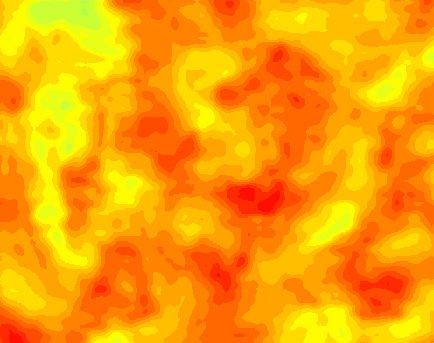}{\fpathTropidual unet_CHM_2D_HHVV.png}
\pgfdeclareimage[width=0.199\textwidth]{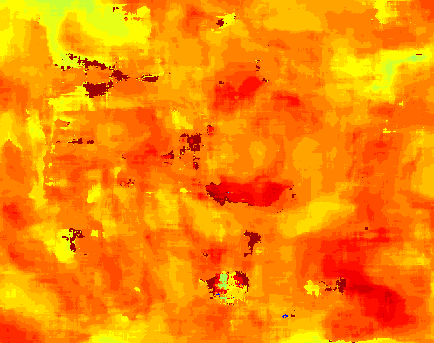}{\fpathTropidual FCN_CHM_2D_HHVV.png}
\pgfdeclareimage[width=0.199\textwidth]{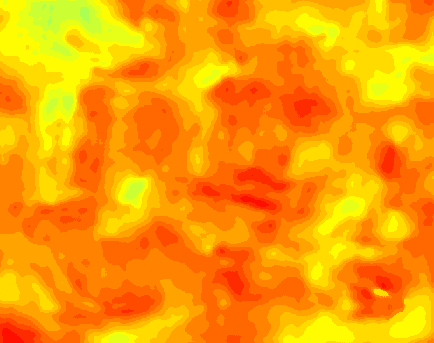}{\fpathTropidual BOSS_CHM_2D_HHVV.png}
\pgfdeclareimage[width=0.199\textwidth]{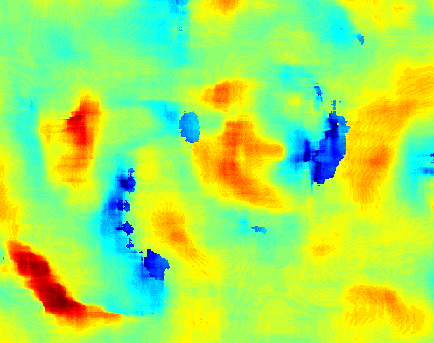}{\fpathTropidual SKP_CHM_2D_HHVV.png}

\pgfdeclareimage[width=0.199\textwidth]{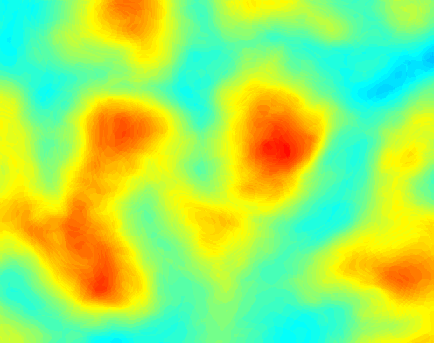}{\fpathTropidual unet_SUM_2D_HHVV.png}
\pgfdeclareimage[width=0.199\textwidth]{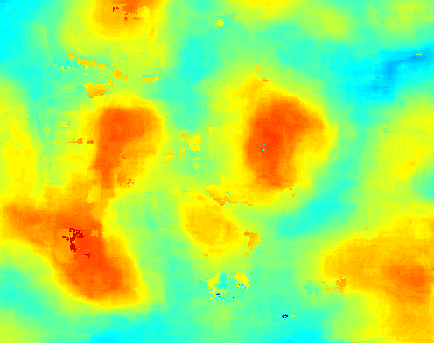}{\fpathTropidual FCN_SUM_2D_HHVV.png}
\pgfdeclareimage[width=0.199\textwidth]{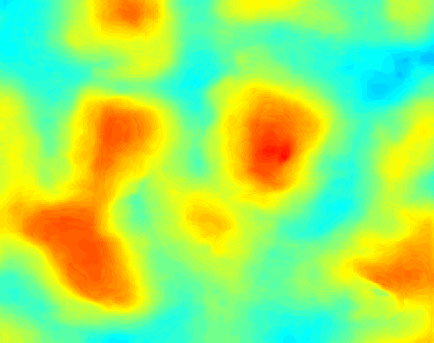}{\fpathTropidual BOSS_SUM_2D_HHVV.png}
\pgfdeclareimage[width=0.199\textwidth]{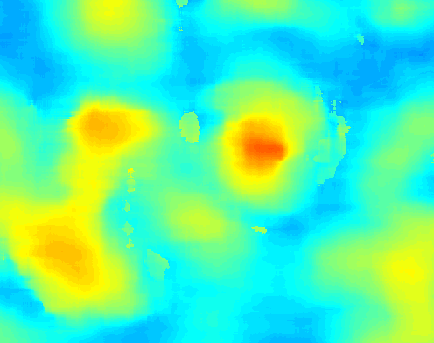}{\fpathTropidual SKP_SUM_2D_HHVV.png}

\pgfdeclareimage[width=0.199\textwidth]{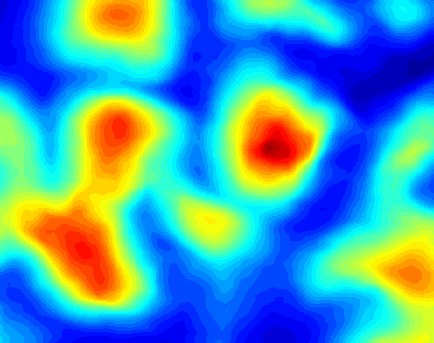}{\fpathTropidual unet_DTM_2D_HHVV.png}
\pgfdeclareimage[width=0.199\textwidth]{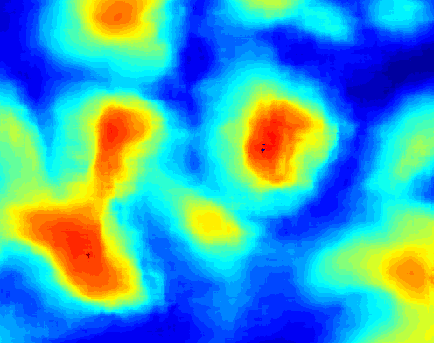}{\fpathTropidual FCN_DTM_2D_HHVV.png}
\pgfdeclareimage[width=0.199\textwidth]{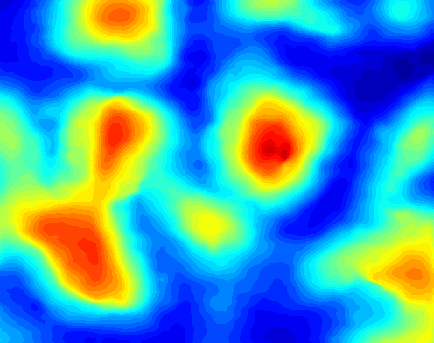}{\fpathTropidual BOSS_DTM_2D_HHVV.png}
\pgfdeclareimage[width=0.199\textwidth]{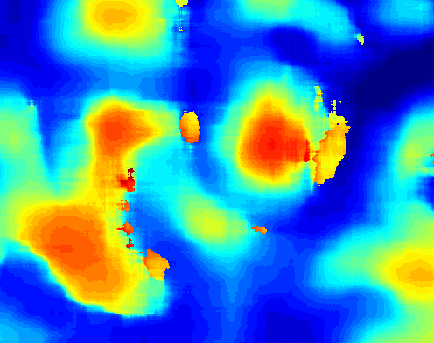}{\fpathTropidual SKP_DTM_2D_HHVV.png}

\pgfdeclareimage[width=0.199\textwidth]{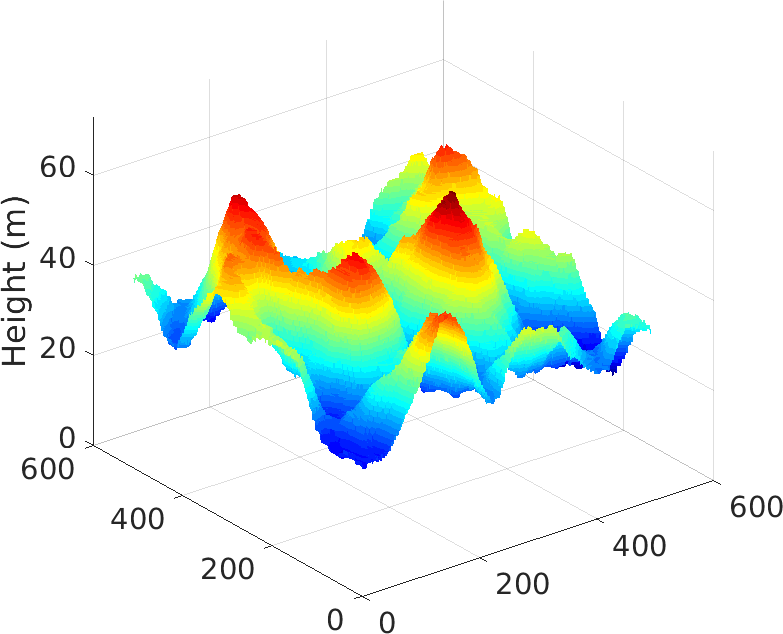}{\fpathTropidual unet_SUM_3D_HHVV.png}
\pgfdeclareimage[width=0.199\textwidth]{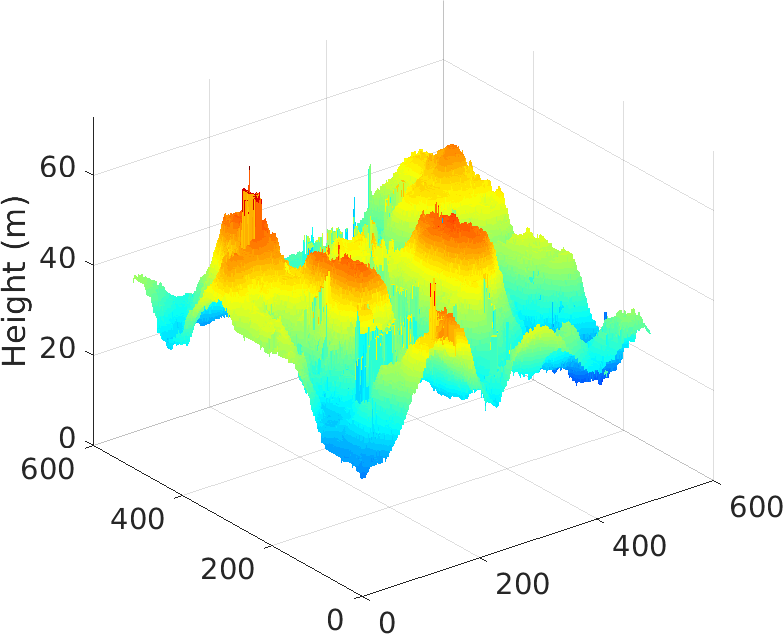}{\fpathTropidual FCN_SUM_3D_HHVV.png}
\pgfdeclareimage[width=0.199\textwidth]{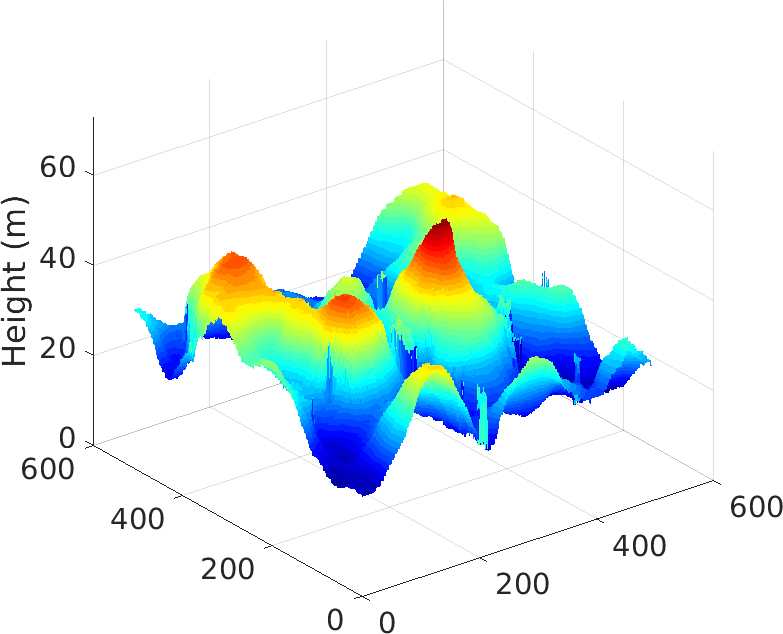}{\fpathTropidual SKP_SUM_3D_HHVV.png}

\pgfdeclareimage[width=0.199\textwidth]{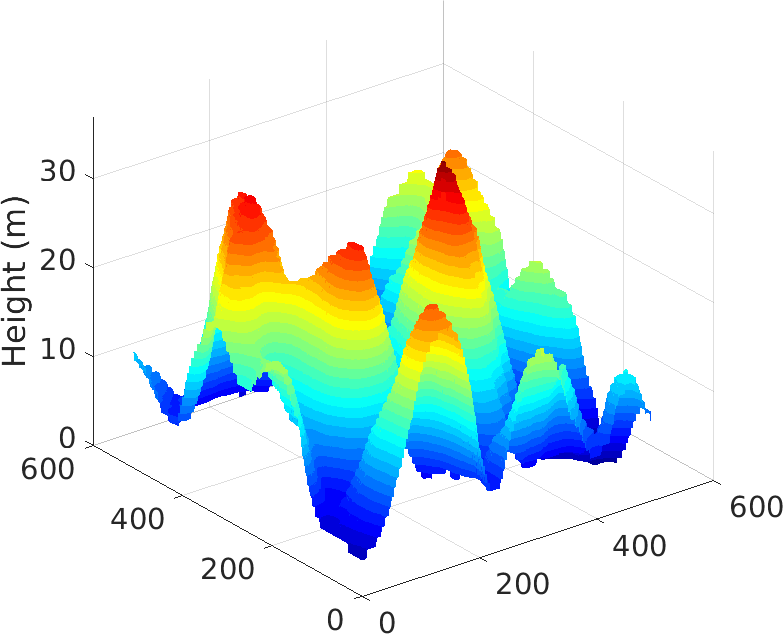}{\fpathTropidual unet_DTM_3D_HHVV.png}
\pgfdeclareimage[width=0.199\textwidth]{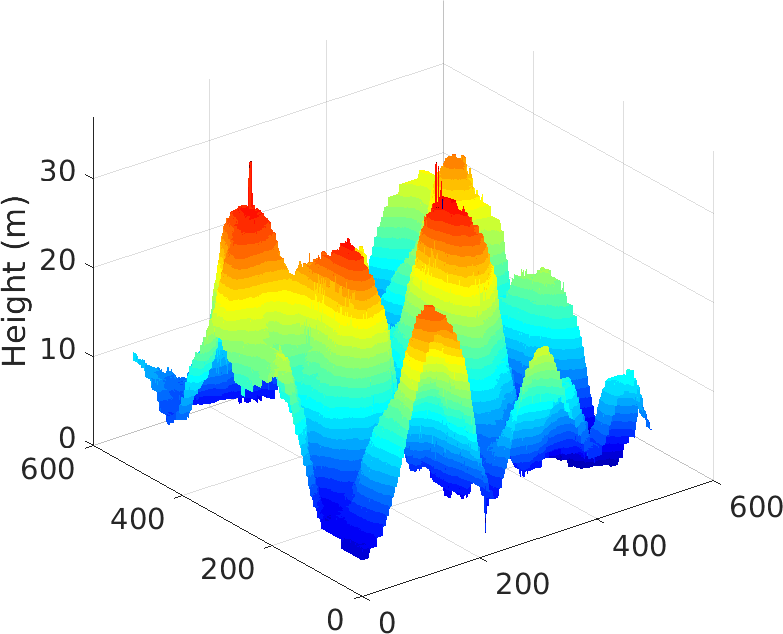}{\fpathTropidual FCN_DTM_3D_HHVV.png}
\pgfdeclareimage[width=0.199\textwidth]{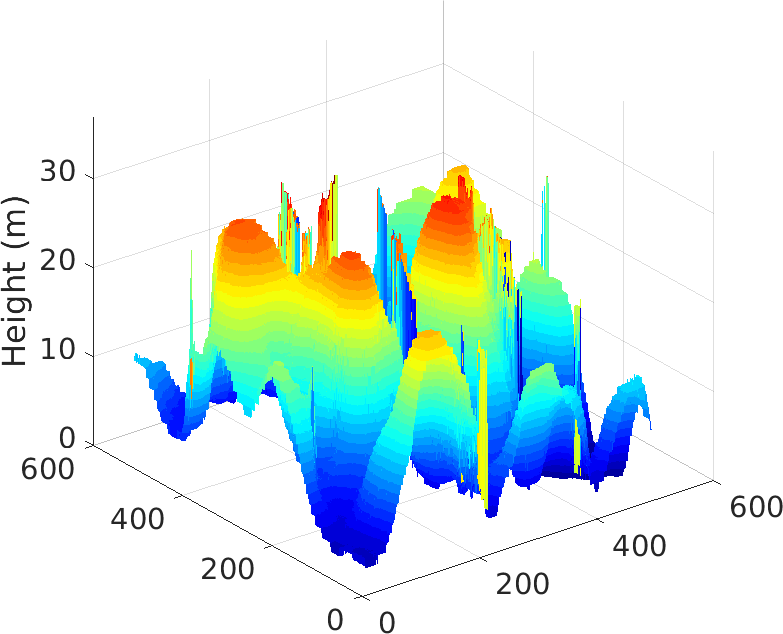}{\fpathTropidual SKP_DTM_3D_HHVV.png}

\pgfdeclareimage[width=0.2100\textwidth]{CHM_profile_1_HHVV}{\fpathTropidual CHM_profile_line1_HHVV.png}
\pgfdeclareimage[width=0.2100\textwidth]{CHM_profile_170_HHVV}{\fpathTropidual CHM_profile_line170_HHVV.png}
\pgfdeclareimage[width=0.2100\textwidth]{CHM_profile_280_HHVV}{\fpathTropidual CHM_profile_line280_HHVV.png}
\pgfdeclareimage[width=0.2100\textwidth]{DTM_profile_1_HHVV}{\fpathTropidual DTM_profile_line1_HHVV.png}
\pgfdeclareimage[width=0.2100\textwidth]{DTM_profile_170_HHVV}{\fpathTropidual DTM_profile_line170_HHVV.png}
\pgfdeclareimage[width=0.2100\textwidth]{DTM_profile_280_HHVV}{\fpathTropidual DTM_profile_line280_HHVV.png}
\pgfdeclareimage[width=0.2100\textwidth]{SUM_profile_1_HHVV}{\fpathTropidual SUM_profile_line1_HHVV.png}
\pgfdeclareimage[width=0.2100\textwidth]{SUM_profile_170_HHVV}{\fpathTropidual SUM_profile_line170_HHVV.png}
\pgfdeclareimage[width=0.2100\textwidth]{SUM_profile_280_HHVV}{\fpathTropidual SUM_profile_line280_HHVV.png}

\pgfdeclareimage[width=0.26\columnwidth]{CHM_joint_unet_HHVV}{\fpathTropidual CHM_jointd_unet_HHVV.png}
\pgfdeclareimage[width=0.26\columnwidth]{CHM_joint_fcn_HHVV}{\fpathTropidual CHM_jointd_fcn_HHVV.png}
\pgfdeclareimage[width=0.26\columnwidth]{CHM_joint_boss_HHVV}{\fpathTropidual CHM_jointd_boss_HHVV.png}
\pgfdeclareimage[width=0.26\columnwidth]{CHM_joint_skp_HHVV}{\fpathTropidual CHM_jointd_skp_HHVV.png}

\pgfdeclareimage[width=0.26\columnwidth]{DTM_joint_unet_HHVV}{\fpathTropidual DTM_jointd_unet_HHVV.png}
\pgfdeclareimage[width=0.26\columnwidth]{DTM_joint_fcn_HHVV}{\fpathTropidual DTM_jointd_fcn_HHVV.png}
\pgfdeclareimage[width=0.26\columnwidth]{DTM_joint_boss_HHVV}{\fpathTropidual DTM_jointd_boss_HHVV.png}
\pgfdeclareimage[width=0.26\columnwidth]{DTM_joint_skp_HHVV}{\fpathTropidual DTM_jointd_skp_HHVV.png}

\pgfdeclareimage[width=0.26\columnwidth]{SUM_joint_unet_HHVV}{\fpathTropidual SUM_jointd_unet_HHVV.png}
\pgfdeclareimage[width=0.26\columnwidth]{SUM_joint_fcn_HHVV}{\fpathTropidual SUM_jointd_fcn_HHVV.png}
\pgfdeclareimage[width=0.26\columnwidth]{SUM_joint_boss_HHVV}{\fpathTropidual SUM_jointd_boss_HHVV.png}
\pgfdeclareimage[width=0.26\columnwidth]{SUM_joint_skp_HHVV}{\fpathTropidual SUM_jointd_skp_HHVV.png}

\pgfdeclareimage[width=0.199\textwidth]{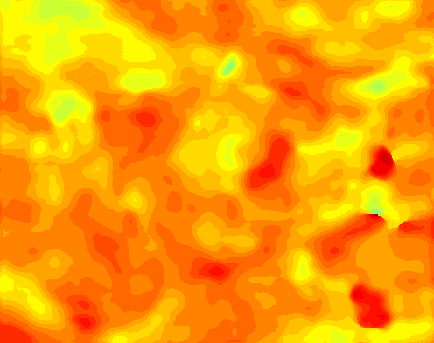}{\fpathTropidual unet_CHM_2D_HVVV.png}
\pgfdeclareimage[width=0.199\textwidth]{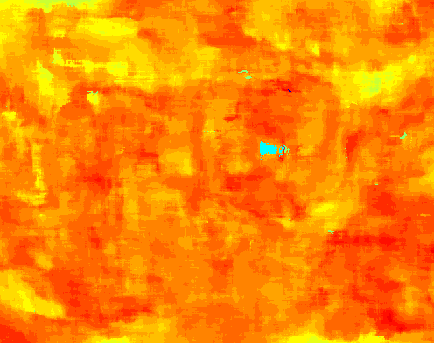}{\fpathTropidual FCN_CHM_2D_HVVV.png}
\pgfdeclareimage[width=0.199\textwidth]{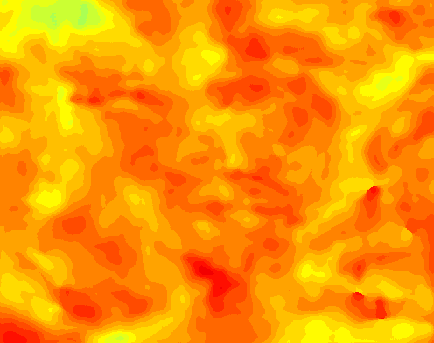}{\fpathTropidual BOSS_CHM_2D_HVVV.png}
\pgfdeclareimage[width=0.199\textwidth]{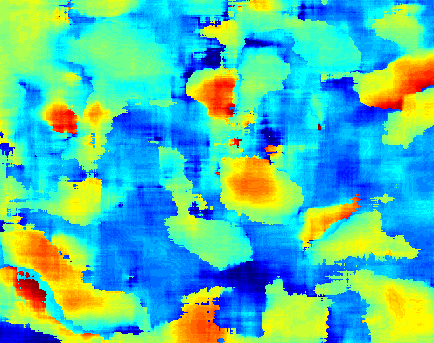}{\fpathTropidual SKP_CHM_2D_HVVV.png}

\pgfdeclareimage[width=0.199\textwidth]{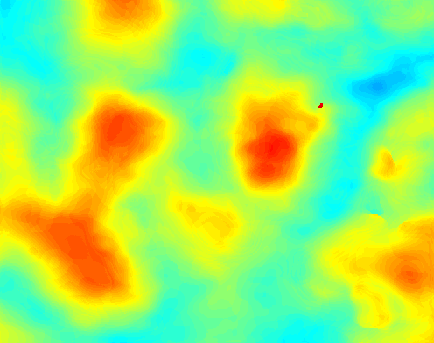}{\fpathTropidual unet_SUM_2D_HVVV.png}
\pgfdeclareimage[width=0.199\textwidth]{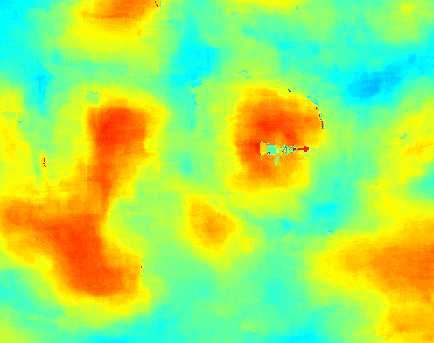}{\fpathTropidual FCN_SUM_2D_HVVV.png}
\pgfdeclareimage[width=0.199\textwidth]{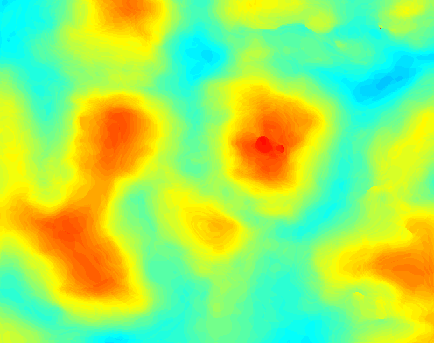}{\fpathTropidual BOSS_SUM_2D_HVVV.png}
\pgfdeclareimage[width=0.199\textwidth]{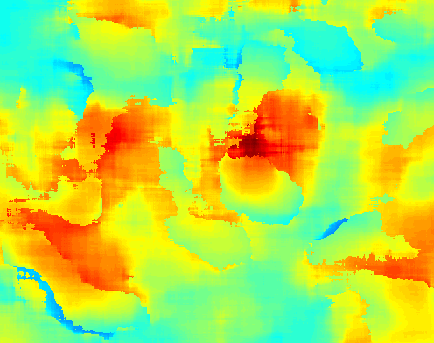}{\fpathTropidual SKP_SUM_2D_HVVV.png}

\pgfdeclareimage[width=0.199\textwidth]{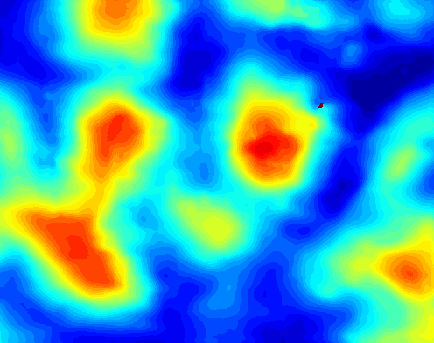}{\fpathTropidual unet_DTM_2D_HVVV.png}
\pgfdeclareimage[width=0.199\textwidth]{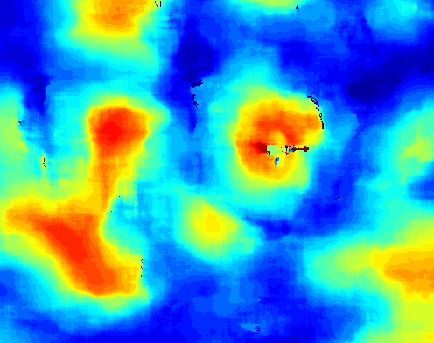}{\fpathTropidual FCN_DTM_2D_HVVV.png}
\pgfdeclareimage[width=0.199\textwidth]{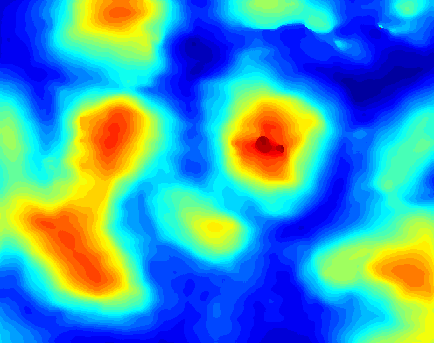}{\fpathTropidual BOSS_DTM_2D_HVVV.png}
\pgfdeclareimage[width=0.199\textwidth]{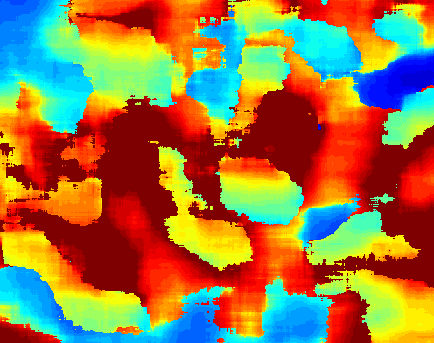}{\fpathTropidual SKP_DTM_2D_HVVV.png}

\pgfdeclareimage[width=0.199\textwidth]{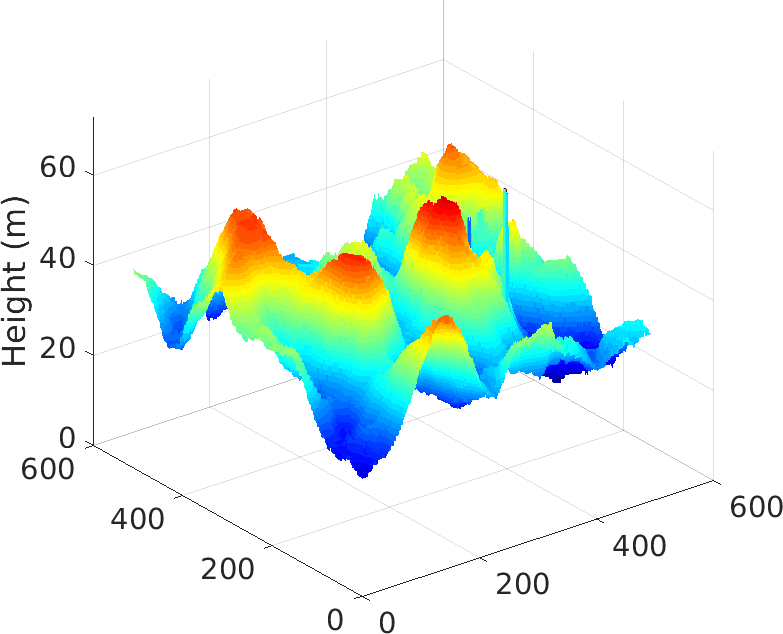}{\fpathTropidual unet_SUM_3D_HVVV.png}
\pgfdeclareimage[width=0.199\textwidth]{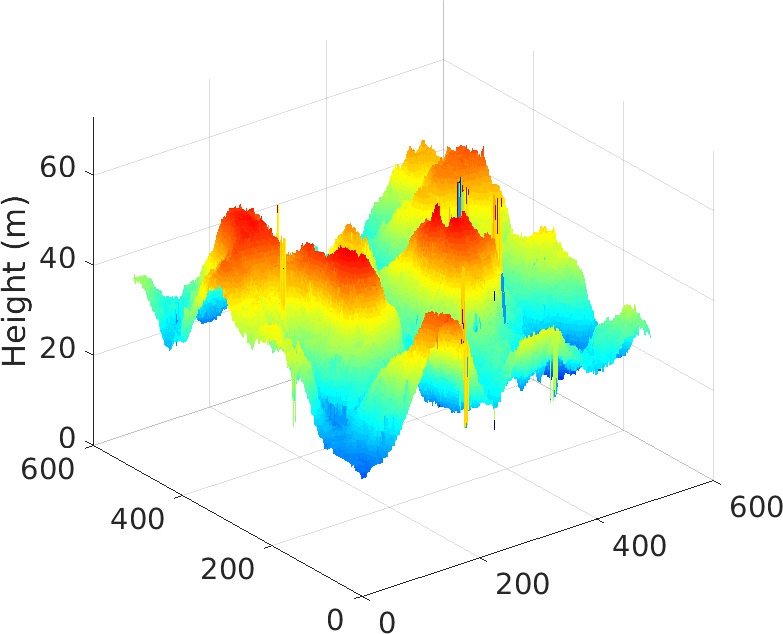}{\fpathTropidual FCN_SUM_3D_HVVV.png}
\pgfdeclareimage[width=0.199\textwidth]{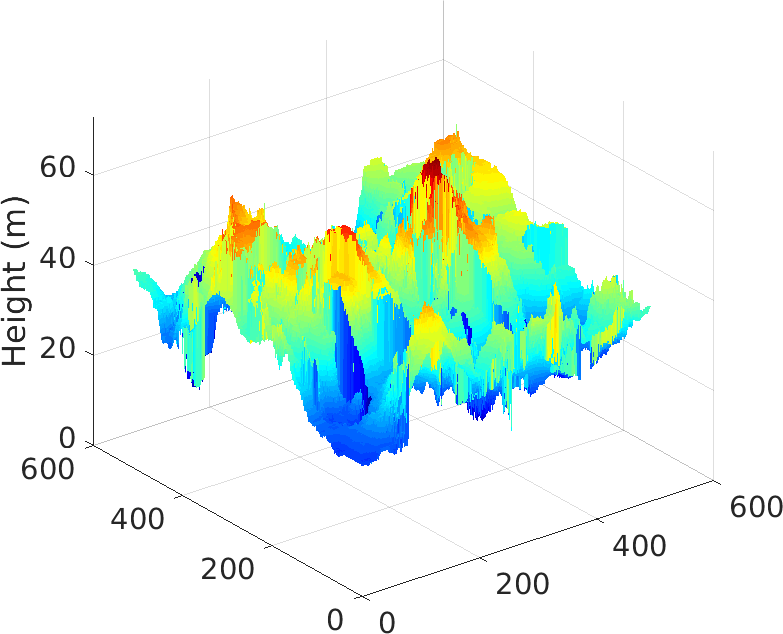}{\fpathTropidual SKP_SUM_3D_HVVV.png}

\pgfdeclareimage[width=0.199\textwidth]{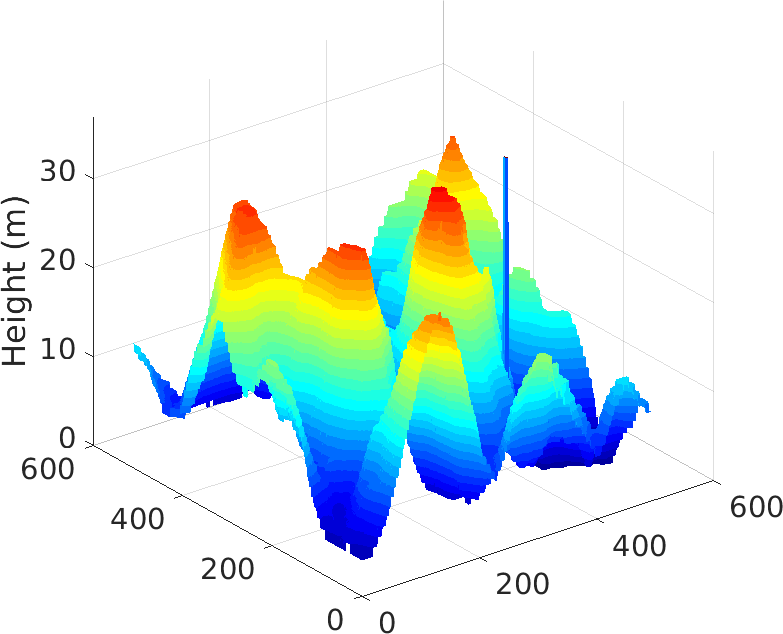}{\fpathTropidual unet_DTM_3D_HVVV.png}
\pgfdeclareimage[width=0.199\textwidth]{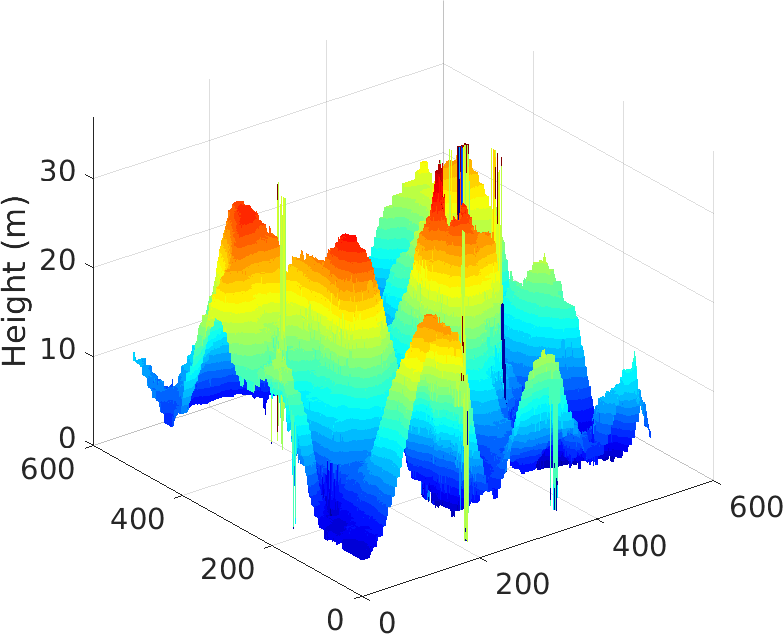}{\fpathTropidual FCN_DTM_3D_HVVV.png}
\pgfdeclareimage[width=0.199\textwidth]{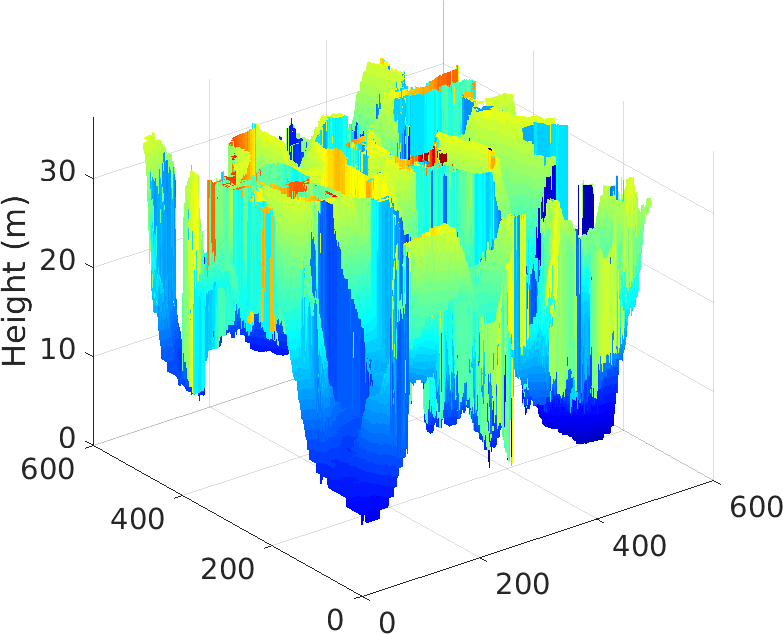}{\fpathTropidual SKP_DTM_3D_HVVV.png}

\pgfdeclareimage[width=0.2100\textwidth]{CHM_profile_1_HVVV}{\fpathTropidual CHM_profile_line1_HVVV.png}
\pgfdeclareimage[width=0.2100\textwidth]{CHM_profile_170_HVVV}{\fpathTropidual CHM_profile_line170_HVVV.png}
\pgfdeclareimage[width=0.2100\textwidth]{CHM_profile_280_HVVV}{\fpathTropidual CHM_profile_line280_HVVV.png}
\pgfdeclareimage[width=0.2100\textwidth]{DTM_profile_1_HVVV}{\fpathTropidual DTM_profile_line1_HVVV.png}
\pgfdeclareimage[width=0.2100\textwidth]{DTM_profile_170_HVVV}{\fpathTropidual DTM_profile_line170_HVVV.png}
\pgfdeclareimage[width=0.2100\textwidth]{DTM_profile_280_HVVV}{\fpathTropidual DTM_profile_line280_HVVV.png}
\pgfdeclareimage[width=0.2100\textwidth]{SUM_profile_1_HVVV}{\fpathTropidual SUM_profile_line1_HVVV.png}
\pgfdeclareimage[width=0.2100\textwidth]{SUM_profile_170_HVVV}{\fpathTropidual SUM_profile_line170_HVVV.png}
\pgfdeclareimage[width=0.2100\textwidth]{SUM_profile_280_HVVV}{\fpathTropidual SUM_profile_line280_HVVV.png}

\pgfdeclareimage[width=0.26\columnwidth]{CHM_joint_unet_HVVV}{\fpathTropidual CHM_jointd_unet_HVVV.png}
\pgfdeclareimage[width=0.26\columnwidth]{CHM_joint_fcn_HVVV}{\fpathTropidual CHM_jointd_fcn_HVVV.png}
\pgfdeclareimage[width=0.26\columnwidth]{CHM_joint_boss_HVVV}{\fpathTropidual CHM_jointd_boss_HVVV.png}
\pgfdeclareimage[width=0.26\columnwidth]{CHM_joint_skp_HVVV}{\fpathTropidual CHM_jointd_skp_HVVV.png}
\pgfdeclareimage[width=0.26\columnwidth]{DTM_joint_unet_HVVV}{\fpathTropidual DTM_jointd_unet_HVVV.png}
\pgfdeclareimage[width=0.26\columnwidth]{DTM_joint_fcn_HVVV}{\fpathTropidual DTM_jointd_fcn_HVVV.png}
\pgfdeclareimage[width=0.26\columnwidth]{DTM_joint_boss_HVVV}{\fpathTropidual DTM_jointd_boss_HVVV.png}
\pgfdeclareimage[width=0.26\columnwidth]{DTM_joint_skp_HVVV}{\fpathTropidual DTM_jointd_skp_HVVV.png}
\pgfdeclareimage[width=0.26\columnwidth]{SUM_joint_unet_HVVV}{\fpathTropidual SUM_jointd_unet_HVVV.png}
\pgfdeclareimage[width=0.26\columnwidth]{SUM_joint_fcn_HVVV}{\fpathTropidual SUM_jointd_fcn_HVVV.png}
\pgfdeclareimage[width=0.26\columnwidth]{SUM_joint_boss_HVVV}{\fpathTropidual SUM_jointd_boss_HVVV.png}
\pgfdeclareimage[width=0.26\columnwidth]{SUM_joint_skp_HVVV}{\fpathTropidual SUM_jointd_skp_HVVV.png}

\pgfdeclareimage[width=0.29\textwidth]{LiDAR_CHM_2D_fine}{\fpathTropifine LiDAR_CHM_2D.png}
\pgfdeclareimage[width=0.29\textwidth]{unet_CHM_2D_fine}{\fpathTropifine unet_CHM_2D.png}
\pgfdeclareimage[width=0.29\textwidth]{FCN_CHM_2D_fine}{\fpathTropifine FCN_CHM_2D.png}
\pgfdeclareimage[width=0.29\textwidth]{unet_CHM_2D_fineog}{\fpathAfri unet_CHM_2D.png}
\pgfdeclareimage[width=0.29\textwidth]{FCN_CHM_2D_fineog}{\fpathAfri FCN_CHM_2D.png}

\pgfdeclareimage[width=0.29\textwidth]{LiDAR_DTM_2D_fine}{\fpathTropifine LiDAR_DTM_2D.png}
\pgfdeclareimage[width=0.29\textwidth]{unet_DTM_2D_fine}{\fpathTropifine unet_DTM_2D.png}
\pgfdeclareimage[width=0.29\textwidth]{FCN_DTM_2D_fine}{\fpathTropifine FCN_DTM_2D.png}
\pgfdeclareimage[width=0.29\textwidth]{unet_DTM_2D_fineog}{\fpathAfri unet_DTM_2D.png}
\pgfdeclareimage[width=0.29\textwidth]{FCN_DTM_2D_fineog}{\fpathAfri FCN_DTM_2D.png}

\pgfdeclareimage[width=0.291\textwidth]{CHM_profile_300_fine_unet}{\fpathTropifine CHM_profile_line300_unet.png}
\pgfdeclareimage[width=0.291\textwidth]{CHM_profile_800_fine_unet}{\fpathTropifine CHM_profile_line800_unet.png}
\pgfdeclareimage[width=0.291\textwidth]{CHM_profile_1020_fine_unet}{\fpathTropifine CHM_profile_line1020_unet.png}
\pgfdeclareimage[width=0.291\textwidth]{CHM_profile_300_fine_fcn}{\fpathTropifine CHM_profile_line300_fcn.png}
\pgfdeclareimage[width=0.291\textwidth]{CHM_profile_800_fine_fcn}{\fpathTropifine CHM_profile_line800_fcn.png}
\pgfdeclareimage[width=0.291\textwidth]{CHM_profile_1020_fine_fcn}{\fpathTropifine CHM_profile_line1020_fcn.png}

\pgfdeclareimage[width=0.291\textwidth]{DTM_profile_300_fine_unet}{\fpathTropifine DTM_profile_line300_unet.png}
\pgfdeclareimage[width=0.291\textwidth]{DTM_profile_800_fine_unet}{\fpathTropifine DTM_profile_line800_unet.png}
\pgfdeclareimage[width=0.291\textwidth]{DTM_profile_1020_fine_unet}{\fpathTropifine DTM_profile_line1020_unet.png}
\pgfdeclareimage[width=0.291\textwidth]{DTM_profile_300_fine_fcn}{\fpathTropifine DTM_profile_line300_fcn.png}
\pgfdeclareimage[width=0.291\textwidth]{DTM_profile_800_fine_fcn}{\fpathTropifine DTM_profile_line800_fcn.png}
\pgfdeclareimage[width=0.291\textwidth]{DTM_profile_1020_fine_fcn}{\fpathTropifine DTM_profile_line1020_fcn.png}

\pgfdeclareimage[width=0.21\textwidth]{CHM_joint_unet_fine}{\fpathTropifine CHM_jointd_unet.png}
\pgfdeclareimage[width=0.21\textwidth]{CHM_joint_fcn_fine}{\fpathTropifine CHM_jointd_fcn.png}
\pgfdeclareimage[width=0.21\textwidth]{CHM_joint_unet_fineog}{\fpathAfri CHM_jointd_unet.png}
\pgfdeclareimage[width=0.21\textwidth]{CHM_joint_fcn_fineog}{\fpathAfri CHM_jointd_fcn.png}
\pgfdeclareimage[width=0.21\textwidth]{DTM_joint_unet_fine}{\fpathTropifine DTM_jointd_unet.png}
\pgfdeclareimage[width=0.21\textwidth]{DTM_joint_fcn_fine}{\fpathTropifine DTM_jointd_fcn.png}
\pgfdeclareimage[width=0.21\textwidth]{DTM_joint_unet_fineog}{\fpathAfri DTM_jointd_unet.png}
\pgfdeclareimage[width=0.21\textwidth]{DTM_joint_fcn_fineog}{\fpathAfri DTM_jointd_fcn.png}
\pgfdeclareimage[width=0.24\textwidth]{SUM_joint_unet_fine}{\fpathTropifine SUM_jointd_unet.png}
\pgfdeclareimage[width=0.24\textwidth]{SUM_joint_fcn_fine}{\fpathTropifine SUM_jointd_fcn.png}

\pgfdeclareimage[width=0.29\textwidth]{LiDAR_CHM_2D_Afri}{\fpathAfri LiDAR_CHM_2D.png}
\pgfdeclareimage[width=0.29\textwidth]{unet_CHM_2D_Afri}{\fpathAfri unet_CHM_2D.png}
\pgfdeclareimage[width=0.29\textwidth]{fcn_CHM_2D_Afri}{\fpathAfri FCN_CHM_2D.png}

\pgfdeclareimage[width=0.29\textwidth]{LiDAR_DTM_2D_Afri}{\fpathAfri LiDAR_DTM_2D.png}
\pgfdeclareimage[width=0.29\textwidth]{unet_DTM_2D_Afri}{\fpathAfri unet_DTM_2D.png}
\pgfdeclareimage[width=0.29\textwidth]{fcn_DTM_2D_Afri}{\fpathAfri FCN_DTM_2D.png}

\pgfdeclareimage[width=0.29\textwidth]{CHM_profile_300_Afri}{\fpathAfri CHM_profile_line300.png}
\pgfdeclareimage[width=0.29\textwidth]{CHM_profile_800_Afri}{\fpathAfri CHM_profile_line800.png}
\pgfdeclareimage[width=0.29\textwidth]{CHM_profile_1020_Afri}{\fpathAfri CHM_profile_line1020.png}

\pgfdeclareimage[width=0.29\textwidth]{DTM_profile_300_Afri}{\fpathAfri DTM_profile_line300.png}
\pgfdeclareimage[width=0.29\textwidth]{DTM_profile_800_Afri}{\fpathAfri DTM_profile_line800.png}
\pgfdeclareimage[width=0.29\textwidth]{DTM_profile_1020_Afri}{\fpathAfri DTM_profile_line1020.png}

\pgfdeclareimage[width=0.4\columnwidth]{CHM_jointd_unet_Afri}{\fpathAfri CHM_jointd_unet.png}
\pgfdeclareimage[width=0.4\columnwidth]{CHM_jointd_fcn_Afri}{\fpathAfri CHM_jointd_fcn.png}

\pgfdeclareimage[width=0.4\columnwidth]{DTM_jointd_unet_Afri}{\fpathAfri DTM_jointd_unet.png}
\pgfdeclareimage[width=0.4\columnwidth]{DTM_jointd_fcn_Afri}{\fpathAfri DTM_jointd_fcn.png}

\pgfdeclareimage[width=0.29\textwidth]{SUM_jointd_unet_Afri}{\fpathAfri SUM_jointd_unet.png}
\pgfdeclareimage[width=0.29\textwidth]{SUM_jointd_fcn_Afri}{\fpathAfri SUM_jointd_fcn.png}

\pgfdeclareimage[width=0.20\textwidth]{LiDAR_CHM_2D_Afrifine}{\fpathTropi LiDAR_CHM_2D.png}
\pgfdeclareimage[width=0.200\textwidth]{unet_CHM_2D_Afrifine}{\fpathAfrifine unet_CHM_2D.png}
\pgfdeclareimage[width=0.200\textwidth]{fcn_CHM_2D_Afrifine}{\fpathAfrifine FCN_CHM_2D.png}
\pgfdeclareimage[width=0.200\textwidth]{unet_CHM_2D_Afrifineog}{\fpathTropi unet_CHM_2D.png}
\pgfdeclareimage[width=0.200\textwidth]{fcn_CHM_2D_Afrifineog}{\fpathTropi FCN_CHM_2D.png}

\pgfdeclareimage[width=0.200\textwidth]{LiDAR_DTM_2D_Afrifine}{\fpathTropi LiDAR_DTM_2D.png}
\pgfdeclareimage[width=0.200\textwidth]{unet_DTM_2D_Afrifine}{\fpathAfrifine unet_DTM_2D.png}
\pgfdeclareimage[width=0.200\textwidth]{fcn_DTM_2D_Afrifine}{\fpathAfrifine FCN_DTM_2D.png}
\pgfdeclareimage[width=0.200\textwidth]{unet_DTM_2D_Afrifineog}{\fpathTropi unet_DTM_2D.png}
\pgfdeclareimage[width=0.200\textwidth]{fcn_DTM_2D_Afrifineog}{\fpathTropi FCN_DTM_2D.png}

\pgfdeclareimage[width=0.200\textwidth]{CHM_profile_1_Afrifine}{\fpathAfrifine CHM_profile_line1.png}
\pgfdeclareimage[width=0.200\textwidth]{CHM_profile_170_Afrifine}{\fpathAfrifine CHM_profile_line170.png}
\pgfdeclareimage[width=0.200\textwidth]{CHM_profile_280_Afrifine}{\fpathAfrifine CHM_profile_line280.png}

\pgfdeclareimage[width=0.200\textwidth]{DTM_profile_1_Afrifine}{\fpathAfrifine DTM_profile_line1.png}
\pgfdeclareimage[width=0.200\textwidth]{DTM_profile_170_Afrifine}{\fpathAfrifine DTM_profile_line170.png}
\pgfdeclareimage[width=0.200\textwidth]{DTM_profile_280_Afrifine}{\fpathAfrifine DTM_profile_line280.png}

\pgfdeclareimage[width=0.200\textwidth]{SUM_profile_1_Afrifine}{\fpathAfrifine SUM_profile_line1.png}
\pgfdeclareimage[width=0.200\textwidth]{SUM_profile_170_Afrifine}{\fpathAfrifine SUM_profile_line170.png}
\pgfdeclareimage[width=0.200\textwidth]{SUM_profile_280_Afrifine}{\fpathAfrifine SUM_profile_line280.png}

\pgfdeclareimage[width=0.200\textwidth]{CHM_jointd_unet_Afrifine}{\fpathAfrifine CHM_jointd_unet.png}
\pgfdeclareimage[width=0.200\textwidth]{CHM_jointd_fcn_Afrifine}{\fpathAfrifine CHM_jointd_fcn.png}
\pgfdeclareimage[width=0.200\textwidth]{CHM_jointd_unet_Afrifineog}{\fpathTropi CHM_jointd_unet.png}
\pgfdeclareimage[width=0.200\textwidth]{CHM_jointd_fcn_Afrifineog}{\fpathTropi CHM_jointd_fcn.png}

\pgfdeclareimage[width=0.200\textwidth]{DTM_jointd_unet_Afrifineog}{\fpathTropi DTM_jointd_unet.png}
\pgfdeclareimage[width=0.200\textwidth]{DTM_jointd_fcn_Afrifineog}{\fpathTropi DTM_jointd_fcn.png}
\pgfdeclareimage[width=0.200\textwidth]{DTM_jointd_unet_Afrifine}{\fpathAfrifine DTM_jointd_unet.png}
\pgfdeclareimage[width=0.200\textwidth]{DTM_jointd_fcn_Afrifine}{\fpathAfrifine DTM_jointd_fcn.png}

\pgfdeclareimage[width=0.200\textwidth]{SUM_jointd_unet_Afrifine}{\fpathAfrifine SUM_jointd_unet.png}
\pgfdeclareimage[width=0.200\textwidth]{SUM_jointd_fcn_Afrifine}{\fpathAfrifine SUM_jointd_fcn.png}

\pgfdeclareimage[width=0.210\textwidth]{LiDAR_CHM_3Dafri}{\fpathAfrithreeD LiDAR_CHM_3D.png}
\pgfdeclareimage[width=0.210\textwidth]{unet_CHM_3Dafri}{\fpathAfrithreeD unet_CHM_3D.png}
\pgfdeclareimage[width=0.210\textwidth]{fcn_CHM_3Dafri}{\fpathAfrithreeD FCN_CHM_3D.png}
\pgfdeclareimage[width=0.210\textwidth]{unet_CHM_3D_ogafri}{\fpathAfrithreeD unet_CHM_3D_og.png}
\pgfdeclareimage[width=0.210\textwidth]{fcn_CHM_3D_ogafri}{\fpathAfrithreeD FCN_CHM_3D_og.png}

\pgfdeclareimage[width=0.210\textwidth]{LiDAR_SUM_3Dafri}{\fpathAfrithreeD LiDAR_SUM_3D.png}
\pgfdeclareimage[width=0.210\textwidth]{unet_SUM_3Dafri}{\fpathAfrithreeD unet_SUM_3D.png}
\pgfdeclareimage[width=0.210\textwidth]{fcn_SUM_3Dafri}{\fpathAfrithreeD FCN_SUM_3D.png}
\pgfdeclareimage[width=0.210\textwidth]{unet_SUM_3D_ogafri}{\fpathAfrithreeD unet_SUM_3D_og.png}
\pgfdeclareimage[width=0.210\textwidth]{fcn_SUM_3D_ogafri}{\fpathAfrithreeD FCN_SUM_3D_og.png}

\pgfdeclareimage[width=0.210\textwidth]{LiDAR_DTM_3Dafri}{\fpathAfrithreeD LiDAR_DTM_3D.png}
\pgfdeclareimage[width=0.210\textwidth]{unet_DTM_3Dafri}{\fpathAfrithreeD unet_DTM_3D.png}
\pgfdeclareimage[width=0.210\textwidth]{fcn_DTM_3Dafri}{\fpathAfrithreeD FCN_DTM_3D.png}
\pgfdeclareimage[width=0.210\textwidth]{unet_DTM_3D_ogafri}{\fpathAfrithreeD unet_DTM_3D_og.png}
\pgfdeclareimage[width=0.210\textwidth]{fcn_DTM_3D_ogafri}{\fpathAfrithreeD FCN_DTM_3D_og.png}

\pgfdeclareimage[width=0.210\textwidth]{fcn_SUM_3D_LiDARDTM}{\fpathnewthreeDDSM FCN_SUM_3D_goodDTM.png}
\pgfdeclareimage[width=0.210\textwidth]{skp_SUM_3D_LiDARDTM}{\fpathnewthreeDDSM skp_SUM_3D_goodDTM.png}
\pgfdeclareimage[width=0.210\textwidth]{unet_SUM_3D_LiDARDTM}{\fpathnewthreeDDSM unet_SUM_3D_goodDTM.png}

\begin{document}
\title{CATSNet: a context-aware network for Height Estimation in a Forested Area based on Pol-TomoSAR data}

\author{Wenyu~Yang,~\IEEEmembership{Student~Member,~IEEE,} Sergio~Vitale,~\IEEEmembership{Member,~IEEE,} Hossein~Aghababaei,~\IEEEmembership{Senior~Member,~IEEE,}
Giampaolo~Ferraioli,~\IEEEmembership{Senior~Member,~IEEE,} Vito~Pascazio,~\IEEEmembership{Fellow,~IEEE}
and~Gilda~Schirinzi,~\IEEEmembership{Senior~Member,~IEEE}
\thanks{Wenyu Yang is with Dipartimento di Ingegneria, Università degli Studi di Napoli “Parthenope,” 80143 Naples, Italy. (email:
wenyu.yang001@studenti.uniparthenope.it)

Sergio Vitale, Vito Pascazio and Gilda Schirinzi are with Dipartimento di Ingegneria, Università degli Studi di Napoli “Parthenope,” 80143 Naples, Italy and also with the National Inter-University Consortium for Telecommunications — CNIT, 80126 Naples, Italy.
(email: sergio.vitale@uniparthenope.it, vito.pascazio@uniparthenope.it, gilda.schirinzi@uniparthenope.it)

Giampaolo Ferraioli is with Dipartimento di Scienze e Tecnologie, Università degli Studi di Napoli “Parthenope,” 80143 Naples, Italy and also with the National Inter-University Consortium for Telecommunications — CNIT, 80126 Naples, Italy. (email: giampaolo.ferraioli@uniparthenope.it)

Hossein Aghababaei is with the Department of Earth Observation Science, Faculty of Geo-Information Science and Earth Observation, University of Twente, Enschede 7514AE, The Netherlands. (e-mail: h.aghababaei@utwente.nl).}
\\
\textcolor{blue}{This paper has been submitted to IEEE TGRS. At the moment is under review (first round).}
}



\maketitle

\begin{abstract}
Tropical forests are a key component of the global carbon cycle. With plans for upcoming space-borne missions like BIOMASS to monitor forestry, several airborne missions, including TropiSAR and AfriSAR campaigns, have been successfully launched and experimented. Typical Synthetic Aperture Radar Tomography (TomoSAR) methods involve complex models with low accuracy and high computation costs. 
In recent years, deep learning methods have also gained attention in the TomoSAR framework, showing interesting performance.
Recently, a solution based on a fully connected Tomographic Neural Network (TSNN) has demonstrated its effectiveness in
accurately estimating forest and ground heights by exploiting the pixel-wise elements of the covariance matrix derived from TomoSAR data.
This work instead goes beyond the pixel-wise approach to define a context-aware deep learning-based solution named CATSNet. A convolutional neural network is considered to leverage patch-based information and extract features from a neighborhood rather than focus on a single pixel.
The training is conducted by considering TomoSAR data as the input and Light Detection and Ranging (LiDAR) values as the ground truth. The experimental results show striking advantages in both performance and generalization ability by leveraging context information within Multiple Baselines (MB) TomoSAR data across different polarimetric modalities, surpassing existing techniques.

\end{abstract}


\section{Introduction}
\IEEEPARstart{A}{l}though tropical forests only cover about $6\%$ of Earth's surface, they provide a wide range of ecosystem services \cite{forestsystem}\cite{labriere2015soil}. They harbor a disproportionately high fraction of global terrestrial biodiversity \cite{simon2005}\cite{ferry2015tree}, stock about a quarter of total terrestrial carbon, and contribute up to a third of net primary production to play an important role in the global carbon cycle \cite{corlett2016impacts}.

In particular, the forest structure is a significant indicator of productivity and biomass level, so it becomes a major element for its measurement \cite{hardiman2011role}\cite{ishii2004exploring}. There are several ways to monitor forests. 
\textit{In-situ} measurement provides the most accurate parameters but represents a costly and time-consuming solution. 
Remote sensing data instead offers a wider perspective thanks to the large-scale, consistent, and short revisiting acquisition intervals. Among remote sensing tools, LiDAR provides the most accurate vertical resolution, making it the optimal solution for monitoring forest structure. Its effectiveness is however constrained by the high cost of airborne-based acquisition campaigns. 
SAR Tomography (TomoSAR) equipped with multiple baselines (MB) acquisitions is the other remote sensing technique that provides 3D imaging with vertical resolution capability. With the under-foliage penetration capability of radar signals, TomoSAR is powerfully suited to 3D measurements of forests. 
Furthermore, with benefits from polarimetric information, TomoSAR has been effectively utilized to differentiate scattering mechanisms in the vertical direction of forestry, enabling the separation of ground and above-ground contributions in SAR signal \cite{ImSKP31}\cite{ImSKP28}.
To reconstruct the forest profiles from TomoSAR data, several models have been proposed ranging from Fourier-based methods to super-resolution solutions. Among them, MUSIC (Multiple Signal Classification) \cite{Lombardini2013} is confined to be applied to sparse data. Traditional Fourier-based algorithms such as beamforming and Capon \cite{stoica2005spectral} suffer from severe sidelobes with worse resolution resulting from uneven and limited baseline sampling \cite{MB9}. Compressive sensing (CS) \cite{MB12} \cite{Aguilera} methods tend to generate high-resolution results, but multiple iterations are highly computation-loaded. Additionally, decomposition-based methods have been studied for the identification and separation of different scattering mechanisms (SM) including double bounce and volumetric scattering \cite{ImSKP30}\cite{Hossein} with the help of Pol-TomoSAR. In \cite{ImSKP29}, the covariance matrix is expressed as a Sum of Kronecker Products which provides the basis for performing the SM separation by model-based, model-free, and hybrid approaches. Methods based on Generalized Likelihood Ratio Test (GLRT) \cite{Hossein} are characterized by the advantages of canopy and ground separation based on Random-Volume-over-Ground (RVOG) with TomoSAR data.

In the past few years, deep learning (DL) solutions have garnered significant interest in the field of TomoSAR applications.
A preliminary DL approach \cite{me17} has been proposed for urban structure height measurement with TomoSAR data. Building heights are reconstructed by formulating the problem as a classification task. The experiment on simulated data primarily shows its potential. Moving to the forest scenario, an unsupervised DL-based approach named PolGAN \cite{me16} treats the forest height estimation task as a pan-sharpening problem to realize the forest height estimation with high accuracy and fine spatial resolution. The main drawback of PolGAN is the requirement for LiDAR data also in the testing stage, which, even in low resolution, is not always available.  Recently, \cite{dlberenger} proposed a lightweight neural network for improving the vertical resolution profile generated by beamforming. Indeed, \textit{Berenger et al.} train the network without the need of LiDAR data but using simulated reference height profiles.
In \cite{yangTGRS2023} the TSNN (Tomographic SAR Neural Network) algorithm for height estimation in forested areas has been proposed. TSNN is a pixel-wise solution that formulates height estimation as a classification task showing remarkable performance on TropiSAR campaign using both full, dual and single polarized data \cite{10286267}. Being pixel-wise, TSNN however suffers from the lack of neighboring pixels' information, resulting in noisy height maps that could negatively impact the measurement of forests' structure. 

To go beyond such limitations, this paper proposes a flexible and context-aware DL solution for forest height reconstruction, which is named CATSNet (Context-Aware TomoSAR Network). Instead of leveraging single-pixel acquisitions, a patch-based solution is depicted for taking advantage of neighborhood information. Leveraging the availability of LiDAR as vertical reference data, a CNN-based solution has been trained for retrieving forest and ground height profiles from Pol-TomoSAR data. In particular, the input patches are derived from the covariance matrix elements of the corresponding patches of the Pol-TomoSAR data stack. The annotations of the input patches are the quantized LiDAR-based heights (LiDAR-based CHM and LiDAR-based DTM). 
The aim is to extract features related to heights not only from the covariance matrix elements of the corresponding pixel but also from the neighboring pixels, which means instead of pixel-wise, the proposed method is conducted at a patch-wise level. The reason is twofold: the information in the surrounding target pixels can help in the final estimation since the ground and canopy height profiles typically exhibit a high spatial correlation, and leveraging contextual information is crucial for limiting the speckle effect and making a more robust and flexible solution easily adaptable to different areas.
As a CNN-based solution, CATSNet's architecture is borrowed from U-Net \cite{ronneberger2015u} for its good use of context while maintaining good localization with few images.
Moreover, special attention has been paid to the generalization ability of the proposed solution in terms of flexibility in new areas.

The paper is organized as follows: in Section \ref{sec: methodology}, the methodology is indicated on network architecture and training parameters; Study area and dataset construction are described in Section \ref{sec: data}. The description of experimental areas and the analysis of the testing results are presented in Section \ref{sec: experiments}.

\section{Methodology}
\label{sec: methodology}
In this paper, the height estimation problem is formulated as a segmentation task: an auto-encoder convolutional network is trained for classifying each pixel of a Pol-TomoSAR stack with the corresponding LiDAR reference value. In Section \ref{sec:input_data}, details about the dataset's construction and the method's workflow are provided. The CATSNet architecture is described in Section \ref{sec:architecture}. Finally, the training parameters are presented in \ref{sec:training}.

\subsection{Data and Workflow }
\label{sec:input_data}

\begin{figure*}[!t]
\centering
\includegraphics[width= 1\textwidth]{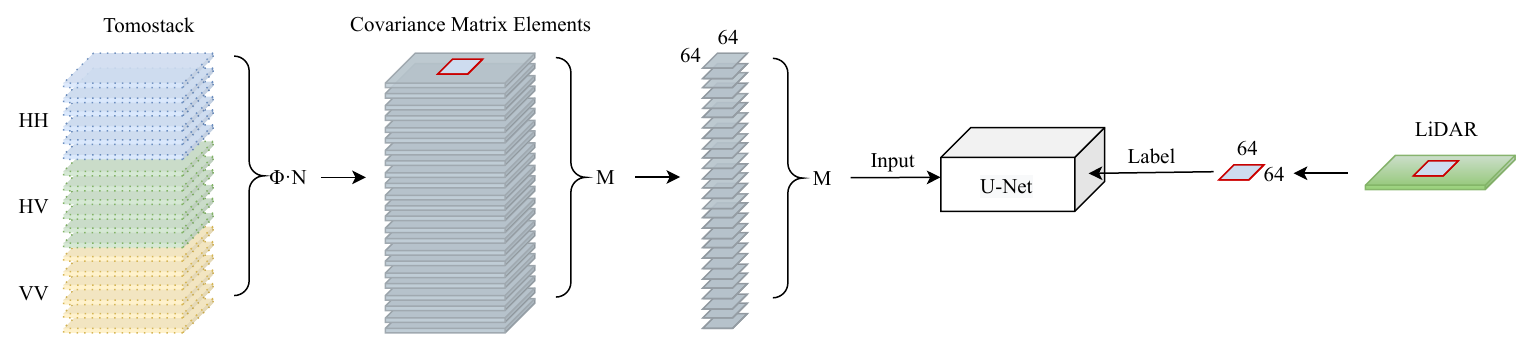}
\caption{Workflow of the proposed method. For each range-azimuth pixel of the MPMB Tomostack, the $ \Phi \times N$ real diagonal elements of the matrix $R$ and the real parts and the imaginary parts of elements in the first are stacked and considered as the channels of the input patch. The annotations consist of the corresponding quantized LiDAR-based heights: LiDAR-based CHM in the case of training targeted for prediction of forest height; and LiDAR-based DTM in the case of training targeted for prediction of ground height.}
\label{Processing}
\end{figure*}

\begin{figure*}[ht]
\centering
\includegraphics[width= 0.8\textwidth]{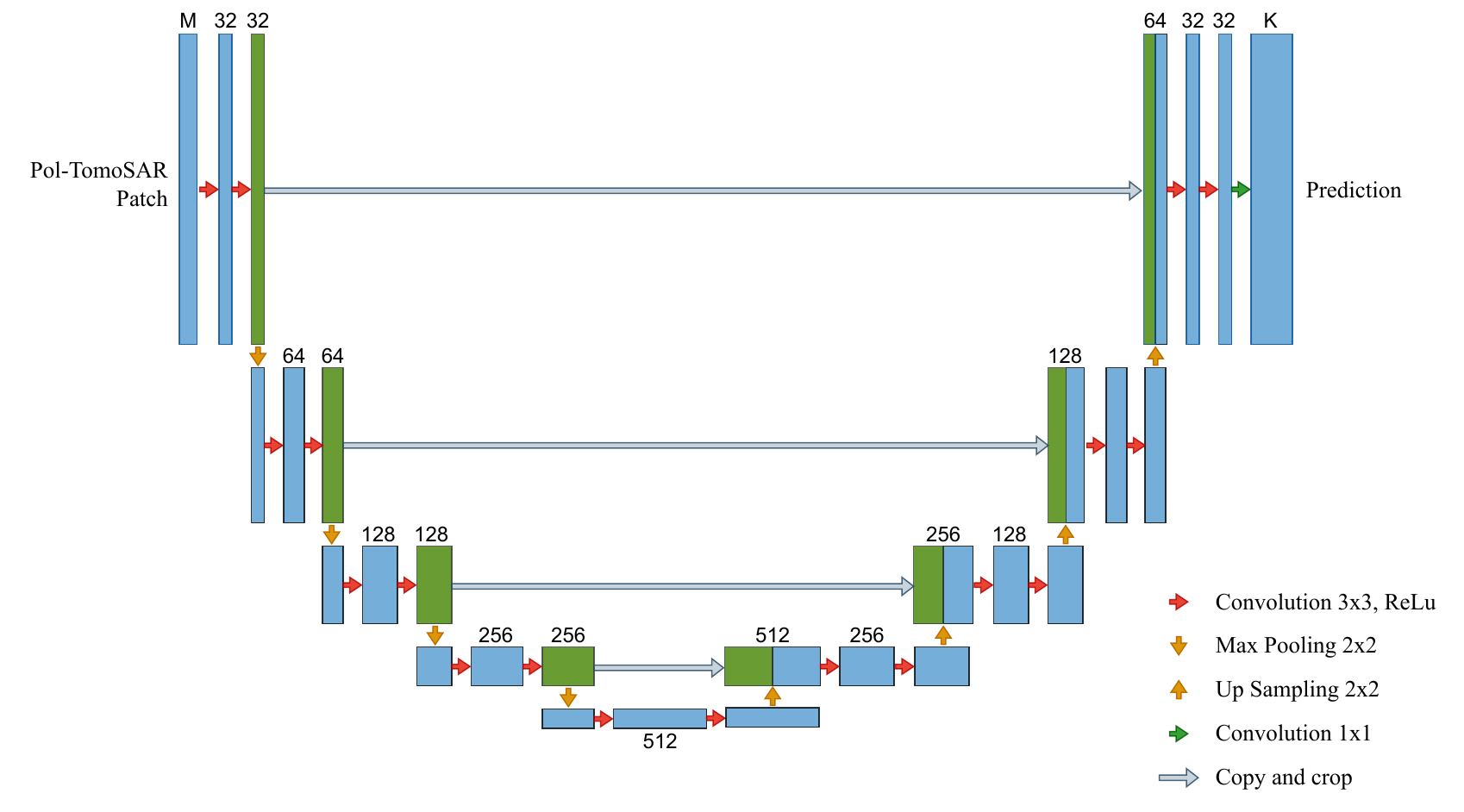}
\caption{Architecture of CATSNet. The architecture is composed of a contracting encoder path (left side) and an expansive decoder path (right side). The encoder path consists of five downsampling levels and the decoder consists of five upsampling levels.}
\label{Unetarchi}
\end{figure*}

TomoSAR methods for forest profile estimation are mostly based on the data covariance matrix. Also in this paper, we rely on the input patches obtained from the covariance matrix. We assume that TomoSAR acquires $N$ multi-baseline SLC images for each polarization channel. After several pre-processing operations applied to the SLC data stack, such as coregistration, phase flattening, and calibration,  we get the covariance matrix of each range-azimuth pixel of the SLC data stack. Let's now consider the fully polarimetric (FP) case.ì, so that the number of polarization channels is $\Phi=3$. In this case, the Pol-TomoSAR MBMP (Multiple Baselines Multiple Polarizations) data stack $\boldsymbol y$ is defined as


\begin{equation}
\label{MPMBvector}
\begin{split}
&\boldsymbol y = [{y_1}^{HH},\quad {y_2}^{HH},\quad \cdots \quad{{y_N}^{HH}},\\
&\quad \quad {y_1}^{HV},\quad {y_2}^{HV},\quad \cdots \quad{{y_N}^{HV}},\\
&\quad \quad {y_1}^{VV},\quad {y_2}^{VV},\quad \cdots \quad{{y_N}^{VV}}]^{T} 
\end{split}
\end{equation}
where $y_i^{PQ}$ denotes the pixel value of the SAR image acquired with the $i$-th baseline using $P$ as the transmitting polarization channel and $Q$ as the receiving polarization channel, where $P, Q = \{H, V\}$, and $T$ is the transpose operator. Theoretically, the computation of the covariance matrix $\mathbf{R}$ is operated on different realizations of the data vector $\mathbf{y}$. However, different realizations of the random data are not available, thereby the statistical expectation is performed by means of a spatial averaging operation on SAR images. In the considered FP model, the size of covariance matrix $\mathbf{R}$ is $3N\times3N$ and is given by:

\begin{equation}
\label{MPMBcovariance}
\begin{split}
\mathbf{R} = E\{\boldsymbol y \boldsymbol y&^{\dagger}\}=
\left[
\begin{matrix}
{\mathbf{C}_{HHHH}} & {\mathbf{\Omega}_{HHHV}} & {\mathbf{\Omega}_{HHVV}} \\
{\mathbf{\Omega}_{HVHH}}& {\mathbf{C}_{HVHV}} & {\mathbf{\Omega}_{HVVV}} \\
 {\mathbf{\Omega}_{VVHH}} & {\mathbf{\Omega}_{VVHV}} & {\mathbf{C}_{VVVV}} \\
\end{matrix}
\right]
\end{split}
\end{equation}
where the superscript $\dagger$ denotes the Hermitian transpose and $E\{ \cdot \}$ is the statistical expectation operator.
It can be noted that in the equation \eqref{MPMBcovariance}, $\mathbf{R}$ consists of $3\times 3$ sub-matrices, each of size $N \times N$. The diagonal sub-matrices ${\mathbf{C}}$ are contributed by interferometric information \cite{ImSKP31}, whereas the off-diagonal sub-matrices ${\mathbf{\Omega}}$ are cross-correlation matrices containing both polarimetric and interferometric information. For the dual polarimetric (DP) model, the number of polarization channels is $\Phi=2$. For the single polarimetric (SP) model, $\Phi=1$ and the covariance matrix $\mathbf{R}$ reduces to one of the diagonal sub-matrices ${\mathbf{C}}$.

We construct our input patches with a size of $W \times W \times M$, where $W$ represents the number of pixels in the range and azimuth directions of a local patch. The parameter $M$ denotes the dimensionality of tomographic data information in the covariance matrix. Considering $\Phi \cdot N$ real elements of the diagonal and $\Phi \cdot N - 1$ complex elements of the first row of the covariance matrix $\mathbf{R}$, the value of $M$ is given by $M=3\Phi \cdot N -2$. This is because the complex elements are split into their real and imaginary parts.

In the proposed method, quantized LiDAR-based heights are used for the annotation. Before the quantization operation, a spatial averaging operation is applied to LiDAR-based data to get a comparable spatial resolution to the SAR data. The size of the adopted averaging window is the same as the one used in the SAR covariance matrix computation. Then, a quantization with a $1$ m step is conducted on the filtered LiDAR-based heights, generating annotations $\mathbf{s} \in 
[ s_1,\quad s_2, \quad \cdots, \quad s_{K} ]$ where $K$ is the number of the height classes. The label patch size ($W \times W $) is set accordingly with the input patch one. In this context, we choose $W = 64$ as indicated in the workflow depicted in Fig.~\ref{Processing}.

\subsection{Neural Network architecture}
\label{sec:architecture}
The architecture considered in this paper is U-Net architecture, as illustrated in Fig. \ref{Unetarchi}. It consists of a contracting encoder path (left side) and an expansive decoder path (right side). The number of the input image tile channels is $M$. 
The contracting part is composed of five downsampling levels: each level is composed of two $3 \times 3$ convolutional layers, followed by a rectified linear unit (ReLU) and a $2 \times 2$ max pooling operation with stride 2 for downsampling (except for the last level). After each downsampling step, the number of feature channels is doubled. On the contrary, the expansive part is composed of five upsampling levels. As for the encoder part, each level consists of two $3 \times 3$ convolutional layers followed by ReLU and a $2 \times 2$ upsampling operation. On the contrary, in the encoder part, at each level, the number of feature maps is halved and concatenated with the corresponding feature map from the contracting path.
At the final layer, a $1\times1$ convolution is used to map each 32-component feature vector to the desired number of classes, which is the number of the forest height or ground height classes. As we can see from Fig. \ref{Unetarchi}, there are $23$ convolutional layers.
The reason behind this choice lies in two main motivations: first, in this work, the height estimation problem is referred to as a segmentation task, and CATSNet is a well-assessed solution for such a task; second, it easily allows patch-based and multi-scale feature extraction thanks to its auto-encoder nature.


\subsection{Network Training}
\label{sec:training}
A multiclass cross-entropy function \cite{murphy2012machine} is used for the loss function. In this case, $\mathbf{p}$ is the reference distribution containing a unitary value at the position corresponding to the $i$-th class embedding the LiDAR quantized value, and $\mathbf{q}$ is the estimated  \textit{a-posteriori} probability distribution for each class obtained by applying the softmax function to the output $\mathbf{\hat{H}}$. So the estimated distribution can be written as $\mathbf{q} = [ e^{h_j}/\sum_je^{h_j} \textit{ } | \textit{ } j=1,\cdots,\zeta]$ and, $\mathbf{p}=[0,\cdots,1,\cdots,0]$ with unitary values at $i$-th position. Thus, the cross-entropy loss is computed as \cite{yangTGRS2023}:
\begin{equation}
    L_{CE} = - log \left( \frac{e^{h_i}}{\sum_je^{h_j}} \right)
    \label{eq:lossfunction}
\end{equation}
CATSNet is implemented with the Pytorch framework and optimized by minimizing (\ref{eq:lossfunction}) through the SGD optimizer \cite{jia2014caffe} with a momentum of $0.9$. We start training on GeForce GTX 1080Ti GPU with 12 GB of memory. The batch size is set as 64 and the learning rate is set as $0.01$ and the factor of learning rate decay is 0.5 with 200 epochs as the period. The Xavier initialization \cite{Glorot2010} has been conducted for the weights of each fully connected layer.

\section{Study Area}
\label{sec: data}
In this paper, two study areas have been considered: the Paracou site acquired during the TropiSAR campaign carried out over forests of French Guiana by ONERA in August 2009 \cite{6146421} and the Lopè site acquired during the AfriSAR campaign carried out over the African tropical forests of Gabon by DLR in February 2016 \cite{7559240}, respectively. At the same time, small-footprint LiDAR was also collected for both datasets. 

\begin{figure}[ht]
\centering
\includegraphics[width=3in]{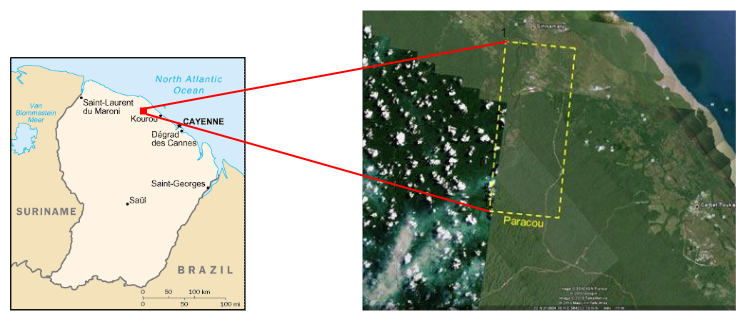}
\caption{The geographic location of Paracou site (left) and the image coverage (right).}
\label{Paracou}
\end{figure}

The Paracou site (5.27°N, 52.93°W) is located around 75 km West of Cayenne as shown in Fig.\ref{Paracou} (left) (https://paracou.cirad.fr). The altitude ranges between 5 and 45 m asl over undulating terrain. The vegetation height ranges between 0 and 60 m, and the biomass level ranges between 300 t/ha and 485 t/ha.
System parameters can be found in Table \ref{parameters}.

\begin{figure}[ht]
	\begin{tabular}{cc}
        \includegraphics[width= 1.63in]{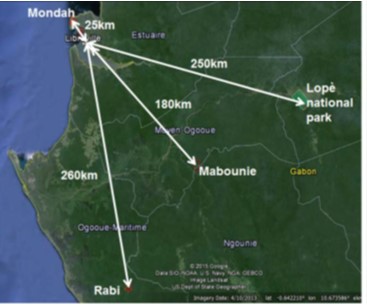}  & \includegraphics[width= 1.37in]{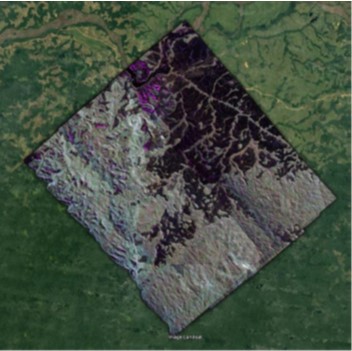}
	\end{tabular}
	\caption{The geographic location of Lopè site (left) and the image coverage (right).}
	\label{Lope}
\end{figure}

The Lopè site (0.20°S, 11.59°E) is located near the geographical center of Gabon, ca. 250 km east of Libreville, as shown in Fig.\ref{Lope} (right). The topography of this site is generally hilly, with an altitude ranging from 200 to 600 m asl \cite{8417912}. Vegetation over the Lopè site is a forest-savanna mosaic with different forest types co-occurring in the area where the forest height is distributed between 30 and 50 m with biomass ranging from about 50 to 600 t/ha \cite{cuni2016african}.
System parameters can be found in Table \ref{parameters}.

\begin{table}[ht]
  \caption{TropiSAR and AfriSAR Acquisition parameters}
  \centering 
   \begin{threeparttable}
    \begin{tabular}{c|c c}
    Parameters  & TropiSAR & AfriSAR \\
     \midrule\midrule
    Wavelength &  0.7542 m & 0.69 m      \\
    \cmidrule(l  r ){1-3}
    Flight height  &  3962 m &  6096 m    \\
    \cmidrule(l  r ){1-3}
    Incidence angle & 35.061° &  25-35° \\
    \cmidrule(l  r ){1-3}
    Range resolution & 1 m  &  3.9 m  \\
    \cmidrule(l  r ){1-3}
    Azimuth resolution & 1.245 m &  2.0 m   \\
    \cmidrule(l  r ){1-3}
    Polarization & Full-Pol & Full-Pol \\
    \cmidrule(l  r ){1-3}
     Baselines [m] & $ \begin{aligned}
    0 \\
    -14.4879 \\
    -30.1163 \\
    -43.7343 \\
    -60.0632 \\
    -74.9683 \end{aligned} $ 
    &$ \begin{aligned}
    20 \\
    0 \\
    -20 \\
    -40 \\
    -60 \\
    -80 \end{aligned} $\\ 
    \midrule\midrule
    \end{tabular}

\end{threeparttable}
\label{parameters}
\end{table}



Aiming to construct a training dataset containing the forest and ground heights, we make use of the SAR tomography stack of the acquisition and relative LiDAR data.
In particular, for each of the selected study areas, the tomographic stack is composed of six temporal SAR baselines, whose descriptions are in Table \ref{parameters}. The stack of covariance matrices estimated from the polarimetric tomographic SAR stack is used as the input for the network, while the reference LiDAR values are used as reference height profiles.
\begin{figure}[ht]
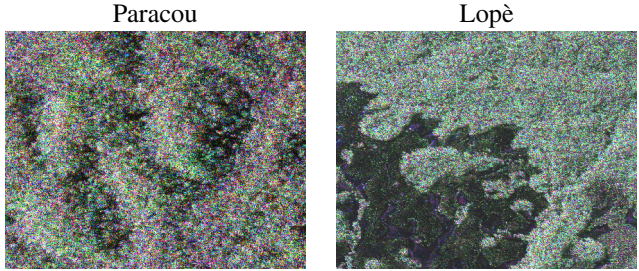

\centering
	\begin{tabular}{cc}
        Paracou & Lopè \\
         \image{TropiSAR_Pauli} & \image{AfriSAR_Pauli} \\
        
	\end{tabular}
	\caption{Pauli RGB images of testing patch extracted by Paracou (left) and Lopè areas (right).}
	\label{fig: testingPataches}
\end{figure}

In particular, the full-polarization Tomo-stack consists of $18$ SLC data contributed from HH, HV, and VV polarization channels, so here $N=6$ and $\Phi=3$. In consequence:
\begin{itemize}
    \item{For fully polarization ($\Phi=3$), the size of the covariance matrix $\mathbf{R}$ size is $18\times18$, so the channels of the input image tile $M$ is $52$. }
    \item{For dual polarization ($\Phi=2$), the size of the covariance matrix $\mathbf{R}$ size is $12\times12$, so the channels of the input image tile $M$ is $34$.}
    \item{For single polarization ($\Phi=1$), the size of the covariance matrix $\mathbf{R}$ size is $6\times6$, so the channels of the input image tile $M$ is $16$.}
\end{itemize}

As reference datasets, LiDAR-based data are available on both sites.

The French Agricultural Research Center for International Development and the Guyafor project provide the LiDAR-based data of the Paracou site. The LiDAR-based DTM was produced by triangular interpolation (TIN) of the ground data. By deducting ground elevation from the raw point cloud and extracting maximum height with a $1$ m resolution grid, the LiDAR-based CHM was generated. The forest height and ground height of a geographic position associated with a SAR image position were obtained by projecting the LiDAR-based data under WGS84 UTM zone 22 into SAR coordinates using an ASCII file supplied by TROPISAR \cite{Improving}. More details about LiDAR-based data processing are given in \cite{Improving38}

Regarding the Lopè site, the LiDAR-based data was obtained by NASA's full-waveform LiDAR LVIS (Land, Vegetation, and Ice Sensor) system. LVIS was mounted on the NASA Langley B200 aircraft and flown at 23000ft and the LiDAR footprint on the ground is around 20m wide. The data access is https://lvis.gsfc.nasa.gov/Data/Maps/Gabon2016Map.html including Level1B and Level2 products.
In our experiment, the ground topography (or Digital Terrian Model, DTM) and the canopy top height (or Canopy Height Model, CHM) in Level2 are used as the reference.

From each of Paracou and Lopè study areas, a testing patch with size $512\times512$ pixels has been extracted from the whole dataset and shown in Fig. \ref{fig: testingPataches}, respectively.
The rest of data is randomly split into the training data set (80\%) and the validating data set (20\%).

\begin{figure*}[t]
\centering
	\begin{tabular}{ccccc}
        & LiDAR & CATSNet & TSNN & SKP\\
         \rotatebox{90}{\parbox[c]{3cm}{\centering{CHM}}} & \image{LiDAR_CHM_3D} &  \image{unet_CHM_3D} &  \image{fcn_CHM_3D} & \image{skp_CHM_3D}\\
           & & \multicolumn{2}{c}{
        \begin{tabular}{c}
		\includegraphics[width = 0.3\textwidth, clip=true, trim = 0mm 0mm 0mm 150mm]{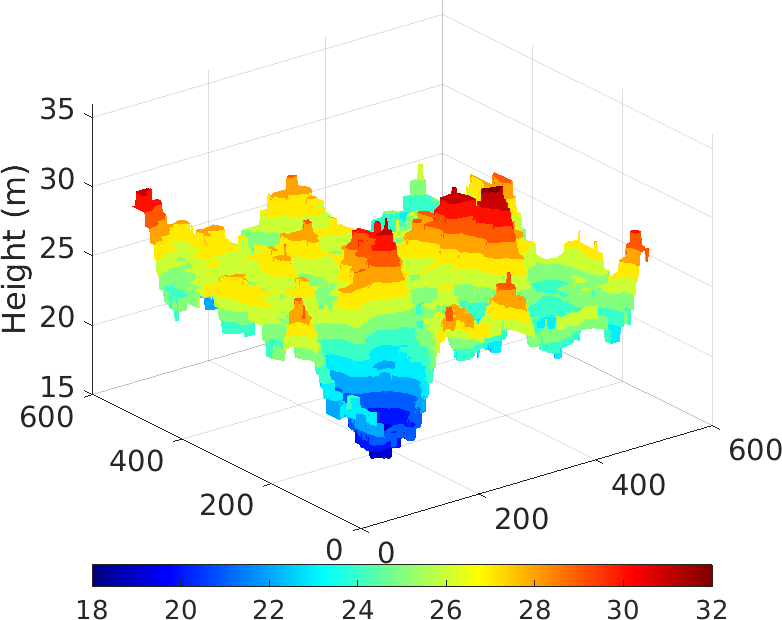}
         \end{tabular}
         } &   \\
           \rotatebox{90}{\parbox[c]{3cm}{\centering{DTM}}} & \image{LiDAR_DTM_3D} &  \image{unet_DTM_3D} & \image{FCN_DTM_3D}  & \image{skp_DTM_3D}  \\
            & & \multicolumn{2}{c}{
        \begin{tabular}{c}
		\includegraphics[width = 0.3\textwidth, clip=true, trim = 0mm 0mm 0mm 150mm]{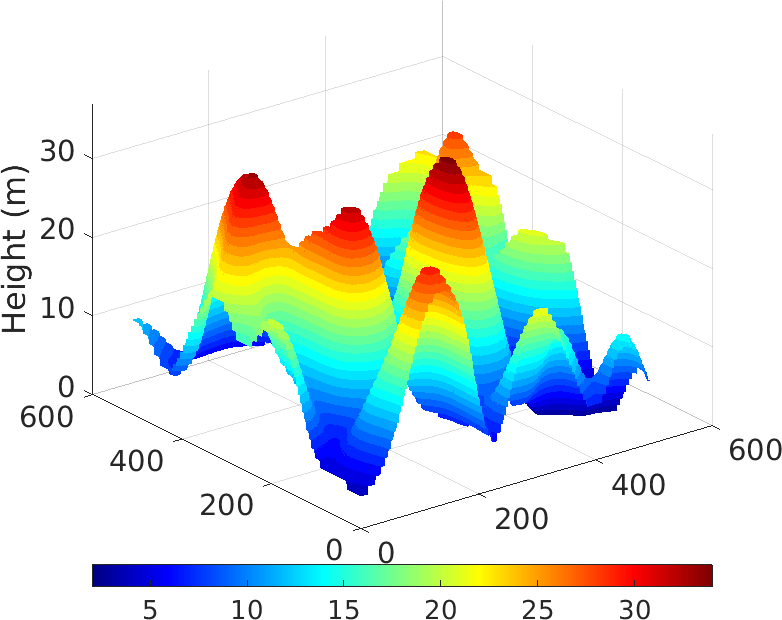}
         \end{tabular}
         } &  \\              
         \rotatebox{90}{\parbox[c]{3cm}{\centering{DSM}}} & \image{LiDAR_SUM_3D} &  \image{unet_SUM_3D_LiDARDTM} &  \image{fcn_SUM_3D_LiDARDTM} & \image{skp_SUM_3D_LiDARDTM}\\
                   & & \multicolumn{2}{c}{
                \begin{tabular}{c}
        		\includegraphics[width = 0.3\textwidth, clip=true, trim = 0mm 0mm 0mm 150mm]{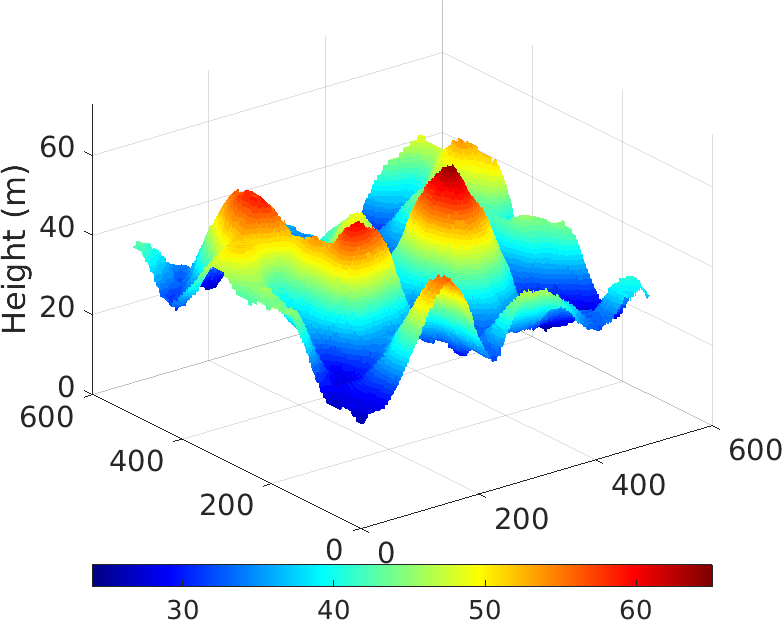}
                 \end{tabular}
                 } &   \\
                 \end{tabular}
	\caption{Full-Polarization results on Paracou. The comparison within the reference LiDAR-based values and heights estimated by CATSNet, TSNN and SKP. The first row is for forest heights, the second row is for ground heights, and the last row is for canopy heights.}
	\label{fig: Tropi3D}
\end{figure*}

\section{Experimental Results}
\label{sec: experiments}
In this section, the height estimation ability of the proposed method is presented. Comparison with three representative Tomographic methods, such as SKP, GLRT, and deep learning-based TSNN, is carried out on the Paracou area with FP, DP, and SP TropiSAR data stack. It is worth mentioning that SKP exploits the separation of different scattering mechanisms based on MPMB TomoSAR data and GLRT is characterized by the advantages of scattering mechanisms separation with single polarization TomoSAR data. Thereby, for FP and DP models, SKP is employed and for the SP model, GLRT is applied.
Moreover, the proposed solution is compared with its more timing efficient version able to predict the forest and ground height in a single shot without the need for a separate training procedure for the CHM and DTM height estimation.

\subsection{Fully polarization}
For the sake of visualization, we present a 3D representation of a real forest scenario by comparing the reference LiDAR-based CHM, DTM, and DSM (CHM+DTM) with the estimated ones by CATSNet, TSNN, and SKP in Fig.~\ref{fig: Tropi3D}, respectively.

For the forest, TSNN presents generally overestimated forest heights, including several outliers, and for SKP, there is a serious underestimation problem. Among them, CATSNet performs the best. For the ground, CATSNet still presents the best performance. In this case  TSNN also gives a good agreement with the reference, but the reconstruction is more noisy. On the contrary, SKP tends to generate overestimated ground heights, which is a common problem in traditional TomoSAR methods for ground height measurement. The overall results can be observed from the third row of Fig.~\ref{fig: Tropi3D}, showing the canopy heights (forest heights + ground heights), and CATSNet exhibits the best agreement with LiDAR-based DSM.

         

\begin{figure}[ht!]
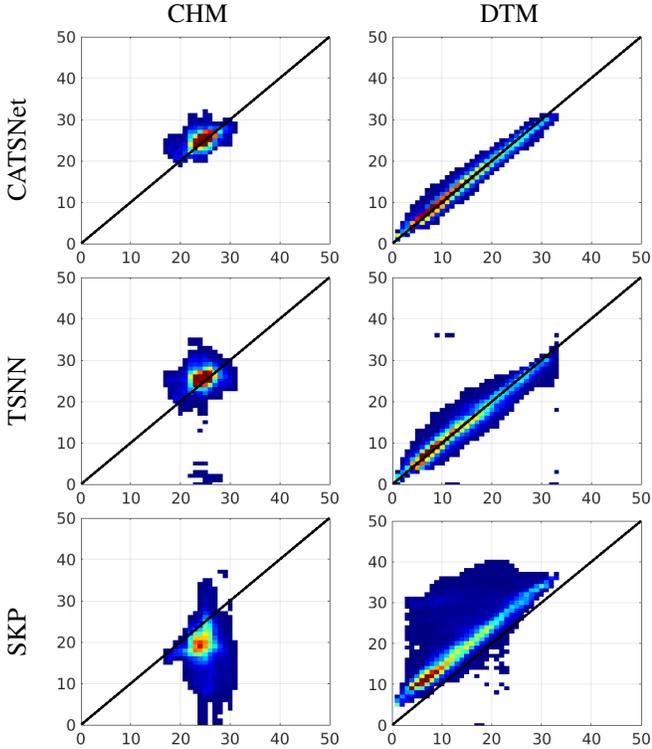

\centering
	\begin{tabular}{ccc}
        & CHM & DTM\\
         \rotatebox{90}{\parbox[c]{3cm}{\centering{CATSNet}}} & \image{CHM_joint_unet}& \image{DTM_joint_unet}  \\
        \rotatebox{90}{\parbox[c]{3cm}{\centering{TSNN}}} &  \image{CHM_joint_fcn}& \image{DTM_joint_fcn}   \\
         \rotatebox{90}{\parbox[c]{3cm}{\centering{SKP}}} & \image{CHM_joint_skp}& \image{DTM_joint_skp}   \\
         
	\end{tabular}
	\caption{Full polarization results over Paracou area. Joint distribution within the reference LiDAR-based values and heights estimated by CATSNet, TSNN, and SKP. The left column is for forest heights and the right column is for ground height. The black line represents the ideal prediction.}
	\label{fig: TropiJoint}
\end{figure}

The previous considerations are confirmed by joint distribution between predicted and expected results of Figure~\ref{fig: TropiJoint}, where the ideal result is represented by the black line. The left column shows the result comparison on forest heights, and the right column is about the ground heights.

As we can see the joint distribution of CATSNet is densely concentrated on the black line and TSNN is sparsely focused. However, the result of SKP for CHM presents a severe forest height underestimation problem. So it leads to the conclusion that CATSNet illustrates a better agreement with the LiDAR-based CHM than TSNN and SKP. The predicted ground height of testing samples perfectly located on the black line means the ideal estimation. The joint distribution of CATSNet presents the best performance compared with TSNN and SKP. In terms of TSNN, it yields noisy predicted ground heights as the result of its element-wise characteristics. Regarding SKP, the overestimated problem is further supported by that the majority of the predicted ground heights are located above the black line.
In short, we can conclude that, with a full polarization TomoSAR data stack, CATSNet outperforms the forest and ground heights estimation problem.

\subsection{Dual and Single Polarizations}
To show the robustness of the proposal, the ability to estimate forest and ground heights based on dual and single polarization TomoSAR data stack is carried out in this section.
\begin{figure}[ht]
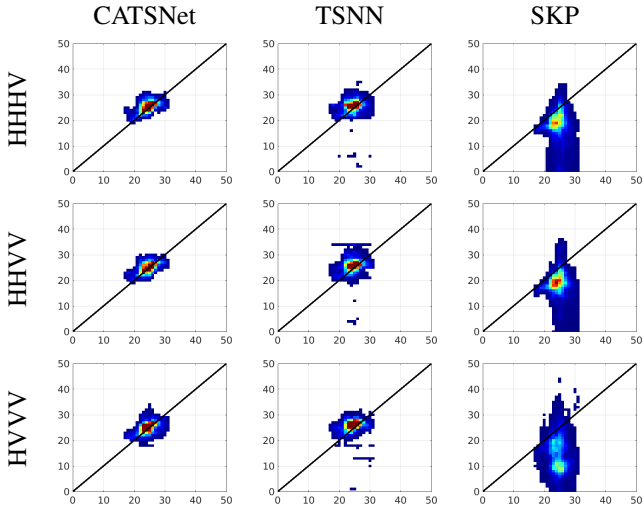

\centering
	\begin{tabular}{cccc}
	 & CATSNet & TSNN & SKP \\
        \rotatebox{90}{\parbox[c]{2cm}{\centering{HHHV}}} & \image{CHM_joint_unet_HHHV} & \image{CHM_joint_fcn_HHHV} & \image{CHM_joint_skp_HHHV}  \\
        \rotatebox{90}{\parbox[c]{2cm}{\centering{HHVV}}} & \image{CHM_joint_unet_HHVV} & \image{CHM_joint_fcn_HHVV} & \image{CHM_joint_skp_HHVV}  \\
        \rotatebox{90}{\parbox[c]{2cm}{\centering{HVVV}}} & \image{CHM_joint_unet_HVVV} & \image{CHM_joint_fcn_HVVV} & \image{CHM_joint_skp_HVVV} 
	\end{tabular}
	\caption{In the Paracou area, the joint distribution between LiDAR-based CHM and forest heights by CATSNet, TSNN, and SKP which are based on dual-polarization channels. The black line represents the ideal prediction.}
	\label{fig: TropiCHMjointdual}
\end{figure}

\begin{figure}[ht]
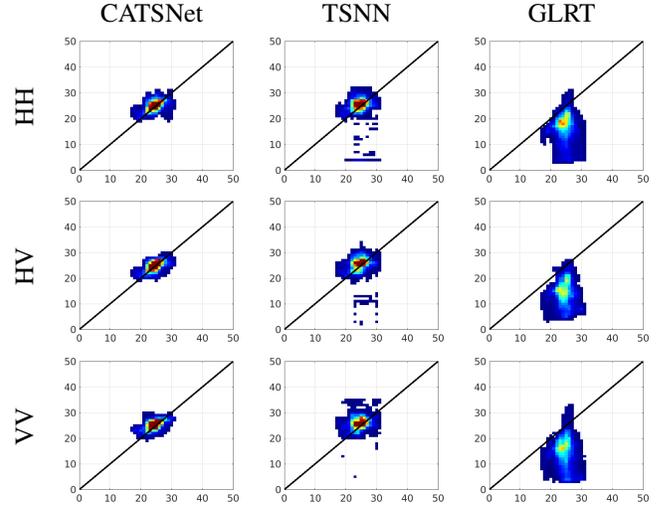

\centering
	\begin{tabular}{cccc}
	 & CATSNet & TSNN & GLRT \\
        \rotatebox{90}{\parbox[c]{2cm}{\centering{HH}}} & \image{CHM_joint_unet_HH} & \image{CHM_joint_fcn_HH} & \image{CHM_joint_m2_HH}  \\
        \rotatebox{90}{\parbox[c]{2cm}{\centering{HV}}} & \image{CHM_joint_unet_HV} & \image{CHM_joint_fcn_HV} & \image{CHM_joint_m2_HV}  \\
        \rotatebox{90}{\parbox[c]{2cm}{\centering{VV}}} & \image{CHM_joint_unet_VV} & \image{CHM_joint_fcn_VV} & \image{CHM_joint_m2_VV}  \\
	\end{tabular}
	\caption{In the Paracou area, the joint distribution between LiDAR-based CHM and forest heights by CATSNet, TSNN, and GLRT which are based on single polarization channels. The black line represents the ideal prediction.}
	\label{fig: TropiDTMjointsingle}
\end{figure}

To this aim, the SAR tomographic stack has been limited in the number of polarization channels and CATSNet has been retrained accordingly, as well as TSNN. CATSNet, TSNN, and SKP have been tested on the selected testing patch considering two polarimetric tomographic stacks, while GLRT has been considered for single polarization comparison. For the sake of reading, only joint distribution results on the forest reconstruction (CHM) are reported, but similar consideration can be done for ground reconstruction (DTM).
The joint distribution shown in Fig.~\ref{fig: TropiCHMjointdual} for the dual polarimetric case shows the robustness of CATSNet with respect to any dual polarimetric channel combination. As matter of fact, quality of prediction is almost stable in all cases HHHV, HHVV, HVVV. Similar considerations are obtained in the case of single polarization shown in Fig. \ref{fig: TropiDTMjointsingle}, where the comparison has been carried out with TSNN and GLRT. 

To reinforce the conclusion we draw in this section, quantitative results, Root-Mean-Square-Error (RMSE), of the forest height measurement are shown in Table~\ref{RMSEforestParacou} and RMSE of the ground height measurement are shown in Table~\ref{RMSEgroundParacou}.

In terms of the forest height, the RMSE of CATSNet is around 2.0 m giving the best height accuracy, while the one of TSNN is around 3.0 m and SKP gives the worst result in the full polarization model. For dual and single polarization models, the RMSE of CATSNet and TSNN don't present worse results compared to the results of full polarization. It elucidates the robustness of CATSNet and TSNN that they would maintain good performance even with less polarization information. Even in the single polarization model, the forest height measurement accuracy of CATSNet is still around 2.0 m which strongly supports CATSNet's advantage.

Regarding the measurement of the ground height, CATSNet in full polarization presents the best performance with RMSE, 1.1365 m, which is around 0.4 m better than TSNN and around 5.3 m better than SKP. When moving on to dual polarization models, CATSNet and TSNN both go worse, and SKP even performs well in the HHVV polarization channel. For single polarization, the RMSE of the ground heights by CATSNet are all better than 2.0 m which is around 1.0 m better than TSNN. GLRT shows the worst results.

\begin{table}[ht]
  \caption{RMSE of the forest height measurement on Paracou}
  \begin{threeparttable}
    \begin{tabular}{cccccc}
    Campaign  & Experiments & CATSNet  & TSNN  & SKP & GLRT\\
     \midrule\midrule
    TropiSAR  &  Full-pol  & 2.0220  &  3.0144 & 6.8195 & $ - $  \\
    \cmidrule(l  r ){1-6}
     TropiSAR & $ \begin{aligned}
    HHHV \\
    HHVV \\
    HVVV \end{aligned} $ & 
     $ \begin{aligned}
    2.1601\\
    1.9906 \\
    2.1235 \end{aligned} $ &  
    $ \begin{aligned}
     2.7294 \\
    2.9383 \\
    2.6388 \end{aligned} $ & 
     $
     \begin{aligned}
    8.6684\\
    7.0381 \\
    12.0206 \\
    \end{aligned} $ &
    $ \begin{aligned}
    - \\
    - \\
    - \end{aligned} $ \\ 
    \cmidrule(l  r ){1-6}
     TropiSAR & $ \begin{aligned}
    HH \\
    HV \\
    VV \end{aligned} $ & 
     $ \begin{aligned}
    2.0928 \\
    1.9040 \\
    2.0112 \end{aligned} $ &  
    $ \begin{aligned}
    3.2188 \\
    2.7131 \\
    2.7752 \end{aligned} $ & 
    $ \begin{aligned}
    - \\
    - \\
    - \end{aligned} $ &
    $ \begin{aligned}
     9.5828 \\
    11.2247 \\
    11.6350 \end{aligned} $\\    
\midrule\midrule 
 \end{tabular}
\end{threeparttable}
\label{RMSEforestParacou}
\end{table}

\begin{table}[ht]
  \caption{RMSE of the ground height measurement on Paracou}
  \centering 
  \begin{threeparttable}
    \begin{tabular}{cccccc}
    Campaign  & Experiments & CATSNet  & TSNN & SKP & GLRT \\
     \midrule\midrule
    TropiSAR  &  Full-pol  & 1.1365  & 1.7169  & 6.4001  & $ - $  \\
      \cmidrule(l  r ){1-6}
     TropiSAR & $ \begin{aligned}
    HHHV \\
    HHVV \\
    HVVV \end{aligned} $ & 
     $ \begin{aligned}
    1.4533\\
    1.2335 \\
    1.5966
     \end{aligned} $ &  
     $ \begin{aligned}
    2.2125 \\
    1.7623 \\
    2.3779 \end{aligned} $  &
     $ \begin{aligned}
    7.5480\\
    3.2627 \\
    15.0028 \end{aligned} $  & 
     $ \begin{aligned}
    - \\
    - \\
    - \end{aligned} $  \\ 
    \cmidrule(l  r ){1-6}
     TropiSAR & $ \begin{aligned}
    HH \\
    HV \\
    VV \end{aligned} $ & 
     $ \begin{aligned}
    1.4987 \\
    1.8983 \\
    1.8840 \end{aligned} $ & 
    $ \begin{aligned}
    2.8751 \\
    3.1609 \\
    3.1779 \end{aligned} $ & 
    $ \begin{aligned}
    - \\
    - \\
    - \end{aligned} $ &
    $ \begin{aligned}
     4.1808 \\
    9.6536 \\
    6.2612 \end{aligned} $\\
    \midrule\midrule
    \end{tabular}
\end{threeparttable} 
\label{RMSEgroundParacou}
\end{table}

\subsection{Unified CATSNet}
\label{sec: unified}
The experiment results shown above are realized by implementing the CATSNet solution on forest heights and ground heights individually. Specifically, CATSNet trained with the LiDAR forest heights as the reference is used for the forest height estimation, while CATSNet trained with the LiDAR ground heights as the reference is used for the ground estimation. Such an approach is based on a double training conduct and a double testing phase. With this in mind, we train the CATSNet with both forest height and ground height at the same time and get the trained Unified CATSNet that can be used to predict forest and ground heights in one conduction to reduce computation load and improve time efficiency. We refer to this solution as unified CATSNet. 

The structure of the Unified CATSNet is the same as the one described in section~\ref{sec:architecture} just adapting the final layer to the expected number of classes. 

\begin{figure}[ht!]
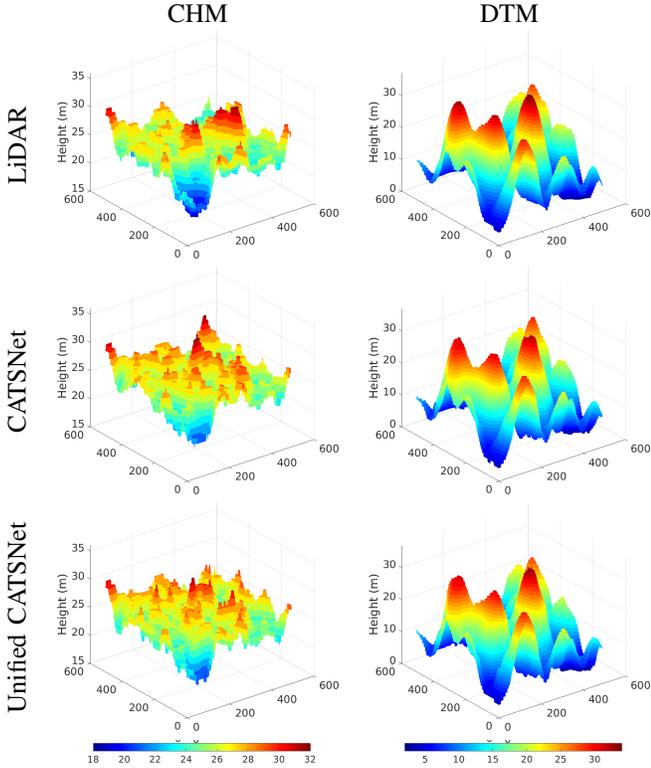

\centering
	\begin{tabular}{ccc}
        & CHM & DTM\\
         \rotatebox{90}{\parbox[c]{3cm}{\centering{LiDAR}}} & 
         \image{LiDAR_CHM_3D}&
         \image{LiDAR_DTM_3D}  \\
        \rotatebox{90}{\parbox[c]{3cm}{\centering{CATSNet}}} &  
        \image{unet_CHM_3D}&
        \image{unet_DTM_3D}   \\
         \rotatebox{90}{\parbox[c]{3cm}{\centering{Unified CATSNet}}} & 
         \image{BOSS_CHM_3D}&
         \image{BOSS_DTM_3D}   \\
         & \includegraphics[width = 0.20\textwidth, clip=true, trim = 0mm 0mm 0mm 149mm]{Figures_des/18-32Tropi.png} & \includegraphics[width = 0.20\textwidth, clip=true, trim = 0mm 0mm 0mm 149mm]{Figures_des/2-34Tropi.png}
	\end{tabular}
	\caption{Full-Polarization results on Paracou. Reconstruction comparison within the reference LiDAR-based values and heights estimated by CATSNet and Unified CATSNet. The left column is for forest heights and the right column is for ground heights.}
	\label{fig: uni3D}
\end{figure}

\begin{figure}[ht!]
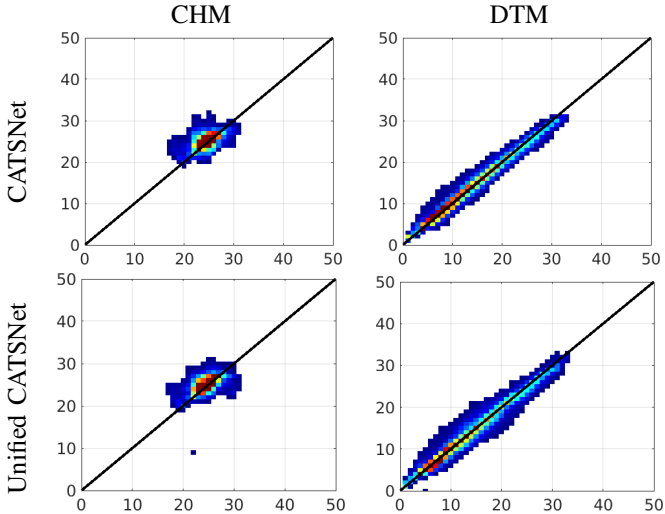

\centering
	\begin{tabular}{ccc}
        & CHM & DTM\\
         \rotatebox{90}{\parbox[c]{3cm}{\centering{CATSNet}}} & \image{CHM_joint_unet}& \image{DTM_joint_unet}  \\
        \rotatebox{90}{\parbox[c]{3cm}{\centering{Unified CATSNet}}} &  \image{CHM_joint_boss}& \image{DTM_joint_boss}   \\
        
	\end{tabular}
	\caption{Full-Polarization results on Paracou. Joint distribution within the reference LiDAR-based values and heights estimated by CATSNet and Unified CATSNet. The left column is for forest heights and the right column is for ground heights. The black line represents the ideal prediction.}
	\label{fig: unijoint}
\end{figure}

Experiments on Unified CATSNet are also conducted over the Paracou area in the full polarization model. The 3D reconstruction results are shown in Fig. ~\ref{fig: uni3D} where the left column shows the comparison of CATSNet and Unified CATSNet with LiDAR values as the reference on the forest height estimation. For the forest, it is hard to visually observe the performance difference and decide on a better one. Quantitatively, Unified CATSNet (RMSE = 1.9156 m) performs a little better than CATSNet (RMSE = 2.0220 m).
Similarly, the ground height estimation results are shown in the right column of Fig.~\ref{fig: uni3D}, CATSNet and Unified CATSNet both have excellent agreement with the LiDAR-based DTM. Regarding RMSE, CATSNet is equipped with 1.1365 m which is shown in Table~\ref{RMSEgroundParacou} and the RMSE of unified CATSNet is 1.3618 m. Combining the joint distribution shown in Fig.~\ref{fig: unijoint}, all results demonstrate that Unified CATSNet is reliable for both forest and ground heights estimation in one conduct. Additionally, compared with CATSNet, we can interestingly notice that unified CATSNet shows a little better performance for forest height estimation and a little worse performance for ground height estimation. This suggests that in case of limited computation resources, unified CATSNet can be a good alternative to the separated ones.


Moreover, it has to be noted that in the training process, we didn't put any balanced weight on the loss functions for the forest height prediction or the ground height prediction, so CATSNet automatically chose a compromised performance balance between the forest height and ground height prediction. To this end, users can choose the CATSNet model and the loss function based on their needs in terms of performance and computation loads.

\section{Generalization on new areas: fine-tuning strategy}

In the previous section, CATSNet has proven adept at measuring forest and ground heights with TropiSAR data on FP, DP, and SP models. To evaluate the robustness of CATSNet to areas with diverse characteristics and different acquisition parameters such as baselines, wavelength, flight height, etc., this section provides the experimental results of CATSNet on another area, Lopè, with AfriSAR data. Considering the outcomes of Section \ref{sec: unified} suggest the Unified CATSNet being a sub-optimal solution, we prefer to focus on CATSNet for such evaluation. In particular, the generalization ability of CATSNet is tested following two approaches: training from scratch and fine-tuning. The former consists of directly training from scratch the proposed solution on the new dataset; instead, the latter considers the CATSNet pre-trained on Paracou data and fine-tuning it on the new dataset.

\subsection{Training from the scratch}
\begin{figure*}[ht!]
\centering
	\begin{tabular}{cc|cc}
        \multicolumn{2}{c}{LiDAR-CHM} &\multicolumn{2}{c}{LiDAR-DTM} \\
        \multicolumn{2}{c}
        {\image{LiDAR_CHM_3Dafri}}
        &\multicolumn{2}{c}{\image{LiDAR_DTM_3Dafri}} \\
        \image{unet_CHM_3D_ogafri}&\image{unet_CHM_3Dafri}    & 
        \image{unet_DTM_3D_ogafri} & \image{unet_DTM_3Dafri}   \\
        CATSNet & CATSNet + Fine-tuning &CATSNet & CATSNet + Fine-tuning\\  
        \image{fcn_CHM_3D_ogafri} &
        \image{fcn_CHM_3Dafri}& 
        \image{fcn_DTM_3D_ogafri} &
        \image{fcn_DTM_3Dafri} \\
        TSNN & TSNN + Fine-tuning & TSNN & TSNN + Fine-tuning\\
        \multicolumn{2}{c}{
        \begin{tabular}{c}
		\includegraphics[width = 0.3\textwidth, clip=true, trim = 0mm 0mm 0mm 150mm]{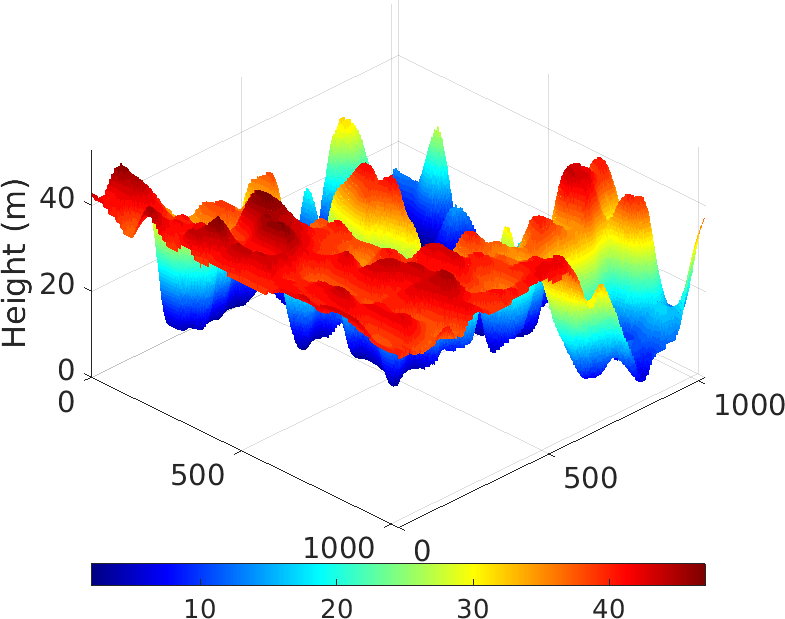}
         \end{tabular}
         } &  \multicolumn{2}{c}{
        \begin{tabular}{c}
		\includegraphics[width = 0.3\textwidth, clip=true, trim = 0mm 0mm 0mm 150mm]{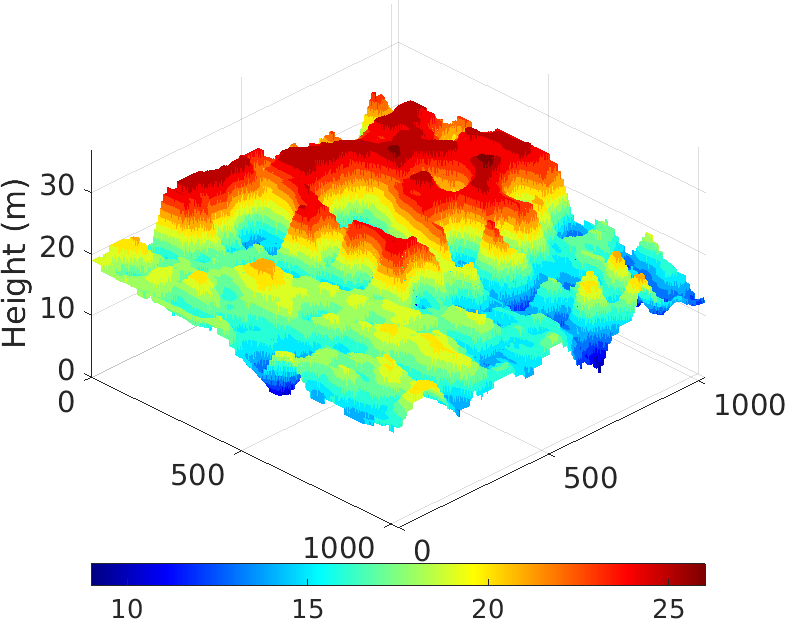}
         \end{tabular}
         }  \\
	\end{tabular}
	\caption{Full-Polarization results on Lopè. The comparison within the reference LiDAR-based values and heights estimated by CATSNet, CATSNet + Fine-tuning, TSNN, and TSNN + Fine-tuning. The left column is for forest heights and the right column is for ground heights.}
	\label{fig: Afri3D}
\end{figure*}

In particular, CATSNet and TSNN have been retrained on the Lopè dataset and tested on the extracted testing patch.
The results of the forest and ground reconstruction are shown in Fig. \ref{fig: Afri3D} (the first and the third column).

\begin{figure*}[ht!]
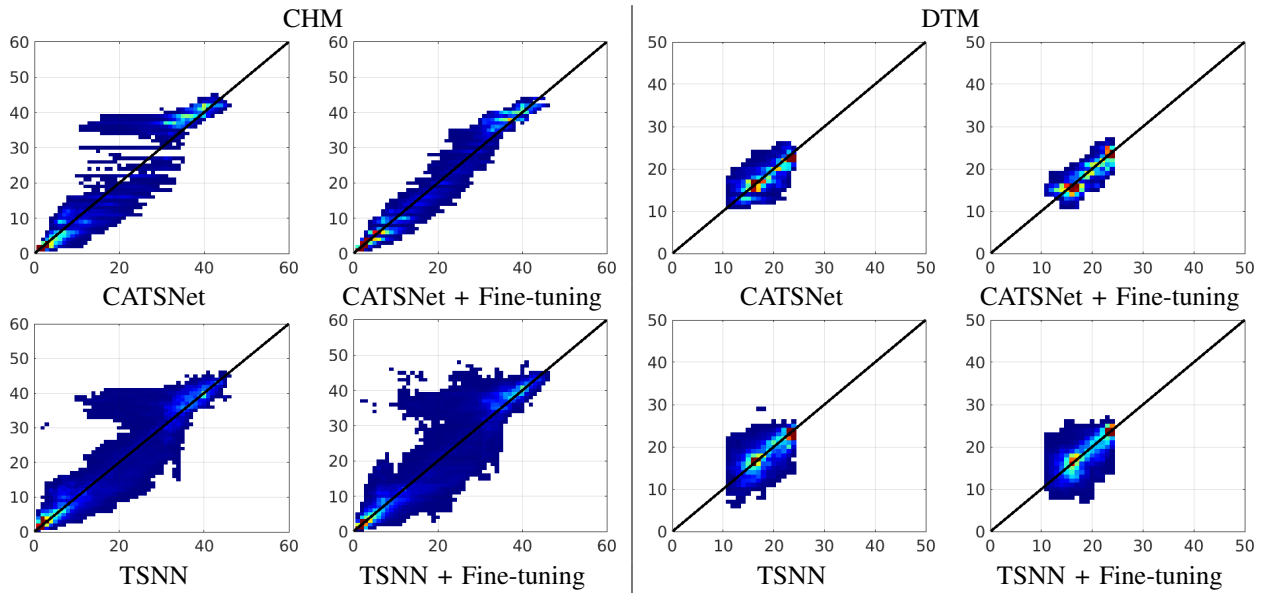

\centering
	\begin{tabular}{cc|cc}
        \multicolumn{2}{c|}{CHM} & \multicolumn{2}{c}{DTM} \\
        \image{CHM_joint_unet_fineog}& \image{CHM_joint_unet_fine} & \image{DTM_joint_unet_fineog}& \image{DTM_joint_unet_fine}  \\
        CATSNet & CATSNet + Fine-tuning &  CATSNet & CATSNet + Fine-tuning \\
        \image{CHM_joint_fcn_fineog} & \image{CHM_joint_fcn_fine}& \image{DTM_joint_fcn_fineog}& \image{DTM_joint_fcn_fine}\\
                TSNN  & TSNN  + Fine-tuning &  TSNN & TSNN + Fine-tuning \\

	\end{tabular}
	\caption{Joint distribution between predicted and estimated results on Lopè: CHM in the left column and DTM on the right one. The black line represents the ideal prediction.}
	\label{fig: TropiJoint-fine}
\end{figure*}

         

Although the robustness of CATSNet is evident in both reconstructions, at a more detailed analysis some difficulties arise also for the proposed solution. Considering the joint distribution shown in Figure \ref{fig: TropiJoint-fine} (first and the third column), both forest and ground suffer from many misleading predictions. The joint distribution of CHM shows many outliers, while DTM joint distributions are more concentrated but far from the qualitative results achieved on Paracou data.

         

These results show how different datasets may lead to different quality outcomes. There could be several causes: different numbers of training examples, nonidentical vertical resolutions, diverse coherence among stack acquisition, and so on. 

\subsection{Fine-tuning}
To improve the performance on different datasets, a fine-tuning step may be performed on a robust pre-trained solution. In particular, taking advantage of the good results on Paracou, instead of training the proposed solution from scratch, its pre-trained model on Paracou has been used as a starting point and fine-tuned on the Lopè data. In comparison, the same procedure has been followed for TSNN. The obtained results are shown in Figures \ref{fig: Afri3D}-\ref{fig: TropiJoint-fine} in the second and fourth columns. With a better starting point (Paracou pre-training), the flexibility and adaptability of CATSNet are still more evident. The profile reconstruction provided by CATSNet is getting closer to the reference following smoother variability for the CHM and reducing the underestimation in DTM. Instead, TSNN seems not robust from this perspective, since its results seem to not have any gain from the fine-tuning strategy. The distribution of predicted points lay over the black line much better thanks to the fine-tuning procedure, while the same can not be seen for TSNN.

Let's examine quantitative results, RMSE. For the forest height estimation, the RMSE of CATSNet is around 0.2 m worse than TSNN. With the operation of fine-tuning, CATSNet's performance has been boosted by around 1.3 m. On the contrary, TSNN with fine-tuning doesn't show any advantage over RMSE. In the same vein, for the ground height measurement, CATSNet has better performance with 0.4 m higher accuracy than TSNN. Importantly, the accuracy of ground height measurement is further improved by CATSNet with fine-tuning. TSNN shows similar performance on the ground height estimation; the RMSE is around 2.5 m with and without fine-tuning. 

As we can see, this is a fundamental aspect when designing a deep learning solution. Deep learning solutions are powerful but, as data-driven approaches, need high-quality datasets. In this case, the ability of CATSNet to overcome the issue related to Lopè training with a simple fine-tuning step indicates that a patch-based approach taking into account neighborhood information allows an easy extension and robustness to several user testing cases.

\begin{table}[ht]
  \caption{RMSE of the forest height measurement on Lopè}
  \centering 
  \begin{threeparttable}
    \begin{tabular}{cccc}
    Campaign  & Experiments & CATSNet  & TSNN \\
     \midrule\midrule
    \cmidrule(l r ){1-4}
     AfriSAR & Full-pol &  4.0221 & 3.8389 \\ 
    \cmidrule(l r ){1-4}
     AfriSAR & Fine-tuning &  2.7754 & 3.9101 \\
\midrule\midrule 
 \end{tabular}
\end{threeparttable}
\label{RMSEforestLope}
\end{table}

\begin{table}[ht]
  \caption{RMSE of the ground height measurement on Lopè}
  \centering 
  \begin{threeparttable}
    \begin{tabular}{cccc}
    Campaign  & Experiments & CATSNet  & TSNN \\
     \midrule\midrule
    \cmidrule(l r ){1-4}
     AfriSAR & Full-pol & 2.1912  & 2.5736 \\ 
    \cmidrule(l r ){1-4}
AfriSAR & Fine-tuning &  2.0980 & 2.5979\\
    \midrule\midrule
    \end{tabular}
\end{threeparttable} 
\label{RMSEgroundLope}
\end{table}

\section{Conclusion}
In this study, we propose to apply the U-Net model for height estimation based on the TomoSAR data set and name this method as CATSNet. CATSNet formulates the inversion of the forest and ground heights as the classification task: the elements of the covariance matrix of MB TomoSAR data patches act as the input, and the LiDAR-based heights are used for the reference. It needs to be underlined that CATSNet is conducted at the patch level enables making use of the context and achieves good localization accuracy.

Firstly, CATSNet has been trained and tested in the Paracou area with FP, DP, and SP TropiSAR data. In the comparison with another deep learning-based method, TSNN, and two typical TomoSAR techniques, SKP and GLRT, CATSNet shows the capability to generate precise forest and ground height as well as maintain good spatial resolution, while the pixel-wise TSNN suffers from noisy prediction. Traditional TomoSAR methods are confined to complex models, have high computation costs, and have worse performance. More importantly, not only with FP TomoSAR data but also with DP and SP TomoSAR data, CATSNet shows compelling performance for height construction compared to the other methods. To this end, CATSNet is a valid candidate for height measurement in the situation of limited Pol-TomoSAR data.

In addition, experiments in the Lopè area with AfriSAR data reinforce the robustness of CATSNet, which works well with data sets acquired with different parameters. CATSNet fine-tuning provides us with an alternative solution to apply deep learning-based CATSNet with fewer data sets, broadening the application scenarios. Lastly, unified CATSNet is also recommended for height reconstruction with a lower computation load and easier experimental operation.

In conclusion, we deem that CATSNet is a versatile patch-wise approach with good performance for diverse application situations and needs.

\section*{Acknowledgments}
 
The authors would like to thank the Europe Space Agency (ESA) for providing the TropiSAR and AfriSAR P-band TomoSAR data sets and thank the NASA/ESA/DLR/AGEOS Afrisar Campaign for providing the LiDAR data. Lastly, thank Chuanjun Wu for his kind help with LiDAR data.

\bibliographystyle{IEEE}
\bibliography{refsDL,refsSAR}

\end{document}